\documentclass[letterpaper]{article} 
\usepackage{aaai25}  
\usepackage{times}  
\usepackage{helvet}  
\usepackage{courier}  
\usepackage[hyphens]{url}  
\usepackage{graphicx} 
\usepackage{natbib}  
\usepackage{caption} 
\usepackage{algorithm}
\usepackage{algorithmic}
\usepackage[utf8]{inputenc} 
\usepackage{url}            
\usepackage{booktabs}       
\usepackage{amsfonts}       
\usepackage{nicefrac}       
\usepackage{microtype}      
\usepackage{xcolor}         
\usepackage{amsmath}
\usepackage{amsthm}
\usepackage{amssymb}
\usepackage{amsfonts}
\usepackage{makecell}
\usepackage{bbding}
\usepackage{rotating}
\usepackage{caption}
\usepackage{bm}
\usepackage{tabularx}
\usepackage{subcaption}
\usepackage{tikz}

\def\etal{\emph{et al.}\ }
\newtheorem{theorem}{Proposition}
\newcommand{\norm}[1]{\left\| #1 \right\|}

\usepackage{newfloat}
\usepackage{listings}

\DeclareCaptionStyle{ruled}{labelfont=normalfont,labelsep=colon,strut=off} 
\floatstyle{ruled}
\newfloat{listing}{tb}{lst}{}
\floatname{listing}{Listing}

\title{CODE: Confident Ordinary Differential Editing}
\author{
    Bastien van Delft\textsuperscript{\rm 1},
    Tommaso Martorella\textsuperscript{\rm 1},
    Alexandre Alahi\textsuperscript{\rm 1}
}
\affiliations{
    \textsuperscript{\rm 1}École Polytechnique Fédérale de Lausanne (EPFL)\\
    {firstname.lastname}@epfl.ch
}

\begin{document}
\maketitle

\begin{abstract}
\label{sec:abstract}
Conditioning image generation facilitates seamless editing and the creation of photorealistic images. However, conditioning on noisy or Out-of-Distribution (OoD) images poses significant challenges, particularly in balancing fidelity to the input and realism of the output.
We introduce Confident Ordinary Differential Editing (CODE), a novel approach for image synthesis that effectively handles OoD guidance images. Utilizing a diffusion model as a generative prior, CODE enhances images through score-based updates along the probability-flow Ordinary Differential Equation (ODE) trajectory. This method requires no task-specific training, no handcrafted modules, and no assumptions regarding the corruptions affecting the conditioning image. Our method is compatible with any diffusion model.
Positioned at the intersection of conditional image generation and blind image restoration, CODE operates in a fully blind manner, relying solely on a pre-trained generative model. Our method introduces an alternative approach to blind restoration: instead of targeting a specific ground truth image based on assumptions about the underlying corruption, CODE aims to increase the likelihood of the input image while maintaining fidelity. This results in the most probable in-distribution image around the input.
Our contributions are twofold. First, CODE introduces a novel editing method based on ODE, providing enhanced control, realism, and fidelity compared to its SDE-based counterpart. Second, we introduce a confidence interval-based clipping method, which improves CODE's effectiveness by allowing it to disregard certain pixels or information, thus enhancing the restoration process in a blind manner. 
Experimental results demonstrate CODE's effectiveness over existing methods, particularly in scenarios involving severe degradation or OoD inputs. 
\end{abstract}

%
\begin{links}
    \link{Website}{https://vita-epfl.github.io/CODE/}
    \link{Code}{https://github.com/vita-epfl/CODE/}
\end{links}

\begin{figure}[ht!]
    \centering
    \includegraphics[width=\linewidth]{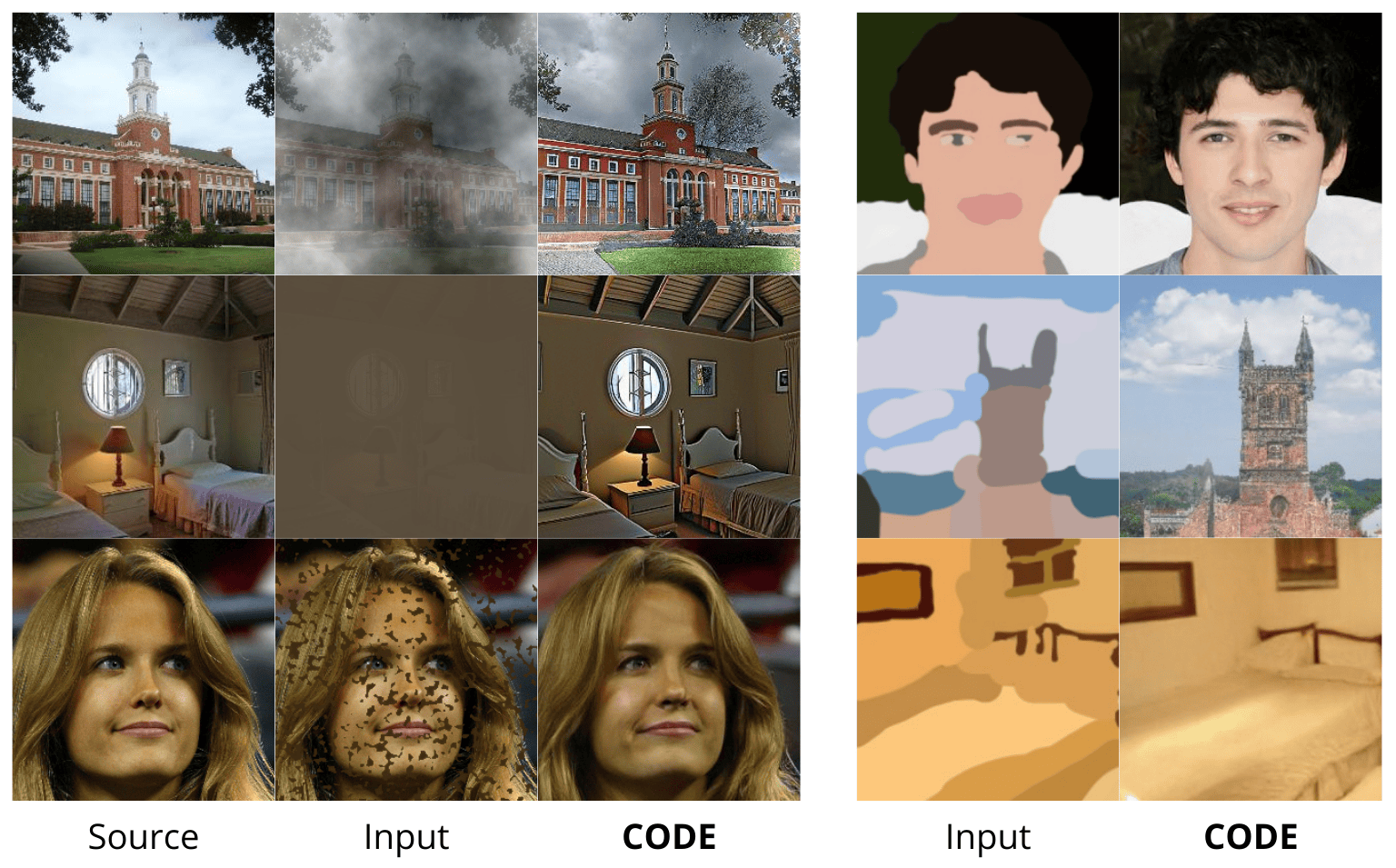}

    \label{fig:pull_fig}
    \caption{CODE: a conditional image generation framework for robust Out-of-Distribution image guidance.}
\end{figure}

\section{Introduction}
Conditional image generation consists of guiding the creation of content using different sorts of conditioning, such as text, images, or segmentation maps. Our research focuses on scenarios where the guidance is an Out-of-Distribution (OoD) image relative to the training data distribution. This is especially relevant for handling corrupted images, similar to denoising or restoration methods. The main challenge in these scenarios is balancing fidelity to the input with realism in the generated images.
Traditional methods for restoring corrupted images, such as Image-to-Image Translation or Style Transfer, are limited by the need for distinct datasets per style or per noise. Another approach models the corruption function as an inverse problem, requiring detailed knowledge of each possible corruption, making it impractical for most real unknown OoD scenarios. Guided image synthesis for OoD inputs aims to rectify corrupted images without prior knowledge of the corruption, positioning it as Blind Image Restoration (BIR). Despite recent advancements, achieving human-level generalization remains challenging.

Our work aims to generate realistic and plausible images from potentially corrupted inputs using only a pre-trained generative model, without additional data augmentation or finetuning on corrupted data, \textit{\textbf{and without any specific assumption about the corruptions.}} Unlike other BIR methods that strive to reconstruct a ground-truth image relying on specific guidance or human-based assumptions, our approach is fully blind, seeking to maximize the input image's likelihood while minimizing modifications to the input image. As such, we differ from traditional BIR approaches.

BIR is inherently ill-posed due to the loss of information from unknown degradation, necessitating auxiliary information to enhance restoration quality. Previous approaches have incorporated domain-specific priors such as facial heatmaps or landmarks \cite{fsrnet,chen2021progressive,heatmap}, but these degrade with increased degradation and lack versatility. Generative priors from pre-trained models like GANs \cite{chan2021glean,zhou2022towards,yang2021gan,luo,pulse} become unstable with severe degradation, leading to unrealistic reconstructions. Methods like \cite{gfpgan} combine facial priors with generative priors to improve fidelity but fail under extreme degradation.
In \cite{meng2021sdedit}, the authors replace GANs with diffusion models \cite{ho2020denoising, song2020score} as generative priors. However, as the degradation increases, the method forces a choice between realism and fidelity.

BIR still fails to achieve faithful and realistic reconstruction for a wide range of corruptions on a wide range of images. Dealing with various unknown corruptions prevents inverse methods from being easily applicable while dealing with a wide range of images prevents relying on carefully designed domain priors. 
We introduce Confident Ordinary Differential Editing (CODE), an unsupervised method that generates faithful and realistic image reconstructions from a single degraded image without information on the degradation type, even under severe conditions. CODE leverages the generative prior of a pre-trained diffusion model without requiring additional training or finetuning. Consequently, it is compatible with any pre-trained Diffusion Model and any dataset. CODE optimizes the likelihood of the generated image while constraining the distance to the input image, framing restoration as an optimization problem.
Similar to GAN-inversion methods \cite{tov2021designing,abdal2020image2stylegan,abdal2019image2stylegan,zhu2020domain,pulse,luo}, CODE inverts the observation to a latent space before optimization but similar to SDEdit \cite{meng2021sdedit} we propose to replace GANs with diffusion models \cite{ho2020denoising, song2020score} as generative priors. Unlike GAN inversion, which relies on an auxiliary trained encoder, diffusion model inversion uses differential equations. In SDEdit, random noise is injected into the degraded observation to partially invert it in order to subsequently revert the process using a stochastic differential equation (SDE). As more noise is injected, a higher degree of realism is ensured, but at the expense of fidelity due to the additional loss of information caused by the noise randomness and the non-deterministic sampling from the SDE. We found that in some cases, as the degree of degradation increases, the method requires too high a degree of noise injection to work, forcing a choice between realism or fidelity. CODE refines SDEdit \cite{meng2021sdedit} by leveraging the probability-flow Ordinary Differential Equation (ODE) \cite{song2020score}, ensuring bijective correspondence with latent spaces. We use Langevin dynamics with score-based updates for correction, followed by the probability-flow ODE to project the adjusted latent representation back into the image space. This decouples noise injection levels, correction levels, and latent spaces, enhancing control over the editing process. Furthermore, CODE introduces a confidence-based clipping method that relies on the marginal distribution of each latent space. This method allows for the disregard of certain image information based on probability, which synergizes with our editing method. Our experimental results show CODE's superiority over SDEdit in realism and fidelity, especially in challenging scenarios.

\section{Background}
\subsection{Related Works}
A detailed comparison of the requirements of state-of-the-art methods is provided in the Appendix A.

\paragraph{Inverse Problems}
In the inverse problem setup, methods are designed to leverage sensible \emph{assumptions on the degradation operators}. When combined with powerful generative models such as diffusion models, these approaches have achieved outstanding results, setting new benchmarks in the field \cite{saharia2022palette, liang2021swinir, kawar2022denoising, murata2023gibbsddrm, zhu2023denoising, chung2023diffusion, wang2022zeroshot}. Several subcategories of the inverse problem setting, like blind linear problems and non-blind non-linear problems, drop some assumptions about the degradation operators and, therefore, extend their applicability. However, while producing exceptional results in controlled applications like deblurring and super-resolution, the necessity for assumptions on the degradation operator makes them often impractical for unknown corruptions or in real-world scenarios. Consequently, these methods are not directly applicable to our context, where such exact information is typically unavailable.  
DDRM \cite{kawar2022denoising}, DDNM \cite{wang2022zeroshot}, GibbsDDRM \cite{murata2023gibbsddrm}, DPS \cite{chung2023diffusion}, and DiffPIR \cite{zhu2023denoising} belong to this category.

\paragraph{Conditional Generative Models with Paired Data}
A parallel approach involves conditioning a generative model on a degraded image. Most methods in this category \emph{require training with pairs of degraded and natural images} \cite{mirza2014conditional, isola2017image, batzolis2021conditional, xia2023diffir, li2023bbdm, liu2023i2sb, chung2023direct}. Additionally, these methods often depend on carefully designed loss functions and guidance mechanisms to enhance performance, as demonstrated by \cite{song2020score}. Conditional Generative Adversarial Networks, as explored in \cite{isola2017image}, exemplify this approach, where generative models are trained to regenerate the original sample when conditioned on its version in another domain. However, when the degradation process is unknown or varies widely, these models struggle to generalize effectively, rendering them less applicable and not comparable to our method, which operates without such constraints. DDB \cite{chung2023direct} and I$^2$SB \cite{liu2023i2sb} fall into this category.

\paragraph{Unsupervised Bridge Problem with Unpaired Datasets}
In scenarios where two distinct datasets of \emph{clean and degraded data are available without direct paired data}, methodologies based on principles like cycle consistency and realism have been developed, as evidenced by the works of \cite{zhu2017unpaired} using GANs \cite{goodfellow2014generative} and \cite{su2023dual} using Diffusion Models \cite{ho2020denoising},\cite{sohl2015deep}. A direct application of such methods to our scenario is not feasible due to the need for datasets of degraded images, which would hamper the ability to generalize to unseen corruptions.

\paragraph{Blind Image Restoration with task-specific or domain-specific information}
Blind Image Restoration methods aim to handle a variety of degradations without restricting themselves to specific types. A recent trend in this field is the transposition of the problem into a latent space where corrections are made based on prior distributions. Notable works in this area include \cite{abdal2019image2stylegan, abdal2020image2stylegan, chan2021glean, zhu2020domain, 10205099} have explored various aspects of this approach, utilizing GAN inversion, or VAE/VQVAE encoding \cite{kingma2013auto, oord2017neural}, and have achieved significant advancements, particularly in scenarios involving light but diverse degradations. Usually, methods for blind image restoration \emph{incorporate domain-specific information} \cite{zhou2022towards, gfpgan,gu2022vqfr} or \emph{task-specific guidances} \cite{fei2023generative}.

Moreover, several methods \cite{lin2023diffbir, yang2023pgdiff} rely on the \emph{combination of many blocks trained separately} (such as Real-ESRGAN \cite{realesrgan}) and \emph{incorporate different task-specific information} (e.g., different restorers), making it even harder to ensure resilience to diverse degradations.

\paragraph{SDEdit and ILVR}
The works by Meng \etal in \cite{meng2021sdedit} and Choi \etal in \cite{choi2021ilvr} inspired the formulation of CODE. SDEdit relies on the stochastic exploration of a given input's neighborhood to yield realistic and faithful outputs within a limited editing range. This method, tested for robustness by \cite{gao2022back}, bears similarities to the proposed gradient updates within the latent space of GAN models but is grounded in more solid theoretical foundations. ILVR is an iterative conditioning method designed to generate diverse images that share semantics with a downsampled guidance image. However, it requires a clean image for downsampling, which is not feasible in our scenario where the input guidance is already corrupted. Downsampling in this context would exacerbate information loss, making ILVR unsuitable for our application.

\subsection{Preliminary - Diffusion Models}
\paragraph{Denoising Diffusion Probabilistic Models}
We denote $\mathbf{x}_0$ the data from the data distribution, in our case natural images, and $\mathbf{x}_1$, ..., $\mathbf{x}_T$ the latent variables. The forward process is in DDPM \cite{ho2020denoising} then defined by: 
\begin{equation*}
\begin{gathered}
\mathbf{x}_{t+1} = \alpha_t \cdot \mathbf{x}_t + (1-\alpha_t) \cdot \epsilon , \text{with } \epsilon \sim \mathcal{N}(0,\mathbf{I}),\\
\end{gathered}
\end{equation*}
Where $\alpha_{t}$ is a schedule predefined as an hyperparameter.\\
The diffusion model $\epsilon_{\theta}$ is then trained to minimize $ \mathbb{E}_{\mathbf{x}_t,t} \norm{\epsilon_{\theta}(\mathbf{x}_t, t) - \epsilon}.$

\paragraph{Score-based Generative Models}
In the case of Score-based Generative Models \cite{song2019generative, song2020score, song2020improved}, the model $s_{\theta}$ learns to approximate the score function, $\nabla_{\mathbf{x}} \log p(\mathbf{x})$, by minimizing:
\begin{equation*}
\begin{gathered}
\mathbb{E}_{p(\mathbf{x})} \norm{s_{\theta}(\mathbf{x}) - \nabla_{\mathbf{x}} \log p(\mathbf{x})}.
\end{gathered}
\end{equation*}
The most common approach to solve this is denoising score matching \cite{vincent}, which is further described in the Appendix C. \\
Crucially, one can sample from $p(x_t)$ while using only the score function through Langevin dynamics \cite{langevin} sampling by repeating the following update step: 
\begin{equation}
    x_{t+1} = x_{t} + \epsilon \cdot s_{\theta}(x_t , t) + \sqrt{2\epsilon} \cdot \eta \ , \text{with} \ \eta \sim \mathcal{N}(0, \sigma^2) .
\label{eq:langevin}
\end{equation}

\section{Method: Confident Ordinary Differential Editing}
\subsection{Editing with Ordinary Differential Equations}\label{sec:odedit}
\begin{figure}[htbp]
    \centering
    \includegraphics[width=\linewidth]{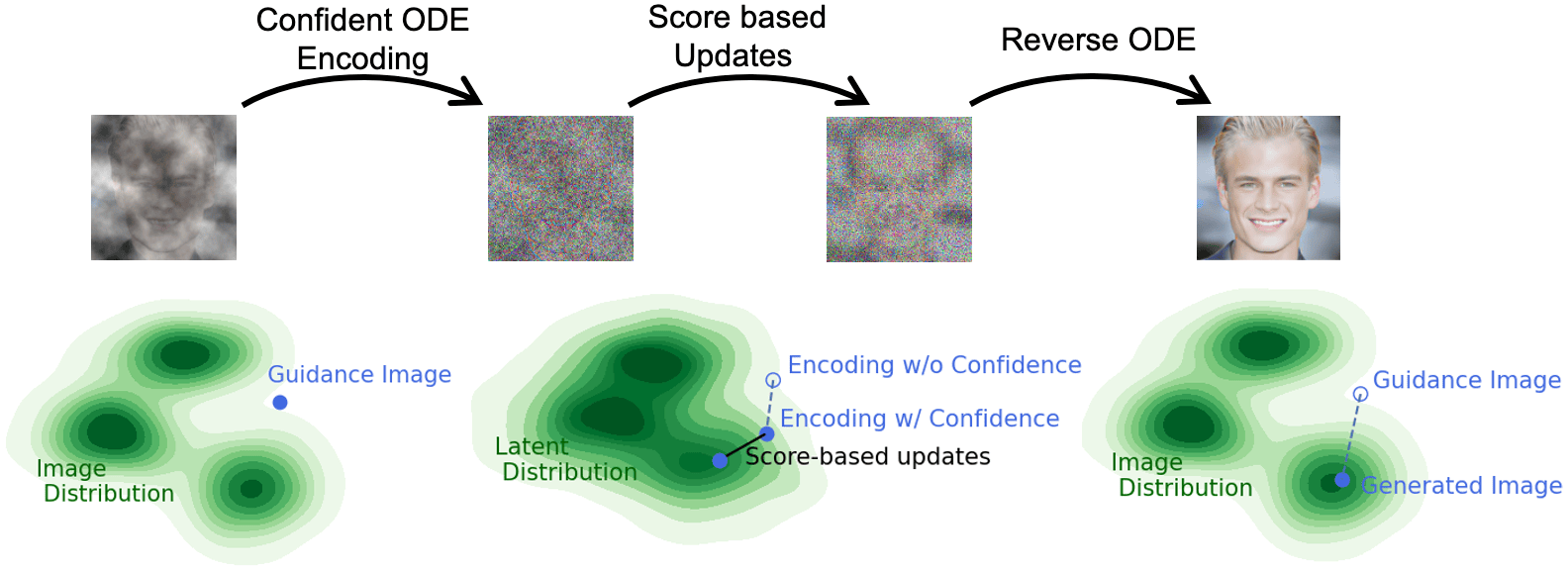}
    \caption{Editing corrupted images with ODE. The green contour plot represents the distribution of images. Given a corrupted image, we encode it into a latent space using the probability-flow ODE and our Confidence-Based Clipping. We use Langevin Dynamics in the latent space to correct the encoded image. Finally, we project the updated latent back into the visual domain.}
    \label{fig:evolution}
\end{figure}

Our approach, described in Figure \ref{fig:evolution}, formulates a theoretically grounded method for mapping OoD samples to in-distribution ones. 

\paragraph{From Gaussian Perturbation to Ordinary Inversion}

Our method draws inspiration from SDEdit \cite{meng2021sdedit} but introduces significant enhancements. SDEdit inverts the diffusion process by injecting Gaussian noise into the input image and then using this noisy image as a starting point to generate an image using DDPM \cite{ho2020denoising}. This process involves a trade-off between fidelity and realism: more noise results in more realistic images but less fidelity to the original input.

In contrast, we propose inverting the Probability-Flow Ordinary Differential Equation (ODE) as a superior alternative to noise injection. This approach maintains the fidelity of the reconstructed image by avoiding extra noise. The inversion process and its reverse operation ensure precise image reconstruction, limited only by approximation errors \cite{su2023dual}. Unlike SDEdit, which requires increasing noise levels to revert to deeper latent spaces, our method allows inversion to any latent space along the ODE trajectory while preserving image integrity. This decouples the noise injection level from the depth of inversion. We use the ODE solver from DDIMs \cite{song2020denoising} in our experiments. 

The primary motivation for inverting the degraded image is the model's ability to process out-of-distribution images. Direct estimation of the score on degraded images is impractical due to the poor performance of the score estimation on OoD data. By mapping the corrupted input back to the latent space, we obtain more accurate estimates within a distribution closely resembling a multivariate Gaussian. This concept was foundational to SDEdit; however, their reliance on noise injection prevented full inversion of the diffusion process without losing information from the observation. 

\paragraph{Langevin Dynamics in Latent Spaces}
There exists a direct correspondence between DDPM, $\epsilon_{\theta} $,  and Noise Conditional Score Network, $s_{\theta} $, such that $s_{\theta}(\mathbf{x}_t , t) = - \frac{\epsilon_{\theta} (\mathbf{x}_t , t) }{\sigma_t}.$

Building upon that, we propose to perform gradient-update in our latent spaces utilizing Langevin dynamics as in equation \eqref{eq:langevin} to increase the likelihood of our latent representation. The method is described in the Appendix D, Algorithm 2. Analogous to SDEdit and contrasting with alternative methods, our editing method can be tailored to prioritize either realism or fidelity by selecting the step size in the Langevin dynamics and the latent spaces where to optimize. 

\begin{algorithm}
\caption{CODE Simple - With Confidence-based Clipping - Algorithm 1}
\label{alg:odedit_clip}
\begin{algorithmic}
\REQUIRE $N$ \textit{(Langevin iterations)}, $\epsilon$ \textit{(step-size)}, $x_0$ \textit{(Observation)}, $L$ \textit{(L-th latent-space)}, $\eta$ \textit{(size of the confidence interval.)}
\STATE $x_{L, 0} = ODE\_SOLVER_{\textit{forward}_{0 \rightarrow L}}(Clip_{CBC_{\eta}}(x_0))$
    \FOR{$k=0$ \TO $N-1$}
        
        \STATE $x_{L,k+1} = x_{L,k} - \epsilon \cdot s_{\theta}(x_{L,k}, L) \ + \sqrt{2 \epsilon } \cdot \eta$ , where $\eta \sim \mathcal{N}(0, \mathbf{I})$ 

    \ENDFOR
     \STATE $\tilde x_{0} = ODE\_SOLVER_{\textit{backward}_{L \rightarrow 0}}(x_{L,N})$
\end{algorithmic}
\end{algorithm}
Our editing technique, relying on updates within a designated latent space, facilitates an extensive array of editing possibilities on the input image, as optimizing in one latent space yields distinct outcomes compared to optimizing in another. 
Whereas SDEdit provides a singular hyper-parameter to govern the editing process, our method bifurcates this control mechanism into two distinct parameters: the step size in the updates and the choice of latent space for optimization. This dual-parameter approach enables our editing method to equal SDEdit's performance on tasks where the latter is effective and to outperform in tasks that are unattainable for SDEdit. 

\subsection{Confidence Based Clipping (CBC)}
Here, we present a clipping method for the latent codes applied during the encoding process that does not depend on the prediction or the original sample. The proof is available in Appendix B. 

\begin{theorem}
\label{th:1}
Let $\Phi$ be the cumulative distribution function of $\mathcal{N}(0, \mathcal{I})$ and let $x_0 \in [-1, 1]$. For $\alpha_t \in [0,1] \text{, } \forall t \in [0,1]$, assume that $x_t \sim \mathcal{N}(\sqrt{\alpha_t} \cdot \alpha_0, \sqrt{1-\alpha_t} \cdot \mathcal{I})$. Then, for all $\eta$:\\
\resizebox{0.8\columnwidth}{!}{\parbox{\linewidth}{
\begin{equation*}
    \mathcal{P}(x_t \in [-\sqrt{\alpha_t} - \eta \cdot \sqrt{1-\alpha_t},\sqrt{\alpha_t} + \eta \cdot \sqrt{1-\alpha_t}]) \geq \Phi(\eta) - \Phi(-\eta).
\end{equation*}
}}\\
Specifically, for $\eta = 2$:\\
\resizebox{0.8\columnwidth}{!}{\parbox{\linewidth}{
\begin{equation*}
    \mathcal{P}(x_t \in [-\sqrt{\alpha_t} - 2 \cdot \sqrt{1-\alpha_t},\sqrt{\alpha_t} + 2 \cdot \sqrt{1-\alpha_t}]) \geq 0.95.
\end{equation*}
}}
\end{theorem}
During the encoding process, we propose to clip the latent codes using a confidence interval derived from Proposition \ref{th:1}. 
Confidence-based clipping is performed as follows:\\
\resizebox{0.8\columnwidth}{!}{\parbox{\linewidth}{
\begin{equation*}
    x_{t}^{clipped} = \text{Clip}(x_{t}, \text{min}=-\sqrt{\alpha_t} - \eta \cdot \sqrt{1-\alpha_t}, \text{max}=\sqrt{\alpha_t} + \eta \cdot \sqrt{1-\alpha_t}),
\end{equation*}
}}\\
where $t$ is the timestep, $\alpha_t$ is the predefined schedule of the DM, and $\eta$ is the chosen confidence parameter.

Similar to our editing method, CBC is agnostic to the input and suitable for blind restoration scenarios. 
We combine CBC with our ODE editing method to form our complete method, CODE, detailed in Algorithm \ref{alg:odedit_clip}. As shown in Figure \ref{fig:ablation}, the two methods synergize efficiently. It is crucial to note that CBC cannot be used in combination with SDEdit.

\section{Experiments}

\begin{figure}[ht!]
    \centering
    \begin{tabular}{@{}c@{}c@{}c@{}c@{}c@{}c@{}c@{}} %
        
        \includegraphics[width=\dimexpr\linewidth/7\relax]{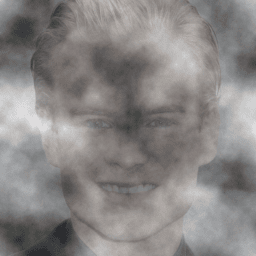} & 
        \includegraphics[width=\dimexpr\linewidth/7\relax]{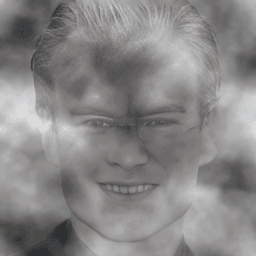} & 
        \includegraphics[width=\dimexpr\linewidth/7\relax]{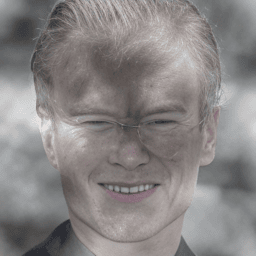} & 
        \includegraphics[width=\dimexpr\linewidth/7\relax]{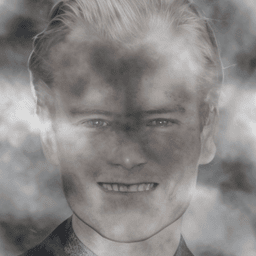} & 
        \includegraphics[width=\dimexpr\linewidth/7\relax]{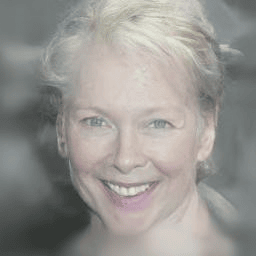} & 
        \includegraphics[width=\dimexpr\linewidth/7\relax]{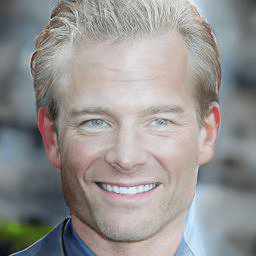} \\
        \includegraphics[width=\dimexpr\linewidth/7\relax]{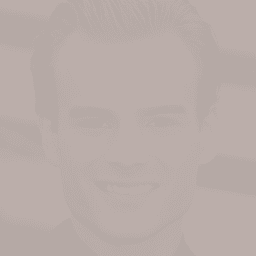} & 
        \includegraphics[width=\dimexpr\linewidth/7\relax]{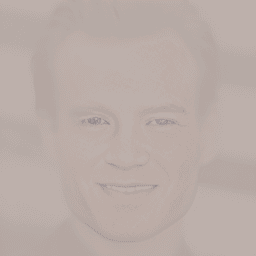} &     
        \includegraphics[width=\dimexpr\linewidth/7\relax]{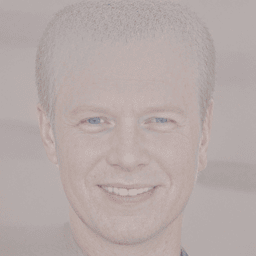} & 
        \includegraphics[width=\dimexpr\linewidth/7\relax]{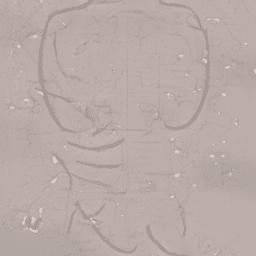} &
        \includegraphics[width=\dimexpr\linewidth/7\relax]{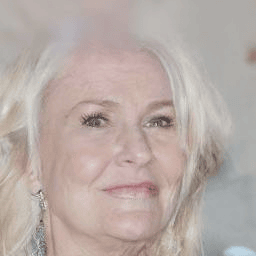} &
        \includegraphics[width=\dimexpr\linewidth/7\relax]{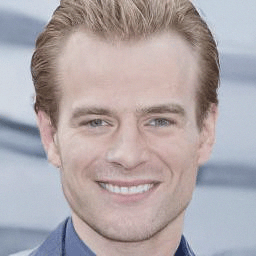} \\   
        \includegraphics[width=\dimexpr\linewidth/7\relax]{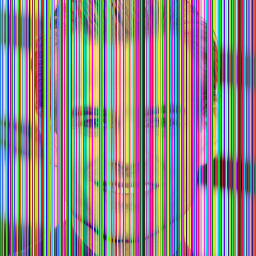} & 
        \includegraphics[width=\dimexpr\linewidth/7\relax]{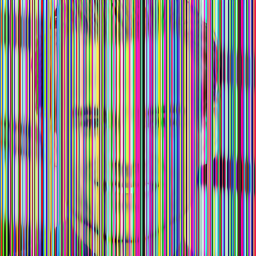} & 
        \includegraphics[width=\dimexpr\linewidth/7\relax]{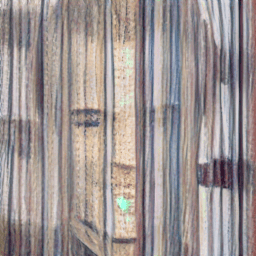} & 
        \includegraphics[width=\dimexpr\linewidth/7\relax]{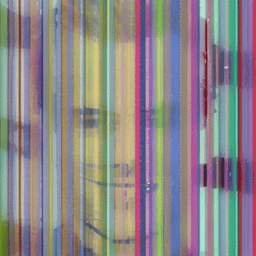} &
        \includegraphics[width=\dimexpr\linewidth/7\relax]{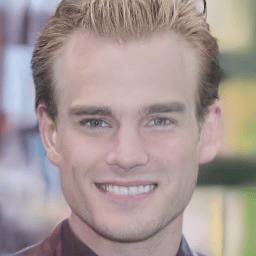} &
        \includegraphics[width=\dimexpr\linewidth/7\relax]{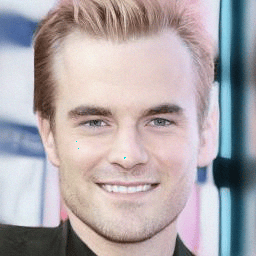}\\
        \includegraphics[width=\dimexpr\linewidth/7\relax]{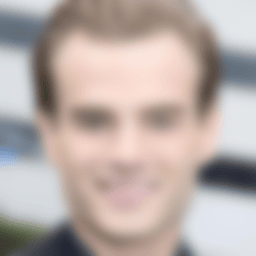} & 
        \includegraphics[width=\dimexpr\linewidth/7\relax]{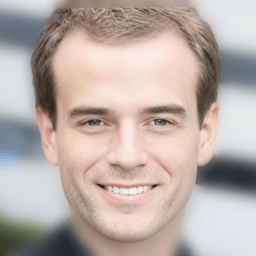} &     
        \includegraphics[width=\dimexpr\linewidth/7\relax]{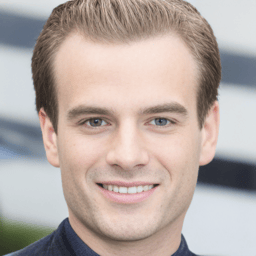} & 
        \includegraphics[width=\dimexpr\linewidth/7\relax]{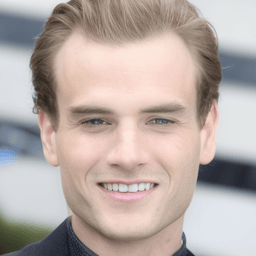} &
        \includegraphics[width=\dimexpr\linewidth/7\relax]{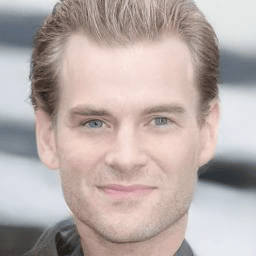} & 
        \includegraphics[width=\dimexpr\linewidth/7\relax]{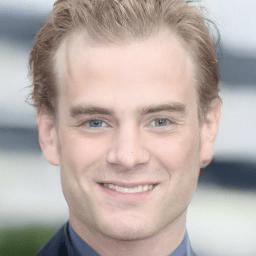} &
        \includegraphics[width=\dimexpr\linewidth/7\relax]{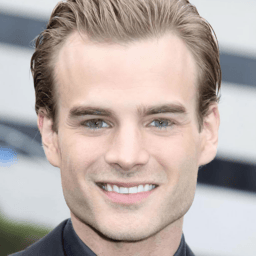} 
        \\
        \multicolumn{1}{@{}c@{}}{\scriptsize{Input}} & 
        \multicolumn{1}{@{}c@{}}{\makebox[0pt][c]{\scriptsize{GFPGAN}}}& 
        \multicolumn{1}{@{}c@{}}{\makebox[0pt][c]{\scriptsize{CodeFormer}}} & 
        \multicolumn{1}{@{}c@{}}{\makebox[0pt][c]{\scriptsize{DiffBIR}}} & 
        \multicolumn{1}{@{}c@{}}{\makebox[0pt][c]{\scriptsize{SDEdit}}} & 
        \multicolumn{1}{@{}c@{}}{\makebox[0pt][c]{\textbf{\scriptsize{CODE}}}} &
        \multicolumn{1}{@{}c@{}}{\scriptsize{Source}} \\
         & 
        \multicolumn{1}{@{}c@{}}{\makebox[0pt][c]{\scriptsize{(2021)}}}& 
        \multicolumn{1}{@{}c@{}}{\makebox[0pt][c]{\scriptsize{(2022)}}} & 
        \multicolumn{1}{@{}c@{}}{\makebox[0pt][c]{\scriptsize{(2023)}}} & 
        \multicolumn{1}{@{}c@{}}{\makebox[0pt][c]{\scriptsize{(2022)}}} & 
        \multicolumn{1}{@{}c@{}}{\makebox[0pt][c]{\textbf{\scriptsize{(ours)}}}} &

    \end{tabular}
    \caption{Visual comparison on CelebAHQ with various corruption types. CODE is the only method performing on all corruptions types, significantly improving over SDEdit on two complex corruptions, Fog and Contrast. Other baselines demonstrate lower versatility while requiring extra training.}
    \label{fig:comparison}
\end{figure}

\paragraph{Setup}

We use open-source pre-trained DDPM models \cite{ho2020denoising} from HuggingFace, specifically the EMA checkpoints of DDPM models trained on CelebA-HQ \cite{karras2018progressive}, LSUN-Bedroom, and LSUN-Church \cite{yu2016lsun}, all at 256x256 resolution. For all experiments, DDIM inversion \cite{song2020improved} with 200 steps is utilized. Enhancement follows the complete Algorithm 3 described in Appendix D. It is used with $N = 200$ Langevin iteration steps, a step size $\epsilon$ of $[10^{-2}, 10^{-3}]$ for shallow latent spaces (up to $L=40$), and $[10^{-5}, 10^{-6}]$ for deeper latent spaces ($L > 100$). We use $K = 4$ annealing steps and $\alpha = 0.8$ as the annealing coefficient. When activating CBC, we use $\eta = 1.7$. A full description of the setup being used to automatically compute the metrics is provided in Appendix E. For SDEdit, samples are generated with $L$ in $[300, 500, 700]$ steps.

We tested our approach on 47 corruption types, including 17 from \cite{hendrycks2019robustness} (noise, blur, weather, and digital artifacts) and 28 from \cite{Mintun2021OnIB}. The corruption codebases are publicly available\footnote{\url{https://github.com/hendrycks/robustness}} \footnote{\url{https://github.com/facebookresearch/augmentation-corruption/blob/fbr_main/imagenet_c_bar/corrupt.py}}. Additionally, we introduced two masking types: masking entire vertical lines and random pixels with random colors. Unlike traditional masking in masked autoencoders \cite{he2022masked}, our method does not assume knowledge of masked pixels' positions, posing a more realistic recovery task. CODE operates completely blind to the corruption type, with no knowledge of the specific task or affected pixels.

For each corruption type, we test on at least 500 corrupted images. For each image, we kept the best 4 samples generated based on PSNR with respect to the original non-degraded images. 

\begin{figure}[htbp]
    \centering
    \begin{tabular}{@{}c@{}c@{\hspace{2pt}}c@{}c@{\hspace{10pt}}c@{}c@{\hspace{2pt}}c@{}c@{}} %
        
        \includegraphics[width=0.11\linewidth]{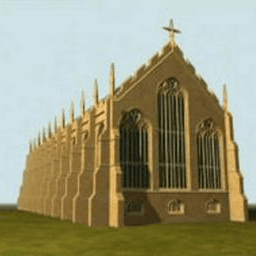} & 
        \includegraphics[width=0.11\linewidth]{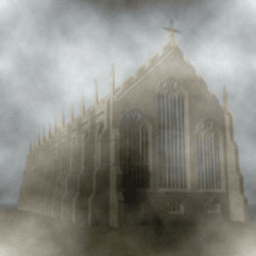} & 
        \includegraphics[width=0.11\linewidth]{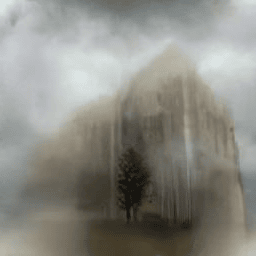} & 
        \includegraphics[width=0.11\linewidth]{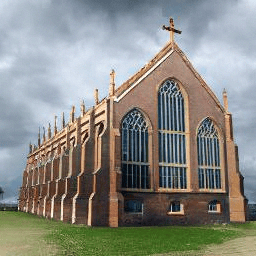} & 
        \includegraphics[width=0.11\linewidth]{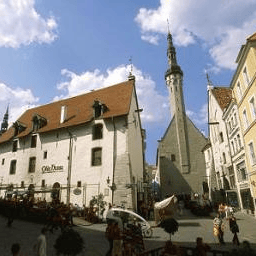} & 
        \includegraphics[width=0.11\linewidth]{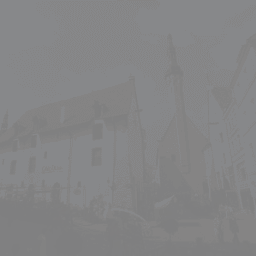} & 
        \includegraphics[width=0.11\linewidth]{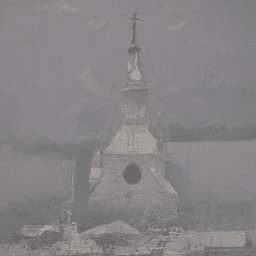} & 
        \includegraphics[width=0.11\linewidth]{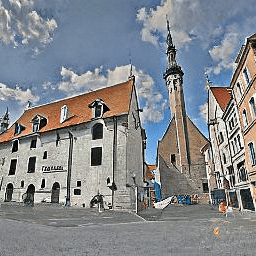}\\
        \includegraphics[width=0.11\linewidth]{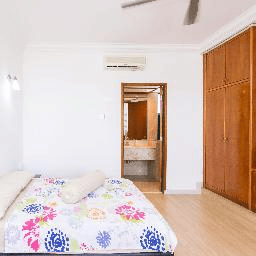} & 
        \includegraphics[width=0.11\linewidth]{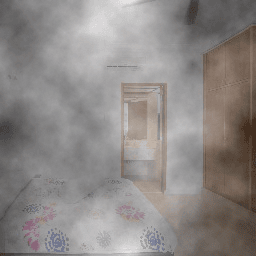} & 
        \includegraphics[width=0.11\linewidth]{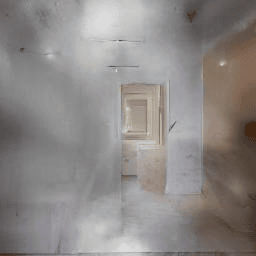} & 
        \includegraphics[width=0.11\linewidth]{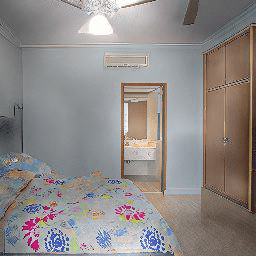} &
        \includegraphics[width=0.11\linewidth]{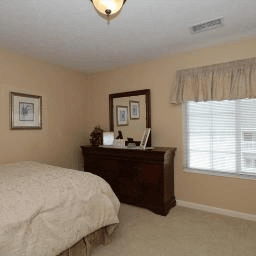} & 
        \includegraphics[width=0.11\linewidth]{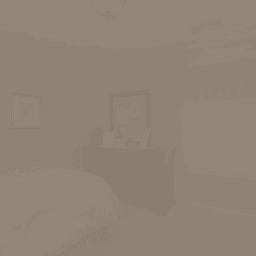} & 
        \includegraphics[width=0.11\linewidth]{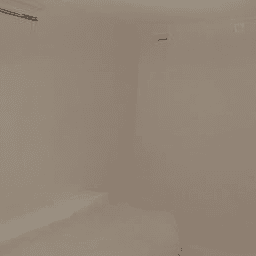} & 
        \includegraphics[width=0.11\linewidth]{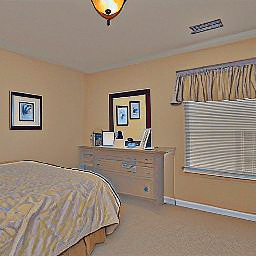}\\
        \includegraphics[width=0.11\linewidth]{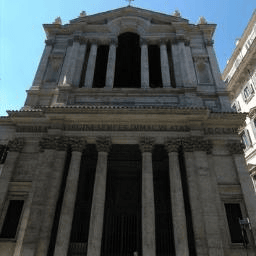} & 
        \includegraphics[width=0.11\linewidth]{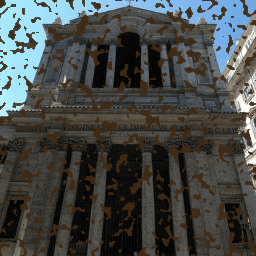} & 
        \includegraphics[width=0.11\linewidth]{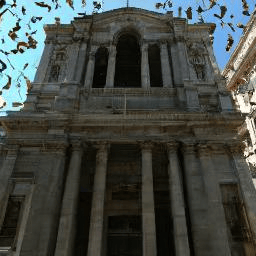} & 
        \includegraphics[width=0.11\linewidth]{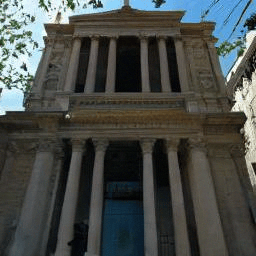} & 
        \includegraphics[width=0.11\linewidth]{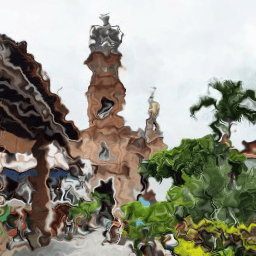} & 
        \includegraphics[width=0.11\linewidth]{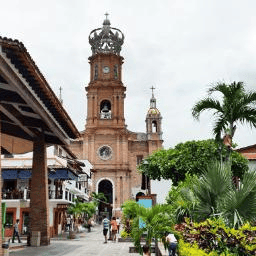} & 
        \includegraphics[width=0.11\linewidth]{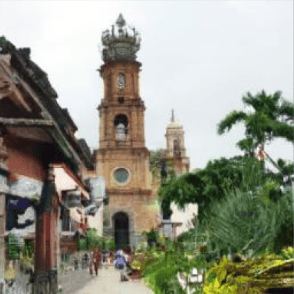} & 
        \includegraphics[width=0.11\linewidth]{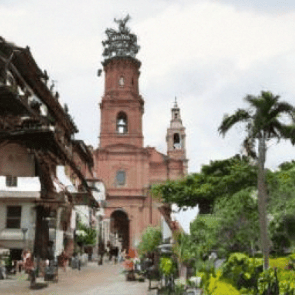} \\
        \multicolumn{1}{@{}c@{}}{\scriptsize{Source}} & 
        \multicolumn{1}{@{}c@{\hspace{2pt}}}{\scriptsize{Input}} & 
        \multicolumn{1}{@{}c@{}}{\makecell{\scriptsize{SDEdit}}} & 
        \multicolumn{1}{@{}c@{\hspace{10pt}}}{\makecell{\textbf{\scriptsize{CODE}}}} & 
        \multicolumn{1}{@{}c@{}}{\scriptsize{Source}} & 
        \multicolumn{1}{@{}c@{\hspace{2pt}}}{\scriptsize{Input}} & 
        \multicolumn{1}{@{}c@{}}{\makecell{\scriptsize{SDEdit}}} & 
        \multicolumn{1}{@{}c@{}}{\makecell{\textbf{\scriptsize{CODE}}}}

    \end{tabular}
    \caption{Visual comparison of general image restoration on various corruptions - LSUN}
    \label{fig:comparison_lsun}
\end{figure}

\paragraph{Baselines}
Our main baseline is the domain-agnostic method SDEdit \cite{meng2021sdedit}, the only one comparable to ours in terms of requirements and assumptions. On CelebAHQ, the performance is also qualitatively benchmarked against domain-specific SOTA models, namely CodeFormer \cite{zhou2022towards}, GFPGAN \cite{gfpgan}, and DiffBIR \cite{lin2023diffbir}. We also conducted visual experiments on LSUN-Bedroom and LSUN-Church to demonstrate the efficacy of CODE over diverse domains similar to SDEdit in \cite{meng2021sdedit}.

\begin{figure}[h!]
    \centering
    \includegraphics[width=\columnwidth]{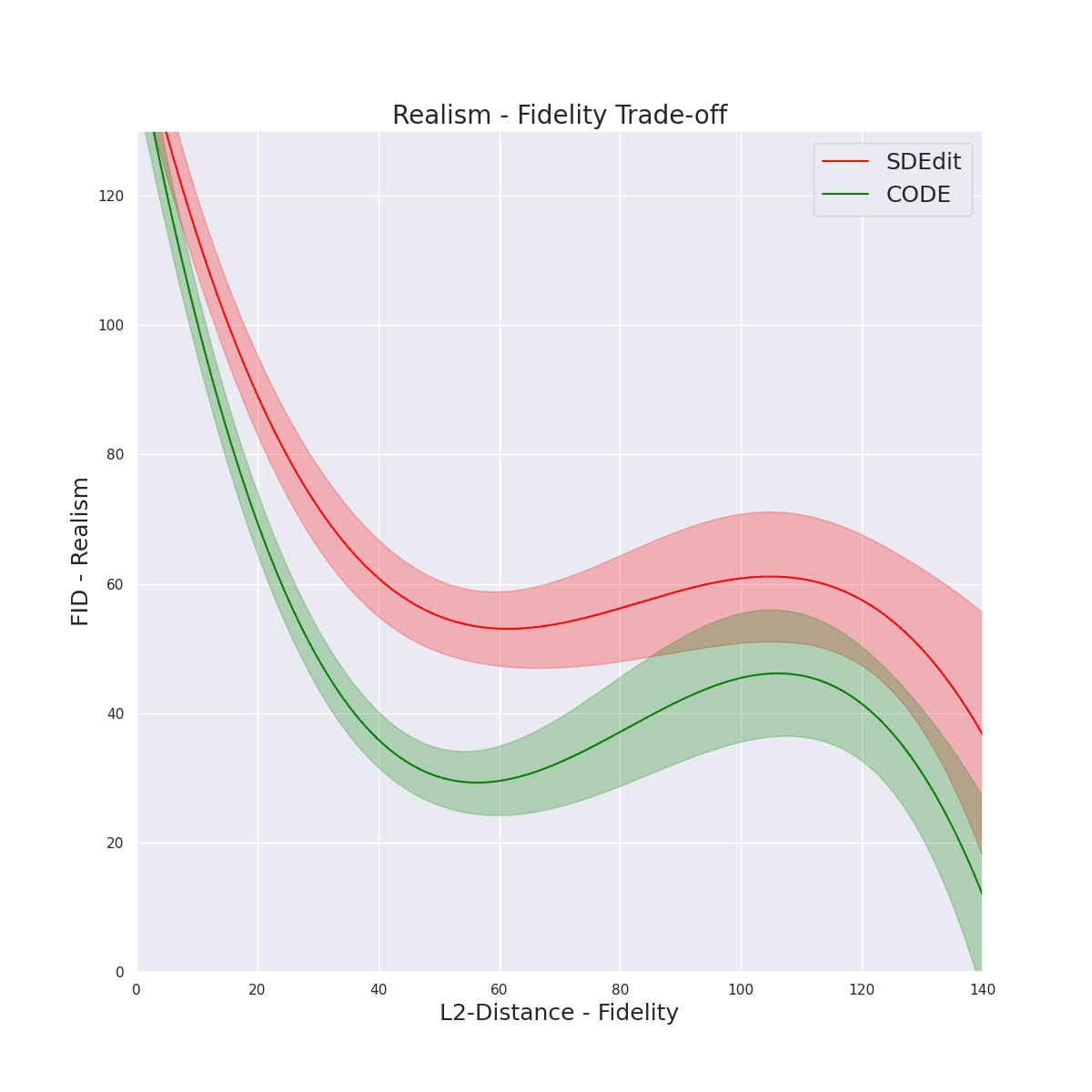}
    \caption{Comparison of realism-fidelity trade-off between SDEdit and CODE. Polynomial regression curves with shaded areas show one standard deviation. CODE produces more realistic images at the same fidelity. Both methods converge when the input distance is large, as they use the same pre-trained model.}
    \label{fig:dna}
\end{figure}

\paragraph{Evaluation Metrics}
We evaluate our results using PSNR, SSIM, LPIPS, and FID. PSNR and SSIM are measured against the corrupted image (input) to assess fidelity to the guidance. FID is used to evaluate the quality of our generated images. Given the absence of assumptions about the input and corruptions, a key metric is the trade-off between realism and fidelity—specifically, the gain in realism relative to a given loss in fidelity. To quantify this, we use L2 distance in the pixel space as a measure of fidelity and FID as a measure of realism, plotting them against each other in Figure \ref{fig:dna}. Additionally, we report LPIPS with respect to the original, non-degraded image (source) to assess reconstruction quality. This metric is particularly informative for evaluating each corruption individually, as it also reflects the complexity of the corruption, with detailed results provided in Appendix G.

\begin{table}[htbp]
    \scriptsize
    \centering
    \begin{tabularx}{\columnwidth}{@{}lX@{}X@{}X@{}X@{}}
     & LPIPS-Source $\downarrow$ &  SSIM-Input $\uparrow$ & PSNR-Input $\uparrow$ & FID $\downarrow$ \\
    \midrule[1pt]
    \textit{Inputs} 
     & 0.48 (0.35) & - & - & 143.49 (96.31) \\
    \textit{SDEdit} 
     & 0.32 (0.13) & 0.46 (0.21) & 18.74 (3.92) & 47.84 (42.29) \\
    \textit{\textbf{CODE}} 
     & \textbf{0.30} (0.12) & \textbf{0.49} (0.22) & \textbf{19.61} (4.66) & \textbf{30.66} (16.21) \\
    \bottomrule[1pt]
    \end{tabularx}
    \caption{Average values of different metrics across the 47 considered corruptions, along with the standard deviations. CODE outperforms SDEdit on all metrics. CODE preserves a higher degree of fidelity while reaching a higher degree of realism using the same pre-trained model.}
    \label{tab:metrics}
\end{table}
\paragraph{Results}
We present a brief qualitative comparison of results in Figure \ref{fig:comparison} to showcase that most methods, without further assumptions, cannot perform properly. In the vast majority of scenarios involving severe corruption like contrast, random pixel masking, or fog, only SDEdit and CODE can generate convincing images. For less intensive corruptions, which typically include erasing fine details or introducing minor noise, most baseline models tend to perform well. We provide extensive results in Appendix E.
For quantitative metrics, we focus on SDEdit and CODE and compare them using the same pre-trained model on CelebA-HQ. Consequently, the differences come only from the way the pre-trained diffusion model is leveraged.
Average metrics across the employed corruptions are detailed in Table \ref{tab:metrics}. \textbf{CODE outperforms SDEdit by 36\% in FID-score} while maintaining \textbf{a fidelity to the input (PSNR-Input) 5\% higher than SDEdit}. Moreover, the standard deviation of the FID score highlights that SDEdit fails in certain cases while CODE is more stable. 
Finally, we report in Figure \ref{fig:dna} the trade-off curves between fidelity and realism for both CODE and SDEdit. We performed a polynomial regression on CODE and SDEdit results to obtain such a curve. Both methods offer hyper-parameters to control such trade-offs. However, we highlight that CODE offers a better possibility. Overall, CODE generates more realistic outputs for a given degree of fidelity.  

\section{Ablation Study}
\subsection{Analysis of Hyperparameters}


\begin{figure}[h!]
    \centering
    \begin{tikzpicture}
        \node[anchor=south west, inner sep=0] at (0,0) {%
        \includegraphics[width=0.99\linewidth]{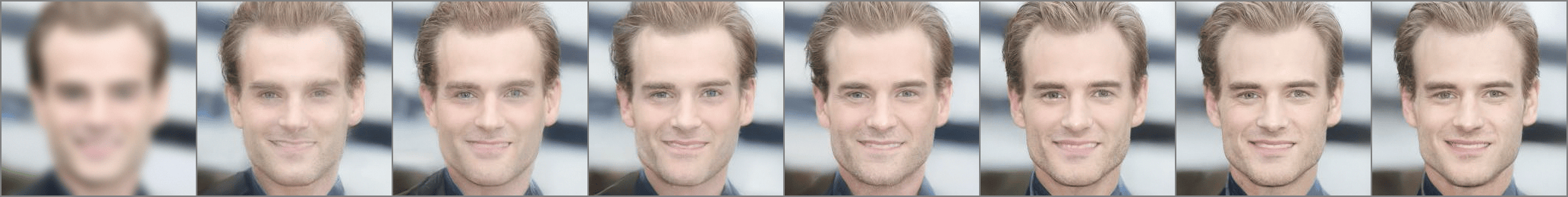}};

        \draw[->,line width=1.5pt] (0,0) -- (8.5,0) node[right] {};

        \node at (4.2,-0.5) {Number of update steps };
    \end{tikzpicture}
    \caption{Analysis of Langevin Dynamics convergence. As the number of updates increases, the output realism increases until convergence and then stabilizes.}
    \label{fig:update_steps}
\end{figure}

\paragraph{Number of Updates.}
As shown in Figure \ref{fig:update_steps}, the number of update iterations conducted in a latent space is pivotal for ensuring convergence and reducing variability. In practice, we employed 300 steps in all our experiments.

\begin{figure}[h!]
    \centering
    \captionsetup[subfigure]{width=\columnwidth, labelformat=parens} 
    \begin{subfigure}{\columnwidth}
        \centering
        \begin{tikzpicture}
            \node[anchor=south west, inner sep=0] at (-1.5,0.1) {%
            \includegraphics[width=0.15\columnwidth]{LaTeX/imgs/Comparison_Celeba/Corrupted/gaussian_blur/9998.png}};
            \node at (-0.8,-0.2) {Input};
            \node[anchor=south west, inner sep=0] at (0.2,0) {%
            \begin{tabular}{@{}c@{}c@{}c@{}c@{}}
            \includegraphics[width=0.6\linewidth]{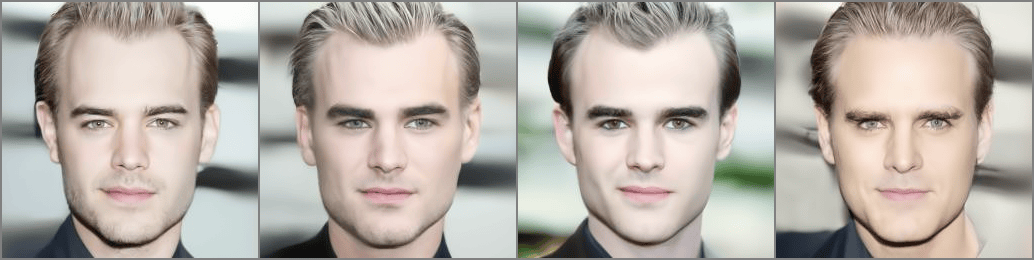} \\
            \includegraphics[width=0.6\linewidth]{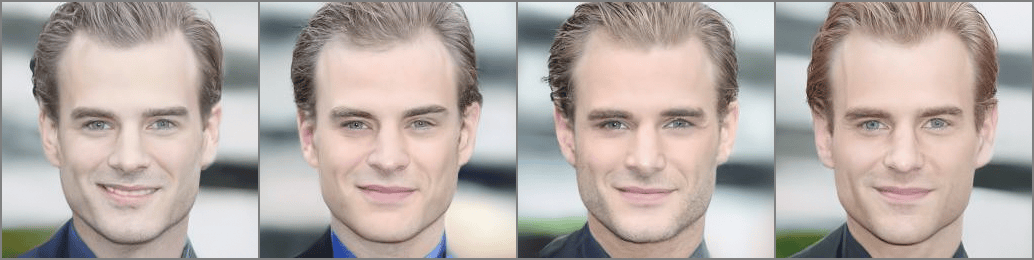} \\
            \includegraphics[width=0.6\linewidth]{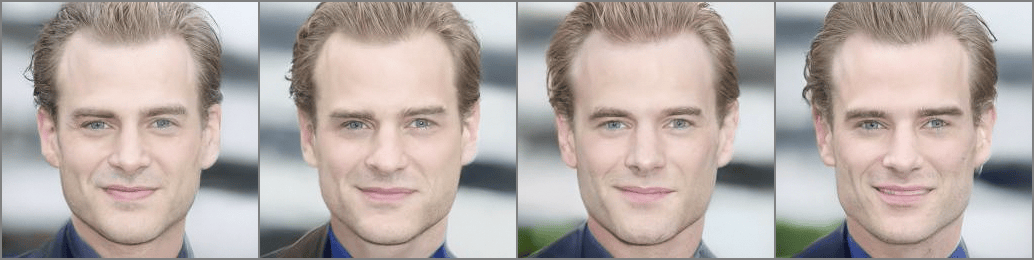} \\
            \includegraphics[width=0.6\linewidth]{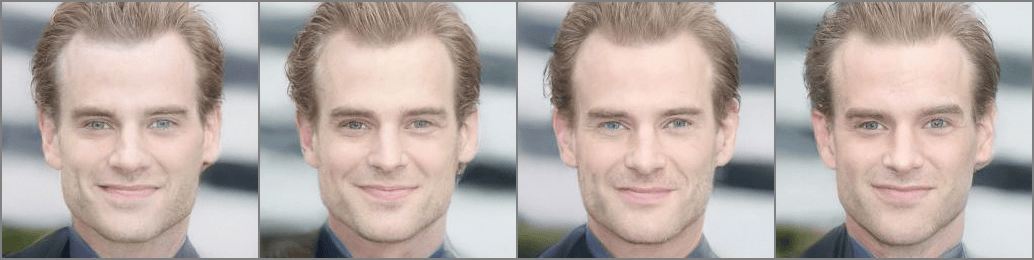} \\
            \includegraphics[width=0.6\linewidth]{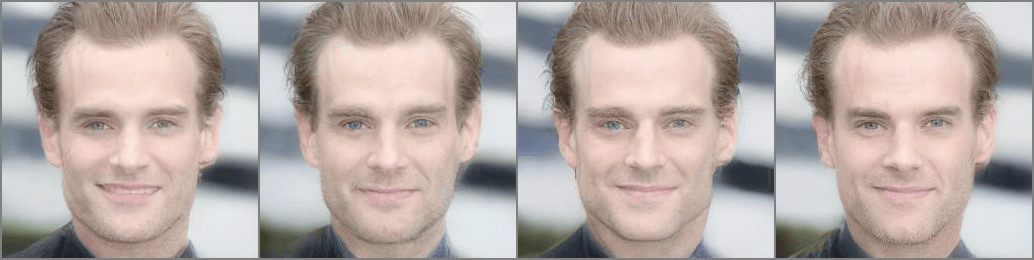} \\
            \end{tabular}};
    
            \draw[->,line width=1.5pt] (0.2,0.1) -- (5.3,0.1) node[right] {};
            \draw[->,line width=1.5pt] (0.2,0.1) -- (0.2,7.05) node[above] {};
    
            \node at (2.5,-0.2) {\# sample};
            \node[rotate=90] at (-0.3,3.0) {\makecell{Step Size \\ $\epsilon$}};
        \end{tikzpicture}
        \subcaption{Impact of step size selection}
        \label{fig:epsilon_size}

    \end{subfigure}
    \hfill
    \begin{subfigure}{0.85\columnwidth}
        \centering
        \begin{tikzpicture}
            \node[anchor=south west, inner sep=0] at (-2,0) {%
            \begin{tabular}{c}
            \includegraphics[width=0.18\linewidth]{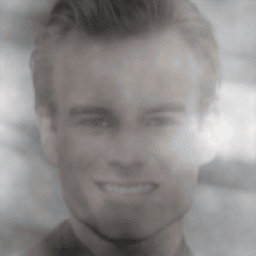} \\
            \includegraphics[width=0.18\linewidth]{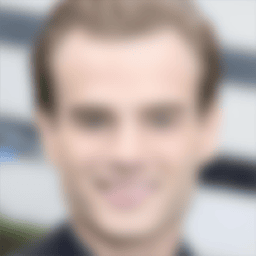} \\
            \includegraphics[width=0.18\linewidth]{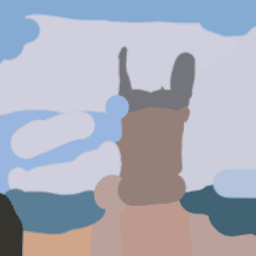}
            \end{tabular}
            }; %
            \node at (-1.1,-0.2) {Input};
            \node[anchor=south west, inner sep=0] at (0,0) {%
            \begin{tabular}{@{}c@{}c@{}c@{}c@{}}
            \includegraphics[width=0.18\linewidth]{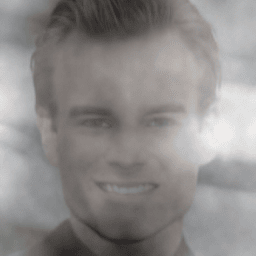} &
            \includegraphics[width=0.18\linewidth]{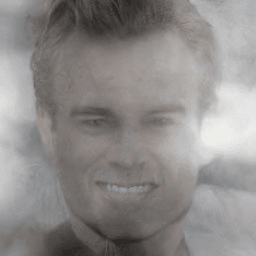} & 
            \includegraphics[width=0.18\linewidth]{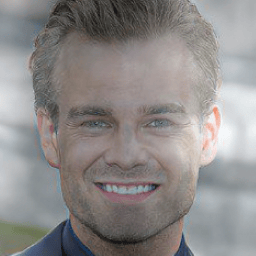} & 
            \includegraphics[width=0.18\linewidth]{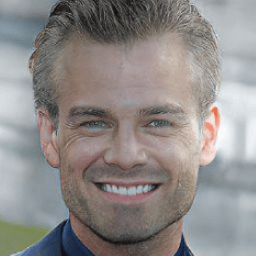}\\
            \includegraphics[width=0.18\linewidth]{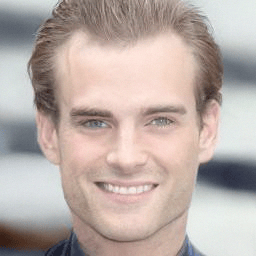} & 
            \includegraphics[width=0.18\linewidth]{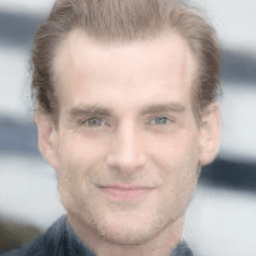} &
            \includegraphics[width=0.18\linewidth]{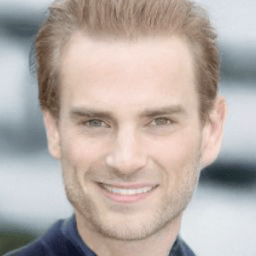} &
            \includegraphics[width=0.18\linewidth]{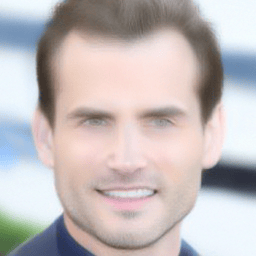}\\
            \includegraphics[width=0.18\linewidth]{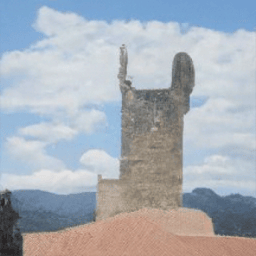} & 
            \includegraphics[width=0.18\linewidth]{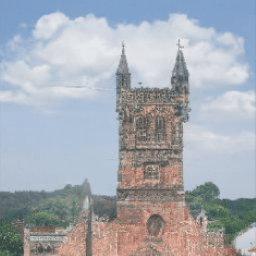} &
            \includegraphics[width=0.18\linewidth]{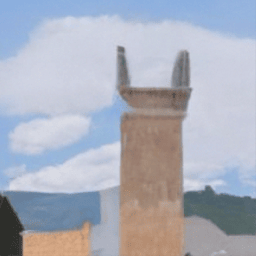} &
            \includegraphics[width=0.18\linewidth]{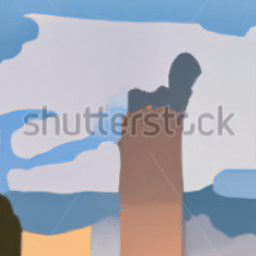}\\
            \end{tabular}
            }; %

            \draw[->,line width=1.5pt] (0,0.1) -- (5.4,0.1) node[right] {};
    
            \node at (2.65,-0.2) {Latent depth};
        \end{tikzpicture}
        
        \subcaption{Impact of latent space selection}
        \label{fig:update_latent}
    \end{subfigure}
    \caption{(a) Impact of step size on sample diversity and realism using CODE: Larger steps increase diversity but reduce fidelity. (b) Impact of latent space choice on the quality and characteristics of generated images across different corruption types.}
    \label{fig:overall_latent}
\end{figure}

\begin{figure}[h]
    \centering
    \begin{tabular}{@{}c@{\hspace{10pt}}c@{}c@{}}
        \hfill
        \includegraphics[width=0.15\linewidth]{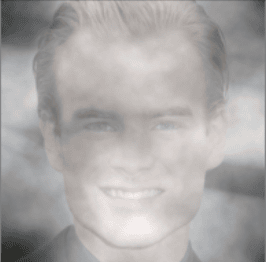} &
        \includegraphics[width=0.15\linewidth]{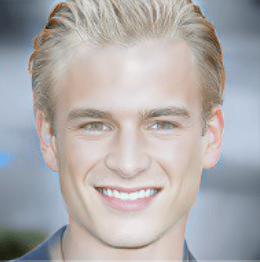} &
        \includegraphics[width=0.15\linewidth]{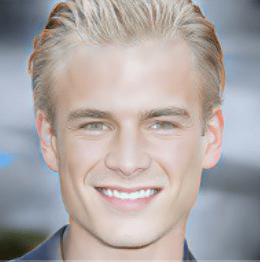} \\
        \includegraphics[width=0.15\linewidth]{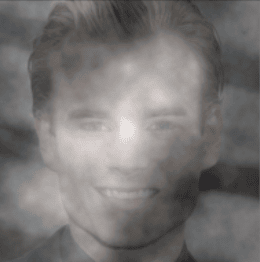} & 
        \includegraphics[width=0.15\linewidth]{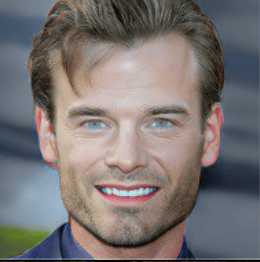} &
        \includegraphics[width=0.15\linewidth]{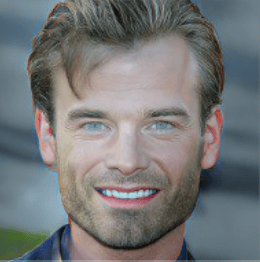} \\
        \includegraphics[width=0.15\linewidth]{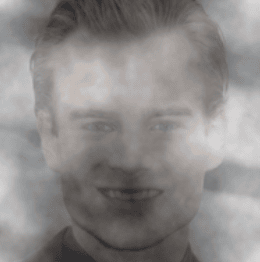} & 
        \includegraphics[width=0.15\linewidth]{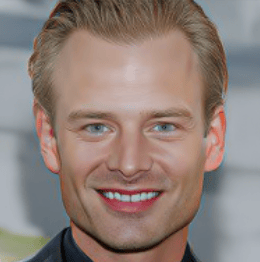} &
        \includegraphics[width=0.15\linewidth]{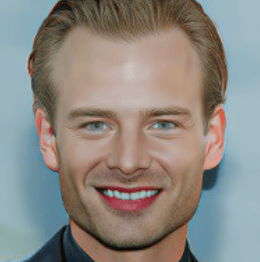} \\
        \multicolumn{1}{c@{\hspace{10pt}}}{\makecell{\scriptsize{Input}}} &
        \multicolumn{2}{c}{\makecell{\scriptsize{CODE samples}}}  
    \end{tabular}
    \caption{CODE outputs adapt faithfully to variations in the image input.}
    \label{fig:man}
\end{figure}

\paragraph{Step Size.}
The step size emerges as a critical parameter. A smaller step size results in high fidelity to the input and low variability among generated samples, albeit compromising realism. Conversely, an increased step size enhances realism and variability, as depicted in Figure \ref{fig:epsilon_size}. As the number of updates is fixed in our experiments, the step size is what governs the size of the explored neighborhood around the input image. As a result, its impact is related to the amount of noise injected in SDEdit.

\paragraph{Latent Space Choice.}
The choice of latent space significantly influences the type of changes made during updates. As shown in Figure \ref{fig:update_latent}, updates in a shallow latent space lead to minor but detailed and realistic modifications. In contrast, updates in deeper latent spaces can cause more significant or complex changes. Interestingly, regarding stroke guidance, optimization in the deepest latent space led to the addition of text and lines to the image. This suggests that the training set likely contained numerous images with these text and lines, implying that their inclusion by the model significantly enhanced the image's likelihood. Empirically, we found that the deeper the latent space, the less the notion of distance is close to an L2 pixel-based distance.\\
The optimal latent space is not one-size-fits-all but depends on the specific input being processed. For complex corruptions, using a mix of updates in different latent spaces proves most effective. On the other hand, shallow latent spaces are best for addressing simple corruptions like blur. This ability to independently select the latent space without affecting other parameters, such as the level of noise injection, is a key strength of our editing method. We disentangle what was previously a single parameter into multiple, allowing for tailored optimization on a per-sample basis. 
\paragraph{Fidelity.}
Our editing method is anchored in the corrupted sample, hence the generation is very impacted by the variations in the corruption. As shown in Figure \ref{fig:man}, the outputs are faithful to the corrupted image and do not map to a single ground-truth image.

\subsection{Ablation Study of Confidence-based clipping}

\begin{figure}[h]
    \begin{subfigure}{\columnwidth}
        \centering
        \begin{tabular}{@{}c@{}c@{\hspace{5pt}}c@{}c@{}c@{}c@{}c@{}}
            \includegraphics[width=0.13\linewidth]{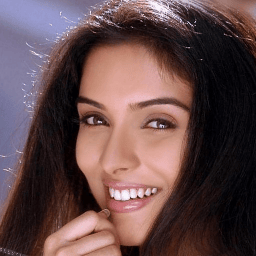} &
            \includegraphics[width=0.13\linewidth]{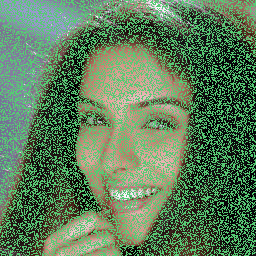} &
            \includegraphics[width=0.13\linewidth]{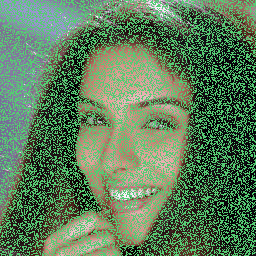} &
            \includegraphics[width=0.13\linewidth]{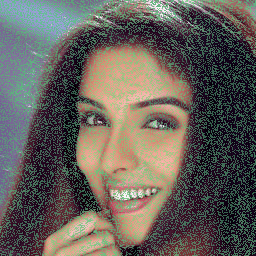} &
            \includegraphics[width=0.13\linewidth]{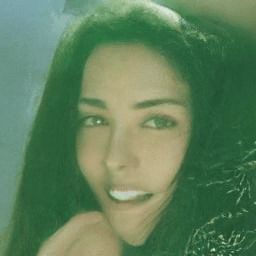} &
            \includegraphics[width=0.13\linewidth]{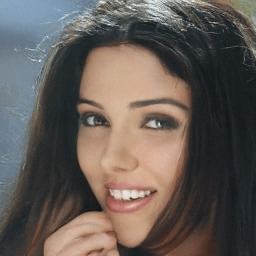} &
            \includegraphics[width=0.13\linewidth]{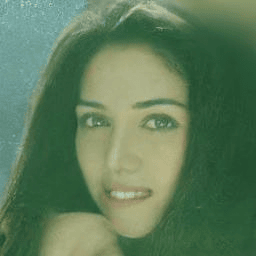} \\
            \includegraphics[width=0.13\linewidth]{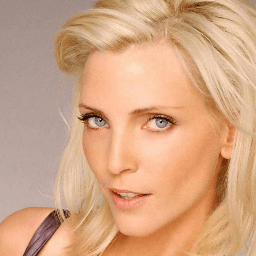} &
            \includegraphics[width=0.13\linewidth]{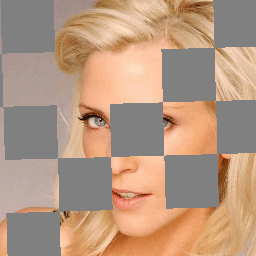} &
            \includegraphics[width=0.13\linewidth]{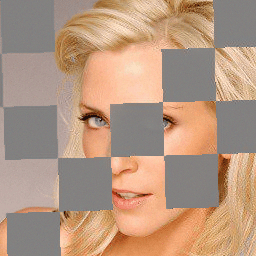} &
            \includegraphics[width=0.13\linewidth]{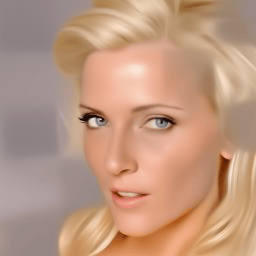} &
            \includegraphics[width=0.13\linewidth]{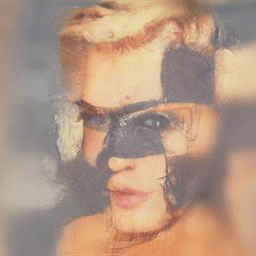} &
            \includegraphics[width=0.13\linewidth]{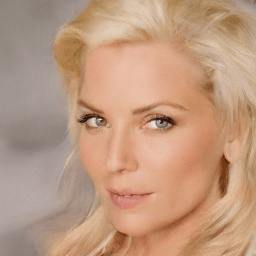} &
            \includegraphics[width=0.13\linewidth]{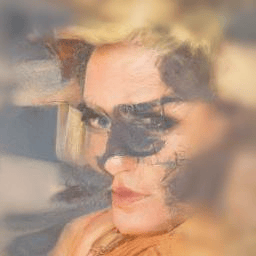} \\
            \includegraphics[width=0.13\linewidth]{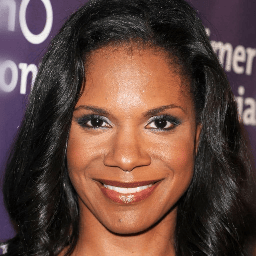} &
            \includegraphics[width=0.13\linewidth]{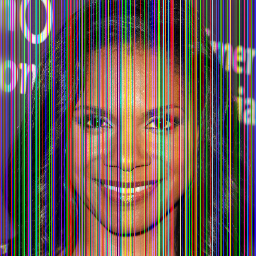} &
            \includegraphics[width=0.13\linewidth]{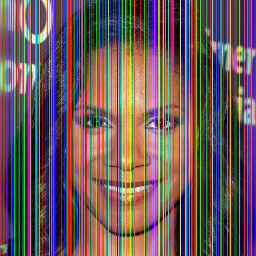} &
            \includegraphics[width=0.13\linewidth]{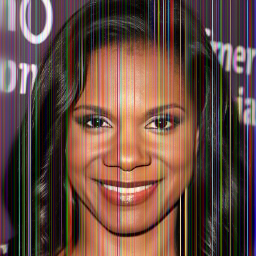} &
            \includegraphics[width=0.13\linewidth]{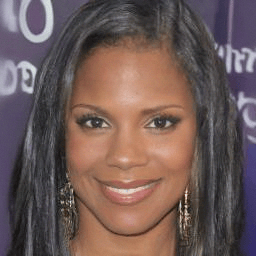} &
            \includegraphics[width=0.13\linewidth]{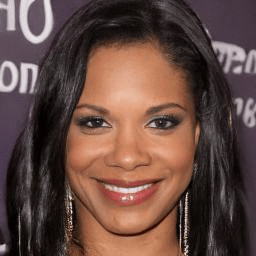} &
            \includegraphics[width=0.13\linewidth]{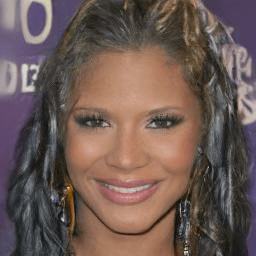} \\
            \includegraphics[width=0.13\linewidth]{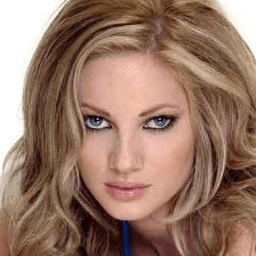} &
            \includegraphics[width=0.13\linewidth]{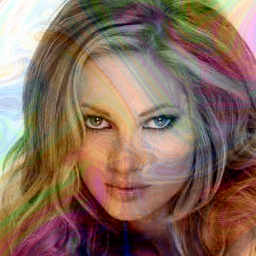} &
            \includegraphics[width=0.13\linewidth]{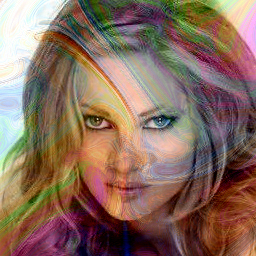} &
            \includegraphics[width=0.13\linewidth]{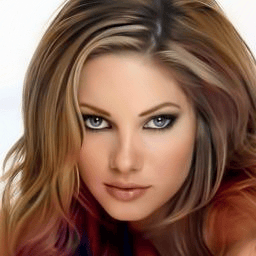} &
            \includegraphics[width=0.13\linewidth]{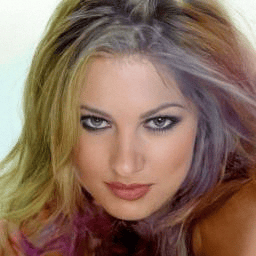} &
            \includegraphics[width=0.13\linewidth]{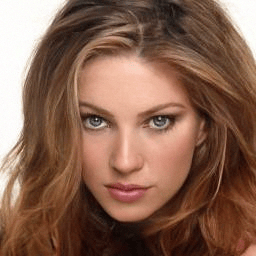} &
            \includegraphics[width=0.13\linewidth]{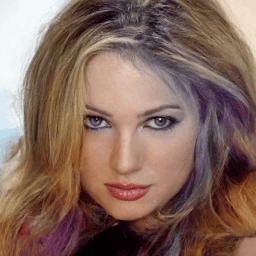} \\
            \makecell{\scriptsize{Source}} &
            \makecell{\scriptsize{Input}} &
            \makecell{\scriptsize{Inversion} \\ \scriptsize{w/o CBC}} &
            \makecell{\scriptsize{Inversion} \\ \scriptsize{w/ CBC}} &
            \makecell{\scriptsize{CODE} \\ \scriptsize{w/o} CBC} &
            \makecell{\scriptsize{CODE}} &
            \makecell{\scriptsize{SDEdit}} \\
    
            \end{tabular}
        \subcaption{Ablation Study.}
        \label{fig:ablation_comparison}
        \vspace{10pt}
    \end{subfigure}
    \hfill
    \begin{subfigure}{\columnwidth}
            \centering
        \begin{tikzpicture}

        \node[anchor=south west, inner sep=0] at (0,0) {
        \begin{tabular}{@{}c@{}c@{}c@{}c@{}c@{}c@{}c@{}}
            \includegraphics[width=0.13\linewidth]{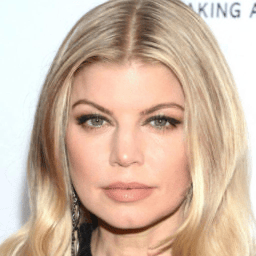}
            \includegraphics[width=0.13\linewidth]{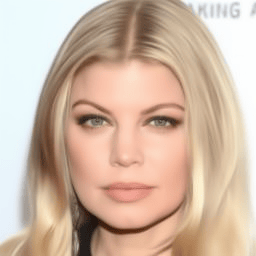}
            \includegraphics[width=0.13\linewidth]{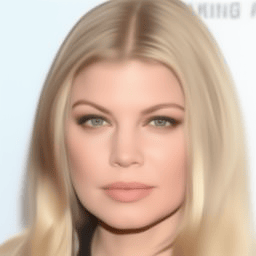}
            \includegraphics[width=0.13\linewidth]{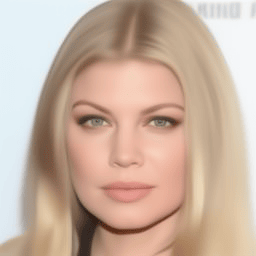}
            \includegraphics[width=0.13\linewidth]{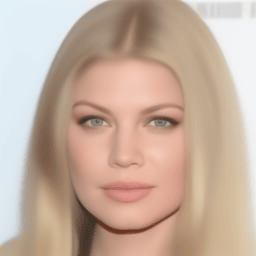}
            \includegraphics[width=0.13\linewidth]{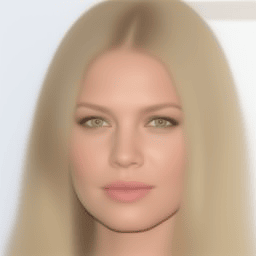}
            \includegraphics[width=0.13\linewidth]{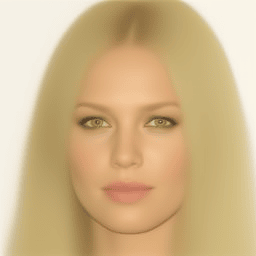} \\
            \includegraphics[width=0.13\linewidth]{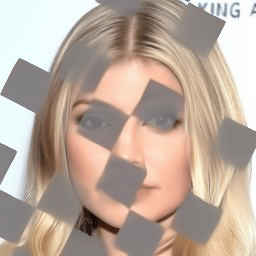}
            \includegraphics[width=0.13\linewidth]{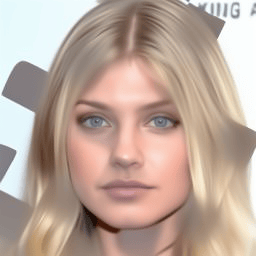}
            \includegraphics[width=0.13\linewidth]{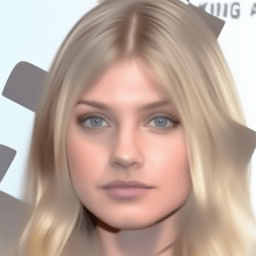}
            \includegraphics[width=0.13\linewidth]{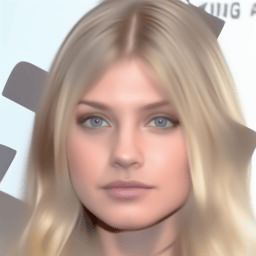}
            \includegraphics[width=0.13\linewidth]{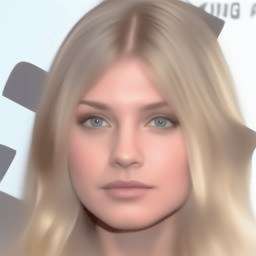}
            \includegraphics[width=0.13\linewidth]{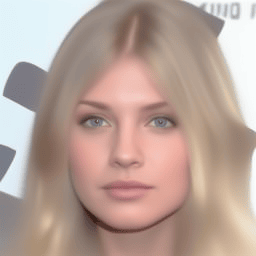}
            \includegraphics[width=0.13\linewidth]{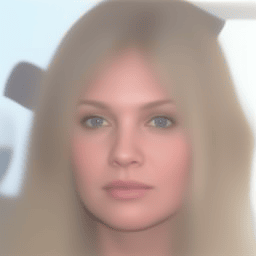} \\
            \includegraphics[width=0.13\linewidth]{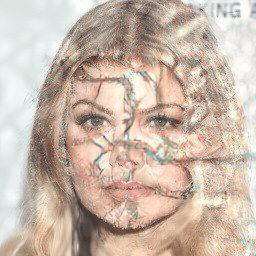}
            \includegraphics[width=0.13\linewidth]{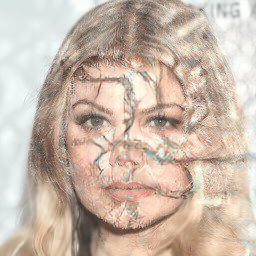}
            \includegraphics[width=0.13\linewidth]{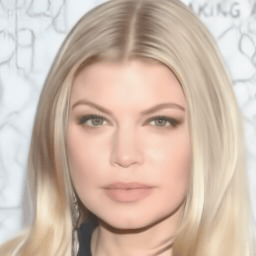}
            \includegraphics[width=0.13\linewidth]{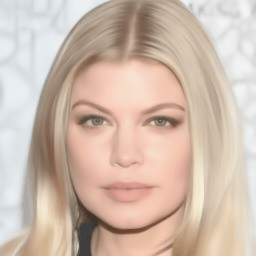}
            \includegraphics[width=0.13\linewidth]{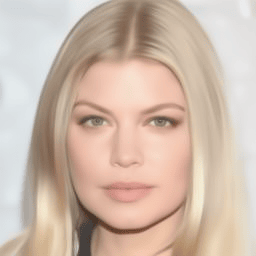}
            \includegraphics[width=0.13\linewidth]{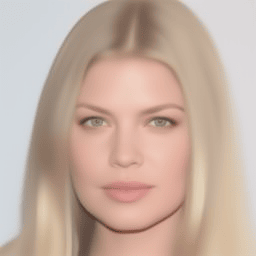}
            \includegraphics[width=0.13\linewidth]{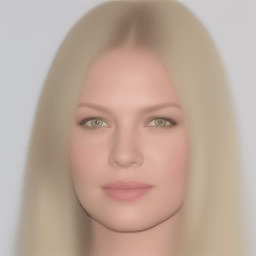} \\
            \includegraphics[width=0.13\linewidth]{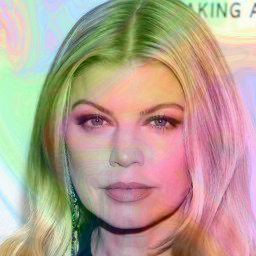}
            \includegraphics[width=0.13\linewidth]{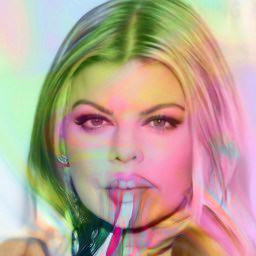}
            \includegraphics[width=0.13\linewidth]{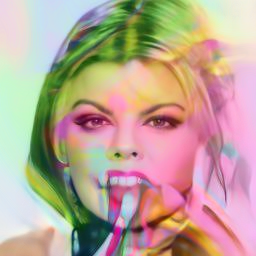}
            \includegraphics[width=0.13\linewidth]{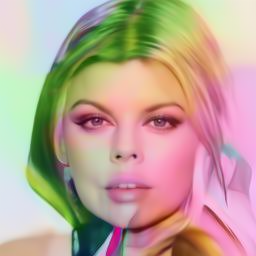}
            \includegraphics[width=0.13\linewidth]{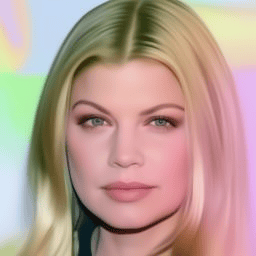}
            \includegraphics[width=0.13\linewidth]{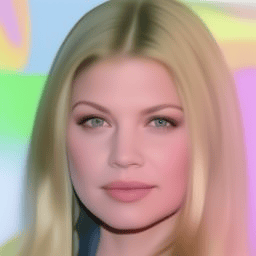}
            \includegraphics[width=0.13\linewidth]{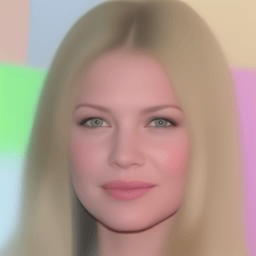} \\              
        \end{tabular}
        };

        \draw[->,line width=1.5pt] (0,0.1) -- (8.4,0.1) node[right] {};
        \node at (4.1,-0.32) {$\eta$};
        \end{tikzpicture}
    \subcaption{DDIM inversion using CBC with different values for confidence parameter $\eta$.}
    \label{fig:ablation_ddim}
    \end{subfigure}
    \caption{(a) Ablation Study (b) DDIM inversion using CBC with different values for the confidence parameter $\eta$.}   
    \label{fig:ablation}
\end{figure}

\paragraph{Impact of Confidence-Based Clipping}
In this section, we study the impact of the confidence parameter $\eta$ in CBC. As we reduce $\eta$, the interval shrinks, keeping only the most likely pixel values. We propose to encode an image using DDIM with different values of $\eta$ and decode it back to see the result. We study this in the case of in-distribution images and of corrupted images. Results can be seen in Figure \ref{fig:ablation_ddim}. A smaller $\eta$ results in a loss of fine-grained details, a shift of the average tone, and the removal of unlikely pixels such as masked pixels.
When applied to corrupted samples, CBC proves efficient in removing part of the noisy artifact while keeping most of the image structure. 
Interestingly, CBC also stabilizes the DDIM inversion, which might sometimes be inconsistent. This allows for fewer steps in the encoding-decoding procedure and speeds up the whole editing process.

\paragraph{Ablation Study}
We propose to study the impact of each block in CODE while keeping SDEdit for comparison. Qualitative results are visible in Figure \ref{fig:ablation_comparison}. While both our editing method alone (w/o CBC) and SDEdit excel at adding extra fine-grained details, they fail at handling unknown masks or color shits efficiently. On the contrary, CBC basically fails at adding extra details but successfully recovers certain color shifts or masked areas. As a result, combining both into CODE leads to powerful synergies.
Quantitative results can be seen in Table \ref{tab:ablation_metrics}.

\begin{table}[h]   
    \scriptsize
    \centering
    \begin{tabularx}{\columnwidth}{@{}l@{\hspace{8pt}}X@{\hspace{5pt}}X@{}X@{}X@{}}
     & LPIPS-Source $\downarrow$ &  SSIM-Input $\uparrow$ & PSNR-Input $\uparrow$ & FID $\downarrow$ \\
    \toprule[1pt]
    \textit{Inputs} & 0.48 (0.35) & - & - & 143.5 (96.3) \\
    \midrule[0.25pt]
    \textit{DDIM} & 0.52 (0.32) & 0.89 (0.08) & 30.7 (5.3) & 152.2 (88.4) \\
    \textit{DDIM w/ CBC} & 0.43 (0.19) & 0.73 (0.17) & 23.5 (5.3) & 90.1 (49.5) \\
    \midrule[0.25pt]
    \textit{\textbf{CODE w/o CBC}} & 0.31 (0.11) & 0.48 (0.22) & 19.4 (4.5) & 34.75 (14.6) \\
    \textit{\textbf{CODE w/ CBC}} & \textbf{0.30 (0.12)} & \textbf{0.49 (0.22)} & \textbf{19.6 (4.7)} & \textbf{30.65 (16.2)} \\
    \bottomrule[1pt]
    \end{tabularx}
    \caption{Confidence-Based Clipping ablation, metrics are averaged across all corruptions.}
    \label{tab:ablation_metrics}
\end{table}

\subsubsection{Discussion.}
While CODE offers enhanced versatility and control in editing, it introduces greater complexity compared to SDEdit. SDEdit's tuning is straightforward, with binary success or failure outcomes, whereas CODE's dual hyperparameter framework requires a more extensive grid search, increasing the search complexity quadratically. However, this added complexity enables CODE to achieve better results across a wider range of scenarios.

\section{Conclusion}
We introduce Confident Ordinary Differential Editing, a novel approach for guided image editing and synthesis that handles OoD inputs and balances realism and fidelity. Our method eliminates the need for retraining, finetuning, data augmentation, or paired data, and it integrates seamlessly with any pre-trained Diffusion Model. CODE excels in addressing a wide array of corruptions, outperforming existing methods that often rely on handcrafted features. As an evolution of SDEdit, our approach provides enhanced control, variety, and capabilities in editing by disentangling the original method and introducing additional hyperparameters. These new parameters not only offer deeper insights into the functioning of Diffusion Models' latent spaces but also enable more diverse editing strategies.
Furthermore, we introduce a Confidence-Based Clipping method that synergizes effectively with our editing technique, allowing the disregard of unlikely pixels or areas in a completely agnostic manner. 
Finally, our extensive study of the different components at play offers a greater understanding of the underlying mechanics of diffusion models, enriching the field’s knowledge base. 
Our findings reveal that CODE surpasses SDEdit in versatility and quality while maintaining its strengths across various tasks, including stroke-based editing. 
We hope our work inspires further innovations in this domain, akin to the transformative impact of GAN inversion. Looking ahead, we see potential in automating the editing and hyperparameter search processes and exploring synergies with text-to-image synthesis.

\clearpage

\appendix
\onecolumn
\section{Comparison to other methods}
\label{app:method-comparison}

\begin{sideways}
\begin{minipage}{1.375\linewidth}
    \centering
    \captionof{table}{Comprehensive comparison of the requirements and assumptions of various state-of-the-art image restoration and editing methods. Most existing methods assume at least some knowledge of the corruption (e.g., the Gaussian blurring kernel operator of a deblurring task), the linearity of the degradation operator, some task knowledge (e.g. if it's deblurring or inpainting that needs to be performed and where's the inpainting mask), or they require training with paired data and/or domain-specific features. The only exceptions are CODE (ours) and SDEdit \cite{meng2021sdedit}, which is our benchmark. \\ *: restorer model(s) only. **: experiments conducted with task-specific parameters initialization}
    \label{tab:method-comparison}
    \renewcommand{\arraystretch}{1.35}
    \begin{tabular}{lccccccp{3.5cm}}
    \toprule[1.5pt]
    & & & \multicolumn{2}{c}{\textbf{Corruption}} \\ \cmidrule(lr){4-5}
    Method & Paired data & Training & Known & Linear & Domain-specific & Task-specific & Other notes \\
    \midrule[1pt]
    CODE (ours) & - & - & - & - & - & - & - \\
    SDEdit \cite{meng2021sdedit} & - & - & - & - & - & - & - \\
    \midrule
    GDP \cite{fei2023generative} & - & - & \CheckmarkBold (for some tasks) & - & - &  \CheckmarkBold & Needs to estimate a task-specific degradation model, which is hard to do for complex corruptions and unknown tasks \\
    PGDiff \cite{yang2023pgdiff} & \CheckmarkBold* & \CheckmarkBold* & \CheckmarkBold (for some tasks) & - & - &  \CheckmarkBold & Handcrafted task-specific guidance and masks \\
    DDRM \cite{kawar2022denoising} & - & - & \CheckmarkBold & \CheckmarkBold & - & \CheckmarkBold & Needs SVD decomposition of linear degradation operator \\
    DDNM \cite{wang2022zeroshot} & - & - & \CheckmarkBold & \CheckmarkBold & - & \CheckmarkBold & Hand-designed linear degradation operator A and pseudoinverse A\textsuperscript{\textdagger} \\
    GibbsDDRM \cite{murata2023gibbsddrm} & - & - & - & \CheckmarkBold & - & \CheckmarkBold** & - \\
    Codeformer \cite{zhou2022towards} & \CheckmarkBold & \CheckmarkBold & - & - & \CheckmarkBold (faces) & - & - \\
    GFPGAN \cite{gfpgan} & \CheckmarkBold & \CheckmarkBold & - & - & \CheckmarkBold (faces) & - & - \\
    \makecell{Robust Style-GAN \\ inversion \cite{10205099}} & - & - & \CheckmarkBold (for some tasks) & - & - & \CheckmarkBold & - \\
    DiffBIR \cite{lin2023diffbir} & \CheckmarkBold & \CheckmarkBold & - & - & - & \CheckmarkBold & - \\
    DPS \cite{chung2023diffusion} & - & - & \CheckmarkBold & - & - & \CheckmarkBold & - \\
    DiffPIR \cite{zhu2023denoising} & - & - & \CheckmarkBold & - & - & \CheckmarkBold & - \\
    DDB \cite{chung2023direct} & \CheckmarkBold & \CheckmarkBold & \CheckmarkBold & - & - & - & - \\
    I$^{2}$SB \cite{liu2023i2sb} & \CheckmarkBold & \CheckmarkBold & \CheckmarkBold & - & - & - & - \\
    \bottomrule[1.5pt]
    \end{tabular}
\end{minipage}
\end{sideways}
\twocolumn

\begin{figure*}[h!]
    \centering
    \setlength{\tabcolsep}{0pt}
    \includegraphics[width=0.75\linewidth]{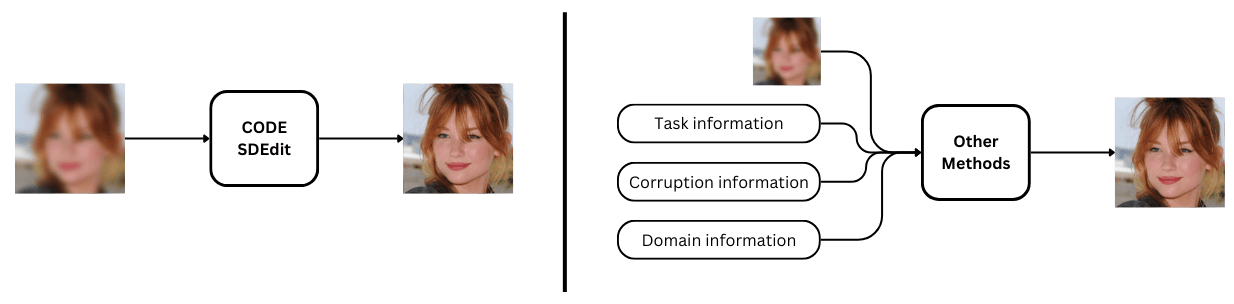}
    \caption{Comparison of method requirements. On the left, CODE and SDEdit directly enhance a degraded image to high quality. On the right, other methods necessitate additional information such as task, corruption, and domain information to enhance the image restoration process}
    \label{fig:method_comparison}
\end{figure*}

\begin{table}[htbp]
    \scriptsize
    \centering
    \begin{tabularx}{\columnwidth}{XXXXXXXX}
        & \multicolumn{3}{c}{CODE (fog)} & \multicolumn{3}{c}{CODE (blur)} & SDEdit  \\ 
        \hline
        & DDIM & Langevin & Total & DDIM & Langevin & Total \\
        \cmidrule(lr){2-4} \cmidrule(lr){5-7}
        NFEs & 2x200 & 2x300 & \textbf{1000} & 2x40 & 1x200 & \textbf{280} & \textbf{400} \\
        Walltime & 2x15.4s ± 0.8s & 2x20.6s ± 1.6s & \textbf{72.0s} & 2x3.1s ± 0.2s & 9.7s ± 0.1s & \textbf{15.9s} & \textbf{25.3s} \\
    \end{tabularx}
     \caption{Comparison of the computational requirements of CODE (ours) and SDEdit in terms of Number of Function Evaluations (NFEs) and wall-clock time. We show CODE runtime requirements are broken down into all the phases, DDIM inversion + DDIM and Langevin sampling for simple corruptions (e.g., blur) and complex corruptions (e.g., fog). Benchmark conducted on an NVIDIA GeForce RTX 3090. Reported average wall-time ± std across 100 trials}
    \label{tab:my_label}
\end{table}

\section{Proof of Proposition 1}
\label{app:th}
Let $\Phi$ be the cumulative distribution function of $\mathcal{N}(0, \mathcal{I})$ and let $x_0 \in [-1, 1]$. Let $\alpha_t \in [0,1] \text{, } \forall t \in [0,1]$. Assume that $x_t \sim \mathcal{N}(\sqrt{\alpha_t} \cdot \alpha_0, \sqrt{1-\alpha_t} \cdot \mathcal{I})$, then for all $\eta$:\\
\resizebox{0.8\columnwidth}{!}{\parbox{\linewidth}{
\begin{equation*}
    \mathcal{P}(x_t \in [-\sqrt{\alpha_t} - \eta \cdot \sqrt{1-\alpha_t},\sqrt{\alpha_t} + \eta \cdot \sqrt{1-\alpha_t}]) \geq \Phi(\eta) - \Phi(-\eta))
\end{equation*}
}}\\
In particular, we have for $\eta = 2$,\\
\resizebox{0.8\columnwidth}{!}{\parbox{\linewidth}{
\begin{equation*}
    \mathcal{P}(x_t \in [-\sqrt{\alpha_t} - 2 \cdot \sqrt{1-\alpha_t},\sqrt{\alpha_t} + 2 \cdot \sqrt{1-\alpha_t}]) \geq 0.95
\end{equation*}
}}\\
and for $\eta = 3$,\\
\resizebox{0.8\columnwidth}{!}{\parbox{\linewidth}{
\begin{equation*}
    \mathcal{P}(x_t \in [-\sqrt{\alpha_t} - 3 \cdot \sqrt{1-\alpha_t},\sqrt{\alpha_t} + 3 \cdot \sqrt{1-\alpha_t}]) \geq 0.99
\end{equation*}
}}


\begin{proof}
\label{proof:1}
Let $x_t \sim \mathcal{N}(x_t, \sqrt{\alpha_t} \cdot x_0, \sqrt{1-\alpha_t} \cdot \mathbf{I})$, then it comes that for all $\eta$ : \\
\resizebox{0.8\columnwidth}{!}{\parbox{\linewidth}{
\begin{equation*}
    P(x_t \in [\sqrt{\alpha_t} \cdot x_0 - \eta \cdot \sqrt{1-\alpha_t}, \sqrt{\alpha_t} \cdot x_0 + \eta \cdot \sqrt{1-\alpha_t}]) = \Phi(\eta) - \Phi(-\eta).
\end{equation*}
}}\\
Let us denote $\mathcal{A}_{x_0, \eta}$ such that:\\
\resizebox{0.8\columnwidth}{!}{\parbox{\linewidth}{
\begin{equation*}
    \mathcal{A}_{x_0, \eta} = [\sqrt{\alpha_t} \cdot x_0 - \eta \cdot \sqrt{1-\alpha_t}, \sqrt{\alpha_t} \cdot x_0 + \eta \cdot \sqrt{1-\alpha_t}].
\end{equation*}
}}\\
However, we know that $x_0 \in [-1,1]$, hence:\\
\resizebox{0.8\columnwidth}{!}{\parbox{\linewidth}{
\begin{equation*}
     \mathcal{A}_{x_0, \eta} \subseteq [-\sqrt{\alpha_t} - \eta \cdot \sqrt{1-\alpha_t}, \sqrt{\alpha_t} + \eta \cdot \sqrt{1-\alpha_t}].
\end{equation*}
}}\\
Let us denote $\mathcal{A}_{\eta}$ such that:\\
\resizebox{0.8\columnwidth}{!}{\parbox{\linewidth}{
\begin{equation*}
    \mathcal{A}_{\eta} = [-\sqrt{\alpha_t} - \eta \cdot \sqrt{1-\alpha_t}, \sqrt{\alpha_t} + \eta \cdot \sqrt{1-\alpha_t}].
\end{equation*}
}}\\
There it comes that:\\
\resizebox{0.8\columnwidth}{!}{\parbox{\linewidth}{
\begin{equation*}
    P(x_t \in \mathcal{A}_{\eta}) \geq P(x_t \in \mathcal{A}_{x_0, \eta}).
\end{equation*}
}}\\
Hence our final result:\\
\resizebox{0.8\columnwidth}{!}{\parbox{\linewidth}{
\begin{equation*}
    P(x_t \in \mathcal{A}_{\eta}) \geq \Phi(\eta) - \Phi(-\eta).
\end{equation*}
}}
\end{proof}


\section{Denoising Score Matching}
\label{sec:diffusion-detail}

In the denoising score matching framework, the data $\mathbf{x}$ is first perturbed by a Gaussian noise. If we note the noised distribution $\mathbf{\Tilde{x}}$ then:
$q_{\sigma}(\mathbf{\Tilde{x}}|\mathbf{x}) = \mathcal{N}(\mathbf{\Tilde{x}},\mathbf{x}, \sigma^2\mathbf{I})$ then $\nabla_{\mathbf{\Tilde{x}}} \log q_{\sigma}(\mathbf{\Tilde{x}}|\mathbf{x}) = -\frac{\mathbf{\Tilde{x}}-\mathbf{x}}{\sigma^2}$ and the denoising score matching objective becomes:

\resizebox{0.8\columnwidth}{!}{\parbox{\linewidth}{
\begin{equation*}    
\mathbb{E}_{p(\mathbf{x})}\mathbb{E}_{\mathbf{\Tilde{x}} \sim \mathcal{N}(\mathbf{x}, \sigma^2\mathbf{I})} \norm{s_{\theta}(\mathbf{\Tilde{x}}, \sigma) - \frac{x-\Tilde{x}}{\sigma^2}}.
\end{equation*}
}}

\section{Detailed Algorithms}
\label{App:alg}

\begin{algorithm}[h!]
\caption{ODE editing}
\label{alg:odedit_1}
\begin{algorithmic}
\REQUIRE $N$ \textit{(Langevin iterations)}, $\epsilon$ \textit{(step-size)}, $x_0$ \textit{(Observation)}, $L$ \textit{(L-th latent-space)}.
\STATE $x_{L, 0} = ODE\_SOLVER_{\textit{forward}_{0 \rightarrow L}}(x_0)$
    \FOR{$k=0$ \TO $N-1$}
        
        \STATE $x_{L,k+1} = x_{L,k} - \epsilon \cdot s_{\theta}(x_{L,k}, L) \ + \sqrt{2 \epsilon } \cdot \eta$ , where $\eta \sim \mathcal{N}(0, \mathbf{I})$ 

    \ENDFOR
     \STATE $\tilde x_{0} = ODE\_SOLVER_{\textit{backward}_{L \rightarrow 0}}(x_{L,N})$
\end{algorithmic}
\end{algorithm}

\begin{algorithm}[h!]
\caption{CODE - Annealed Multi-Latent with CBC}
\label{alg:odedit_2}
\begin{algorithmic}
\REQUIRE $N$ \textit{(Langevin iterations)}, $\epsilon$ \textit{(step-size)}, $\alpha$ \textit{(Annealing coefficient)}, $K$ \textit{Number of annealing steps}, $x_0$ \textit{(Observation)} , $L_1, ..., L_j$ \textit{(List of latent-spaces optimized)}, $\eta$ \textit{(size of the confidence interval.)}

\STATE $L_0 = 0$
\STATE $x_{L_j, 0} = ODE\_SOLVER_{\textit{forward}_{0 \rightarrow L_j}}(Clip_{CBC_{\eta}}(x_0))$
\FOR {$l=j$ \TO $1$}
    \STATE $\epsilon_{0} = \epsilon$
    
    \FOR {$k=0$ \TO $K-1$}
        \FOR{$n=0$ \TO $N-1$}
            \STATE $x_{L_l,n+1} = x_{L_l,n} - \epsilon_{k} \cdot s_{\theta, L_l}(x_{L_l,n}) \ + \sqrt{2 \epsilon_{k} } \cdot \eta$ , where $\eta \sim \mathcal{N}(0, \mathbf{I})$ 
        \ENDFOR 
    \STATE $x_{L_l, 0} = x_{L_l,N}$
    \STATE $\epsilon_{k+1} = \epsilon_{k} \cdot \alpha$
    \ENDFOR
    \STATE $x_{L_{l-1},0} = ODE\_SOLVER_{\textit{backward}_{L_l \rightarrow L_{l-1}}}(x_{L_l,N})$
\ENDFOR
\RETURN $x_{L_0,0}$
\end{algorithmic}
\end{algorithm}

\begin{algorithm}[h!]
\newcommand{\rvx}{\mathbf{x}}
\newcommand{\rvz}{\mathbf{z}}
\newcommand{\vs}{\bm{s}}
\newcommand{\vtheta}{\bm{\theta}}
\caption{SDEdit (VP-SDE formulation)}
\begin{algorithmic}

\REQUIRE $\mathbf{x}^{(g)}$ (guide), $t_0$ (SDE hyper-parameter), $N$ (total denoising steps), $K$ (total repeats)

\STATE $\Delta t \gets \frac{t_0}{N}$
\STATE $\alpha(t_0) \gets \prod_{n=1}^{N} (1-\beta(\frac{nt_0}{N})\Delta t)$

\FOR {$k \gets 1$ \TO $K$}
    \STATE $\mathbf{x} \sim \mathcal{N}(\bm{0}, \bm{I})$
    \STATE $\rvx \gets \sqrt{\alpha(t_0)}\rvx + \sqrt{1-\alpha(t_0)}\rvz$
    
    \FOR {$n \gets N$ \TO $1$}
        \STATE $t \gets t_0 \frac{n}{N}$
        \STATE $\rvz \sim \mathcal{N}(\bm{0}, \bm{I})$
        \STATE $\rvx \gets \frac{1}{\sqrt{1 - \beta(t)\Delta t}}\left(\rvx + \beta(t)\Delta t \vs_{\vtheta}(\rvx, t)\right) + \sqrt{\beta(t)\Delta t} \rvz$
    \ENDFOR
\ENDFOR
\STATE \textbf{return} $\rvx$
\end{algorithmic}
\end{algorithm}


\section{Additional Details}

\subsection{Metrics}
In the initial phase of filtering, we processed a multitude of samples for each corrupted image. This yielded approximately 1.6 million samples across 500 unique faces, 47 distinct types of corruption, 4 algorithms (CODE, CODE w/o CBC, SDEdit, CBC), 2 or 3 latents (3 in case of SDEdit, 2 in case of all CODE variations), 5 epsilons, step sizes, for all CODE variations, and 2 to 4 generated images for stochasticity. We automatically selected the top 4 samples for each based on the psnr to source metric, mimicking a human selection. Consequently, given the number of cases to benchmark, the diversity is 500 unique faces.

\subsection{Hyperparameters}
To generate the different samples we use the following sets of hyperparameters. For SDEdit we generate samples using 300 steps, 500 steps and 700 steps. For CODE, we first use a $eta$ (CBC parameter) of 1.7 to encode the corrupted images. Then we generate samples using two sets of latent, either 200 and 40, either 40 only. The former being for deep latent correction and the latter for shallow corrections. We use 200 updates langevin steps in each selected latent space. We generate samples using five different step size ranging from $1e-5$ to $1e-3$. we use annealing for the step size every 40 langevin update, with a annealing constant of 0.8 . All experiments have been conducted on Nvidia GPUs (A100, V100, RTX3090). Please refer to the Github for packages requirements.

\section{Extra Results - L$_2$ xguidance and LSUN}
\label{App:ablation}
\begin{figure}[htbp]
    \centering
    \setlength{\tabcolsep}{0pt}
    \renewcommand{\arraystretch}{0} 
    \begin{tabular}{c@{\hspace{10pt}}c@{}c@{}c@{}c}
        \includegraphics[width=0.16\linewidth]{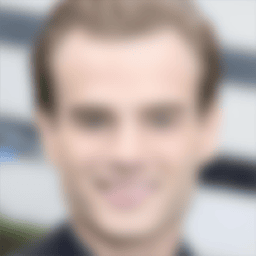} &
        \includegraphics[width=0.16\linewidth]{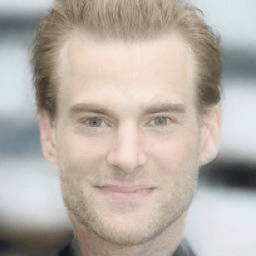} &
        \includegraphics[width=0.16\linewidth]{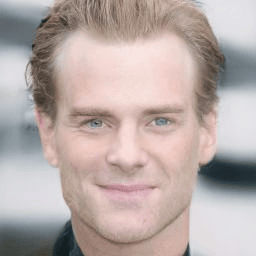} &
        \includegraphics[width=0.16\linewidth]{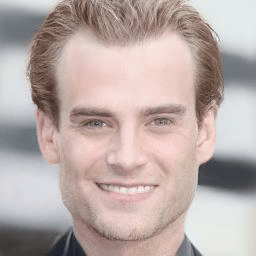} \\
        \includegraphics[width=0.16\linewidth]{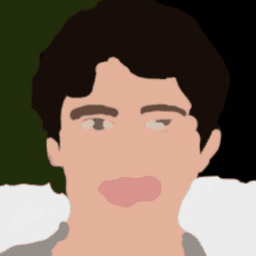} & 
        \includegraphics[width=0.16\linewidth]{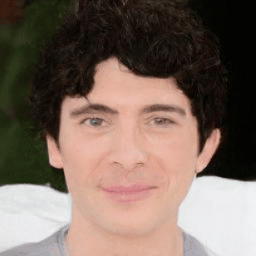} &
        \includegraphics[width=0.16\linewidth]{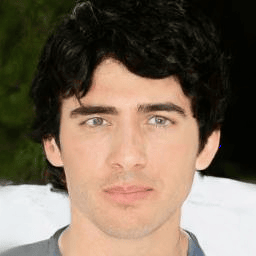} &
        \includegraphics[width=0.16\linewidth]{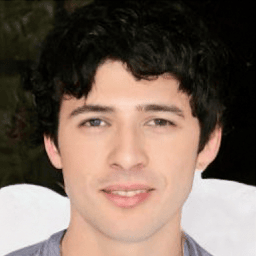} \\
        \includegraphics[width=0.16\linewidth]{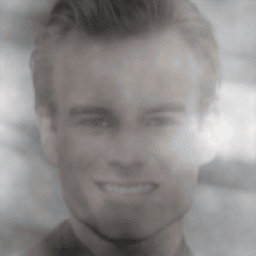} & 
        \includegraphics[width=0.16\linewidth]{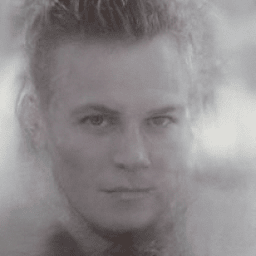} &
        \includegraphics[width=0.16\linewidth]{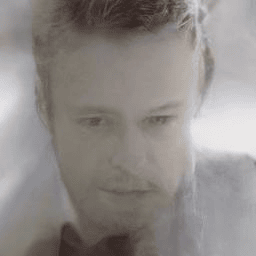} &
        \includegraphics[width=0.16\linewidth]{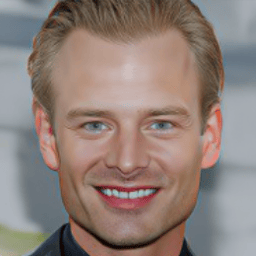} \\
        \multicolumn{1}{c@{\hspace{10pt}}}{\makecell{\scriptsize{Source}}} &
        \multicolumn{1}{c}{\makecell{\scriptsize{L2 guidance}}} &
        \multicolumn{1}{c}{\makecell{\scriptsize{SDEdit}}} &
        \multicolumn{1}{c}{\makecell{\scriptsize{CODE}}} &
        \end{tabular}
    \caption{CODE goes beyond L$_2$ guidance in the visual domain, allowing it to be applied in more diverse and challenging scenarios with severe corruption. L$_2$ guidance and SDEdit only work on corruptions that are more related to visual details.}
    \label{fig:man_L2}
\end{figure}

\begin{figure}[h]
    \centering
        \includegraphics[width=\linewidth]{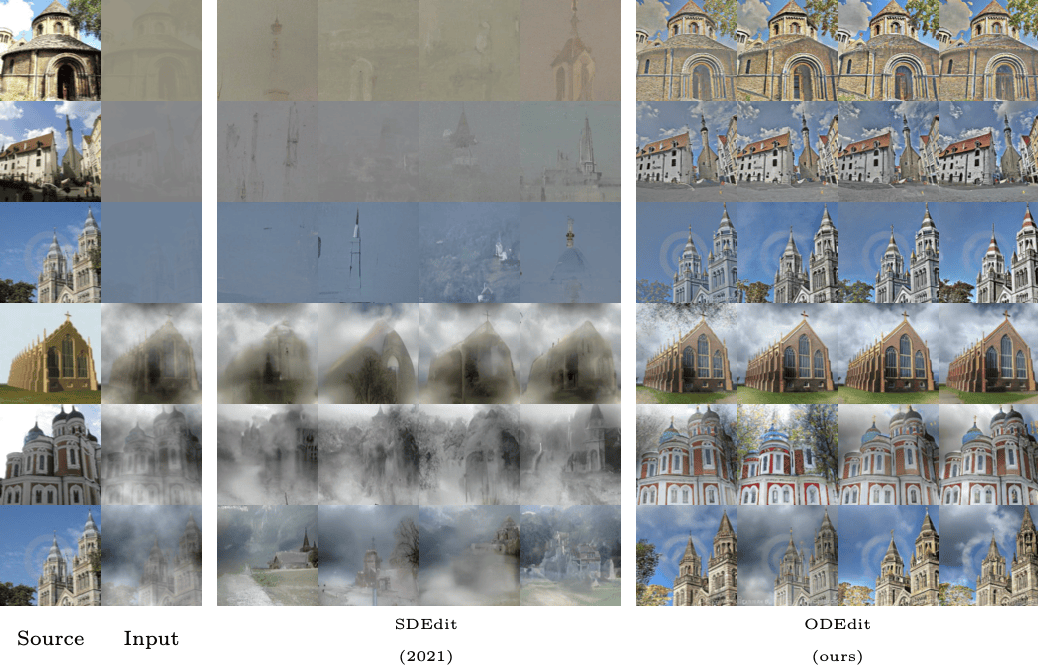} 
    \caption{Qualitative results on LSUN}
\end{figure}
\section{Visuals Results}
\label{App:visuals}
\newpage
\clearpage


\subsection{Caustic Noise}
\begin{table}[htbp]
    \centering
    \scriptsize
\begin{tabularx}{\columnwidth}{@{}ccccc@{}}
    \toprule[1.5pt]
     & LPIPS-Source $\downarrow$ &  SSIM-Input $\uparrow$ & PSNR-Input $\uparrow$ & FID $\downarrow$ \\
    \midrule[1pt]
    \makecell{\textit{Inputs}} & 0.10 & - & - & 35.55 \\
    \midrule[1pt]
    \makecell{\textit{DDIM}\\w/o CBC} & 0.22 & 0.85 & 25.31 & 65.89 \\
    \midrule[1pt]
    \makecell{\textit{DDIM}\\w/ CBC} & 0.26 & 0.80 & 25.86 & 55.11 \\
    \midrule[1pt]
    \makecell{\textbf{CODE}\\\textbf{w/o CBC}} & 0.17 & 0.71 & 24.37 & 19.10 \\
    \midrule[1pt]
    \makecell{\textbf{CODE}\\\textbf{w/ CBC}} & 0.22 & 0.66 & 22.77 & 18.27 \\
    \midrule[1pt]
    \makecell{\textit{SDEdit}} & 0.23 & 0.58 & 20.86 & 27.77 \\
    \bottomrule[1.5pt]
    \end{tabularx}
    \caption{Quantitative results on Caustic Noise}
\end{table}
\begin{figure}[h!]
    \centering
    \setlength{\tabcolsep}{0pt}
    \renewcommand{\arraystretch}{0} 
    \begin{tabular}{c@{}c@{}c}
        \includegraphics[width=0.3\linewidth]{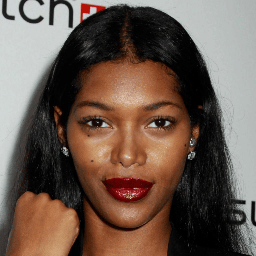} &
        \includegraphics[width=0.3\linewidth]{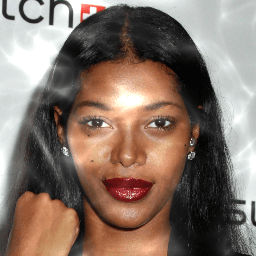} &
        \includegraphics[width=0.3\linewidth]{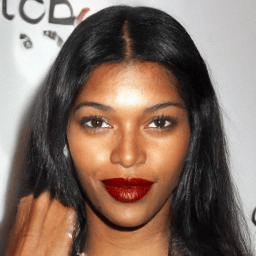} \\
        \includegraphics[width=0.3\linewidth]{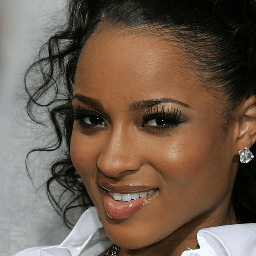} &
        \includegraphics[width=0.3\linewidth]{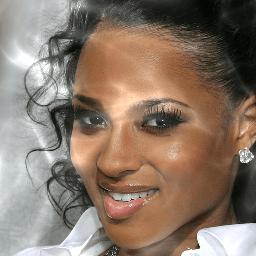} &
        \includegraphics[width=0.3\linewidth]{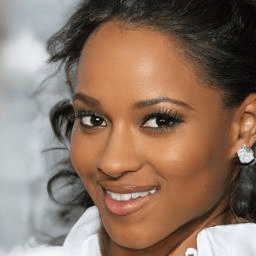} \\
        \includegraphics[width=0.3\linewidth]{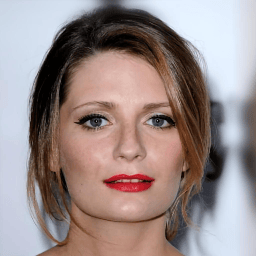} &
        \includegraphics[width=0.3\linewidth]{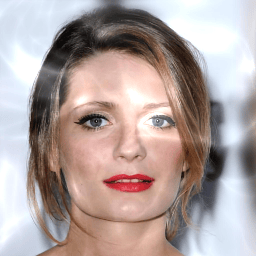} &
        \includegraphics[width=0.3\linewidth]{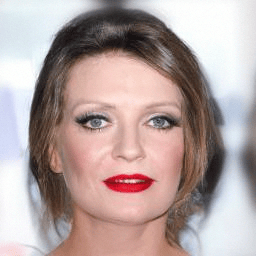} \\
        \includegraphics[width=0.3\linewidth]{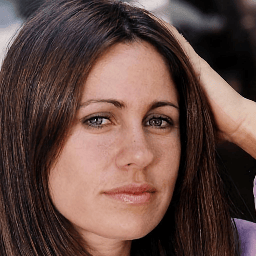} &
        \includegraphics[width=0.3\linewidth]{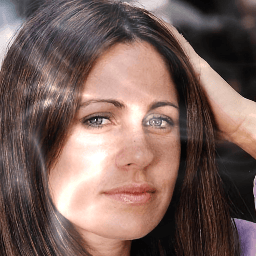} &
        \includegraphics[width=0.3\linewidth]{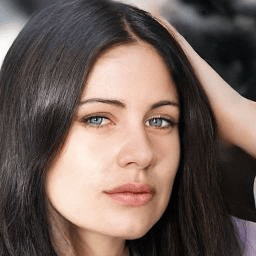} \\
        \includegraphics[width=0.3\linewidth]{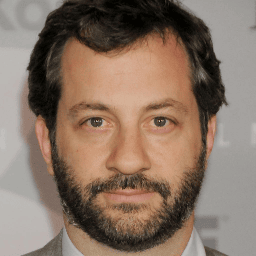} &
        \includegraphics[width=0.3\linewidth]{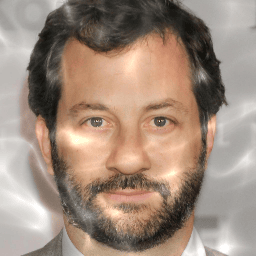} &
        \includegraphics[width=0.3\linewidth]{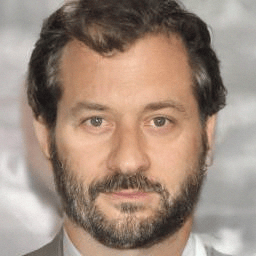} \\
        \multicolumn{1}{c}{\makecell{\scriptsize{Source}}} &
        {\makecell{\scriptsize{Input}}} &
        \multicolumn{1}{c}{\makecell{\scriptsize{\textbf{CODE (ours)}}}}  \\
        \end{tabular}
    \caption{Qualitative results of Caustic Noise}
    \label{fig:appendix-caustic}
\end{figure}
\newpage


\subsection{Caustic Refraction Noise}
\begin{table}[htbp]
    \centering
    \scriptsize
    \begin{tabularx}{\columnwidth}{@{}ccccc@{}}
    \toprule[1.5pt]
     & LPIPS-Source $\downarrow$ &  SSIM-Input $\uparrow$ & PSNR-Input $\uparrow$ & FID $\downarrow$ \\
    \midrule[1pt]
    \makecell{\textit{Inputs}} & - & - & - & - \\
    \midrule[1pt]
    \makecell{\textit{DDIM}\\w/o CBC} & 0.36 & 0.77 & 21.92 & 119.98 \\
    \midrule[1pt]
    \makecell{\textit{DDIM}\\w/ CBC} & 0.26 & 0.75 & 24.17 & 61.80 \\
    \midrule[1pt]
    \makecell{\textbf{CODE}\\\textbf{w/o CBC}} & 0.27 & 0.50 & 19.27 & 24.65 \\
    \midrule[1pt]
    \makecell{\textbf{CODE}\\\textbf{w/ CBC}} & 0.21 & 0.55 & 20.92 & 15.95 \\
    \midrule[1pt]
    \makecell{\textit{SDEdit}} & 0.20 & 0.54 & 20.56 & 26.46 \\
    \bottomrule[1.5pt]
    \end{tabularx}
    \caption{Quantitative results on Caustic Refraction Noise}
\end{table}

\begin{figure}[h!]
    \centering
    \setlength{\tabcolsep}{0pt}
    \renewcommand{\arraystretch}{0} 
    \begin{tabular}{c@{}c@{}c}
        \includegraphics[width=0.3\linewidth]{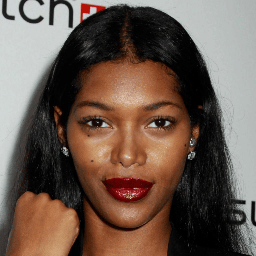} &
        \includegraphics[width=0.3\linewidth]{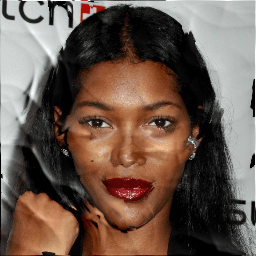} &
        \includegraphics[width=0.3\linewidth]{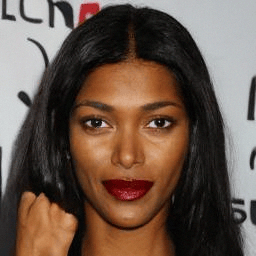} \\
        \includegraphics[width=0.3\linewidth]{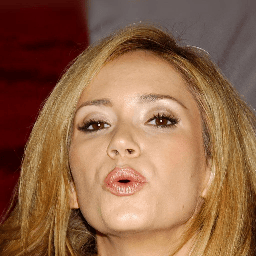} &
        \includegraphics[width=0.3\linewidth]{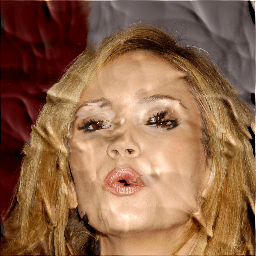} &
        \includegraphics[width=0.3\linewidth]{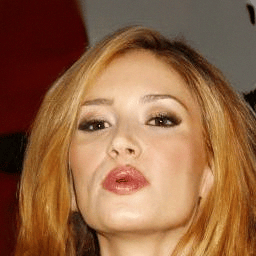} \\
        \includegraphics[width=0.3\linewidth]{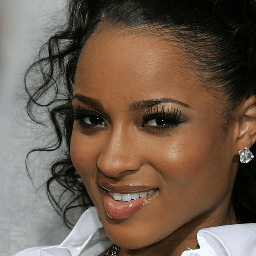} &
        \includegraphics[width=0.3\linewidth]{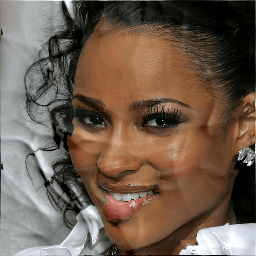} &
        \includegraphics[width=0.3\linewidth]{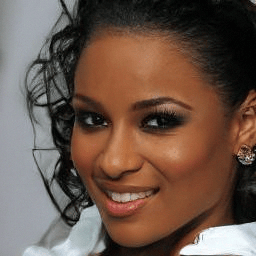} \\
        \includegraphics[width=0.3\linewidth]{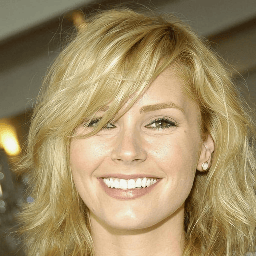} &
        \includegraphics[width=0.3\linewidth]{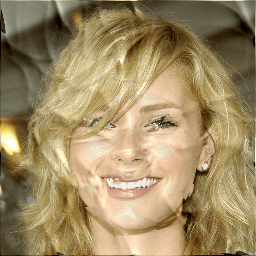} &
        \includegraphics[width=0.3\linewidth]{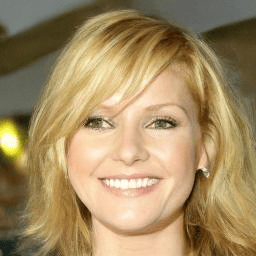} \\
        \includegraphics[width=0.3\linewidth]{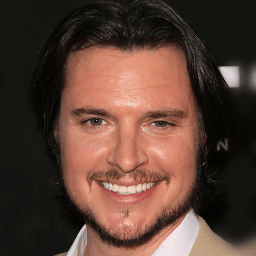} &
        \includegraphics[width=0.3\linewidth]{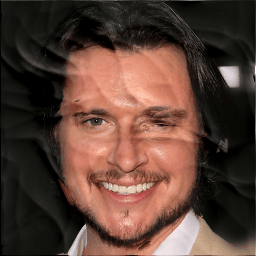} &
        \includegraphics[width=0.3\linewidth]{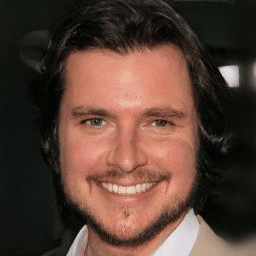} \\
        \multicolumn{1}{c}{\makecell{\scriptsize{Source}}} &
        {\makecell{\scriptsize{Input}}} &
        \multicolumn{1}{c}{\makecell{\scriptsize{\textbf{CODE (ours)}}}}  \\
        \end{tabular}
    \caption{Qualitative results of Caustic Refraction Noise}
    \label{fig:appendix-caustic-refraction}
\end{figure}
\newpage


\subsection{Brownish Noise}
\begin{table}[htbp]
    \centering
    \scriptsize
    \begin{tabularx}{\columnwidth}{@{}ccccc@{}}
    \toprule[1.5pt]
     & LPIPS-Source $\downarrow$ &  SSIM-Input $\uparrow$ & PSNR-Input $\uparrow$ & FID $\downarrow$ \\
    \midrule[1pt]
    \makecell{\textit{Inputs}} & 0.29 & - & - & 75.83 \\
    \midrule[1pt]
    \makecell{\textit{DDIM}\\w/o CBC} & 0.32 & 0.93 & 34.1 & 87.18 \\
    \midrule[1pt]
    \makecell{\textit{DDIM}\\w/ CBC} & 0.40 & 0.83 & 27.66 & 82.67 \\
    \midrule[1pt]
    \makecell{\textbf{CODE}\\\textbf{w/o CBC}} & 0.34 & 0.60 & 20.91 & 35.52 \\
    \midrule[1pt]
    \makecell{\textbf{CODE}\\\textbf{w/ CBC}} & 0.36 & 0.57 & 20.65 & 32.83 \\
    \midrule[1pt]
    \makecell{\textit{SDEdit}} & 0.37 & 0.53 & 19.80 & 43.78 \\
    \bottomrule[1.5pt]
    \end{tabularx}
    \caption{Quantitative results on Brownish Noise}
\end{table}
\begin{figure}[h!]
    \centering
    \setlength{\tabcolsep}{0pt}
    \renewcommand{\arraystretch}{0} 
    \begin{tabular}{c@{}c@{}c}

        \includegraphics[width=0.3\linewidth]{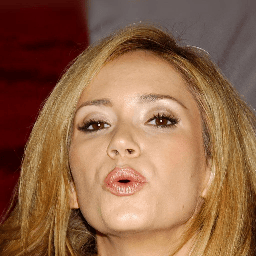} &
        \includegraphics[width=0.3\linewidth]{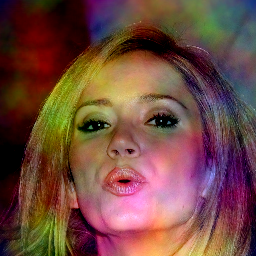} &
        \includegraphics[width=0.3\linewidth]{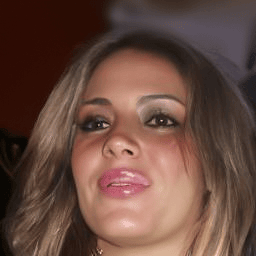} \\
        \includegraphics[width=0.3\linewidth]{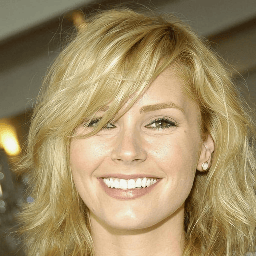} &
        \includegraphics[width=0.3\linewidth]{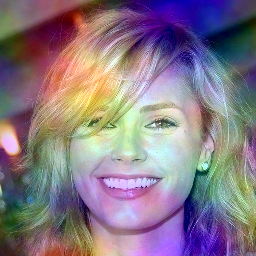} &
        \includegraphics[width=0.3\linewidth]{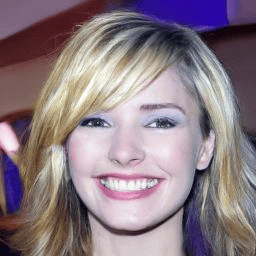} \\
        \includegraphics[width=0.3\linewidth]{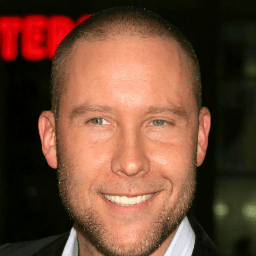} &
        \includegraphics[width=0.3\linewidth]{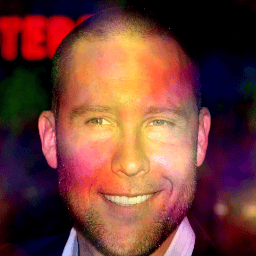} &
        \includegraphics[width=0.3\linewidth]{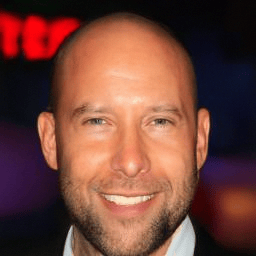} \\
        \includegraphics[width=0.3\linewidth]{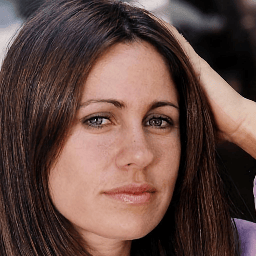} &
        \includegraphics[width=0.3\linewidth]{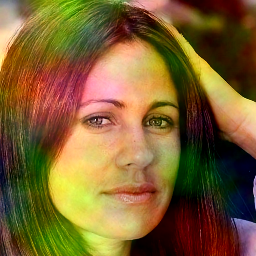} &
        \includegraphics[width=0.3\linewidth]{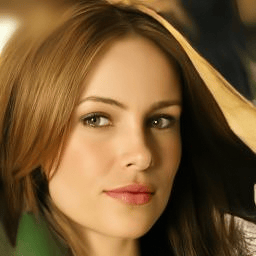} \\
        \includegraphics[width=0.3\linewidth]{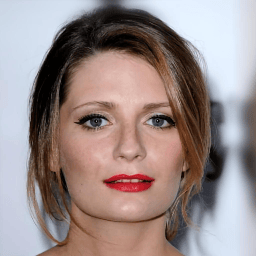} &
        \includegraphics[width=0.3\linewidth]{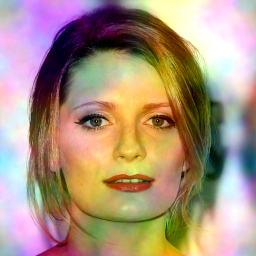} &
        \includegraphics[width=0.3\linewidth]{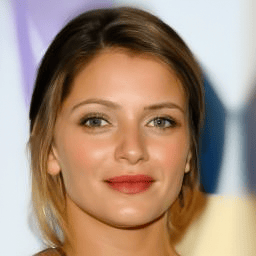} \\
        \multicolumn{1}{c}{\makecell{\scriptsize{Source}}} &
        {\makecell{\scriptsize{Input}}} &
        \multicolumn{1}{c}{\makecell{\scriptsize{\textbf{CODE (ours)}}}}  \\
        \end{tabular}
    \caption{Qualitative results of Brownish Noise}
    \label{fig:appendix-brownish}
\end{figure}
\newpage


\subsection{Concentric Sine Waves Noise}
\begin{table}[htbp]
    \centering
    \scriptsize
    \begin{tabularx}{\columnwidth}{@{}ccccc@{}}
    \toprule[1.5pt]
     & LPIPS-Source $\downarrow$ &  SSIM-Input $\uparrow$ & PSNR-Input $\uparrow$ & FID $\downarrow$ \\
    \midrule[1pt]
    \makecell{\textit{Inputs}} & 0.37 & - & - & 221.30 \\
    \midrule[1pt]
    \makecell{\textit{DDIM}\\w/o CBC} & 0.44 & 0.85 & 22.42 & 235.53 \\
    \midrule[1pt]
    \makecell{\textit{DDIM}\\w/ CBC} & 0.21 & 0.59 & 22.75 & 63.07 \\
    \midrule[1pt]
    \makecell{\textbf{CODE}\\\textbf{w/o CBC}} & 0.18 & 0.49 & 22.00 & 29.30 \\
    \midrule[1pt]
    \makecell{\textbf{CODE}\\\textbf{w/ CBC}} & 0.15 & 0.43 & 21.42 & 18.38 \\
    \midrule[1pt]
    \makecell{\textit{SDEdit}} & 0.17 & 0.35 & 20.00 & 25.00 \\
    \bottomrule[1.5pt]
    \end{tabularx}
    \caption{Quantitative results on Concentric Sine Waves Noise}
\end{table}
\begin{figure}[h!]
    \centering
    \setlength{\tabcolsep}{0pt}
    \renewcommand{\arraystretch}{0} 
    \begin{tabular}{c@{}c@{}c}
        \includegraphics[width=0.3\linewidth]{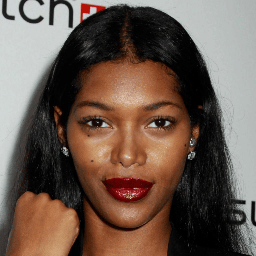} &
        \includegraphics[width=0.3\linewidth]{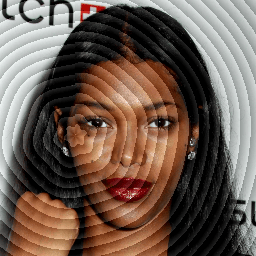} &
        \includegraphics[width=0.3\linewidth]{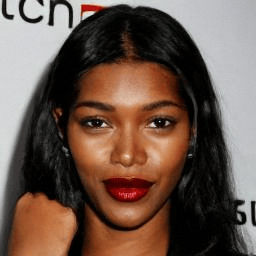} \\
        \includegraphics[width=0.3\linewidth]{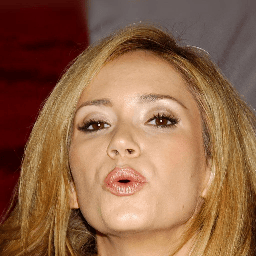} &
        \includegraphics[width=0.3\linewidth]{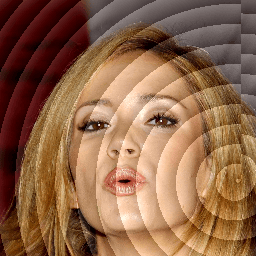} &
        \includegraphics[width=0.3\linewidth]{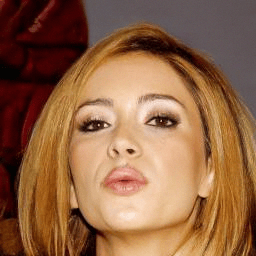} \\
        \includegraphics[width=0.3\linewidth]{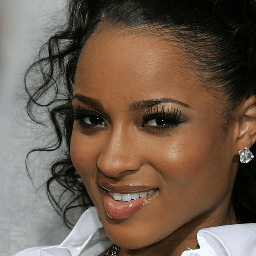} &
        \includegraphics[width=0.3\linewidth]{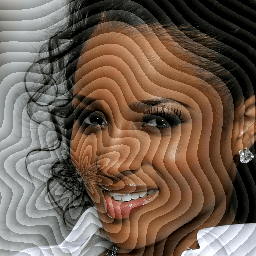} &
        \includegraphics[width=0.3\linewidth]{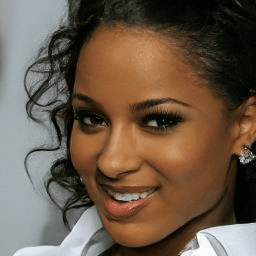} \\
        \includegraphics[width=0.3\linewidth]{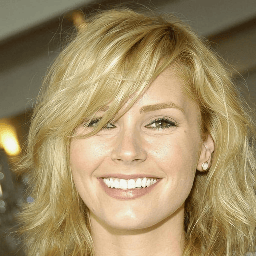} &
        \includegraphics[width=0.3\linewidth]{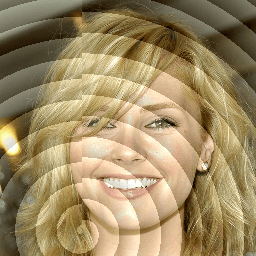} &
        \includegraphics[width=0.3\linewidth]{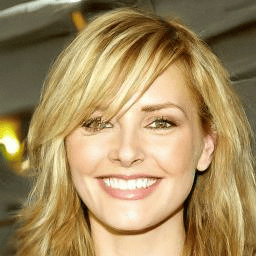} \\
        \includegraphics[width=0.3\linewidth]{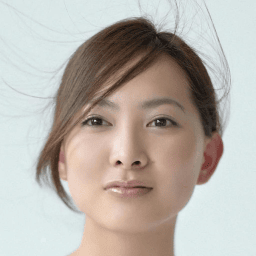} &
        \includegraphics[width=0.3\linewidth]{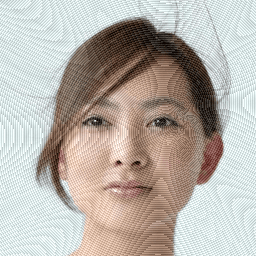} &
        \includegraphics[width=0.3\linewidth]{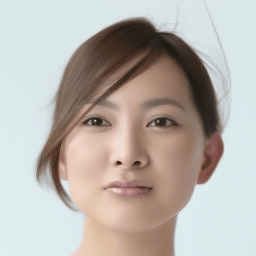} \\

        \multicolumn{1}{c}{\makecell{\scriptsize{Source}}} &
        {\makecell{\scriptsize{Input}}} &
        \multicolumn{1}{c}{\makecell{\scriptsize{\textbf{CODE (ours)}}}}  \\

        \end{tabular}
    \caption{Qualitative results of Concentric Sine Waves Noise}
    \label{fig:appendix-sine-waves}
\end{figure}
\newpage


\subsection{Fish-Eye Noise}
\begin{table}[htbp]
    \centering
    \scriptsize
    \begin{tabularx}{\columnwidth}{@{}ccccc@{}}
    \toprule[1.5pt]
     & LPIPS-Source $\downarrow$ &  SSIM-Input $\uparrow$ & PSNR-Input $\uparrow$ & FID $\downarrow$ \\
    \midrule[1pt]
    \makecell{\textit{Inputs}} & 0.45 & - & - & 341.37 \\
    \midrule[1pt]
    \makecell{\textit{DDIM}\\w/o CBC} & 0.46 & 0.81 & 25.85 & 300.68 \\
    \midrule[1pt]
    \makecell{\textit{DDIM}\\w/ CBC} & 0.39 & 0.76 & 23.81 & 85.72 \\
    \midrule[1pt]
    \makecell{\textbf{CODE}\\\textbf{w/o CBC}} & 0.29 & 0.59 & 21.85 & 34.02 \\
    \midrule[1pt]
    \makecell{\textbf{CODE}\\\textbf{w/ CBC}} & 0.29 & 0.62 & 22.01 & 25.54 \\
    \midrule[1pt]
    \makecell{\textit{SDEdit}} & 0.29 & 0.58 & 20.92 & 41.90 \\
    \bottomrule[1.5pt]
    \end{tabularx}
    \caption{Quantitative results on Fish-Eye Noise}
\end{table}
\begin{figure}[h!]
    \centering
    \setlength{\tabcolsep}{0pt}
    \renewcommand{\arraystretch}{0} 
    \begin{tabular}{c@{}c@{}c}
        \includegraphics[width=0.3\linewidth]{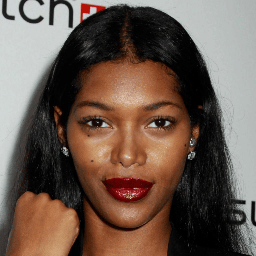} &
        \includegraphics[width=0.3\linewidth]{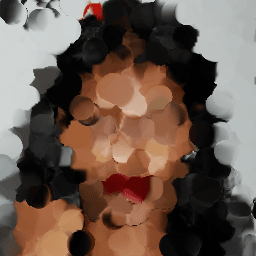} &
        \includegraphics[width=0.3\linewidth]{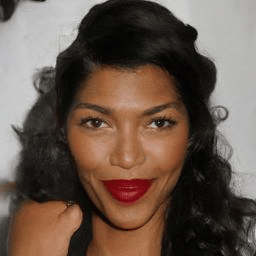} \\
        \includegraphics[width=0.3\linewidth]{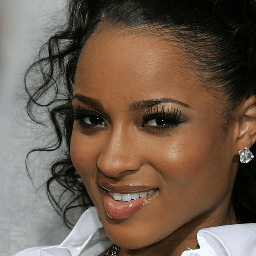} &
        \includegraphics[width=0.3\linewidth]{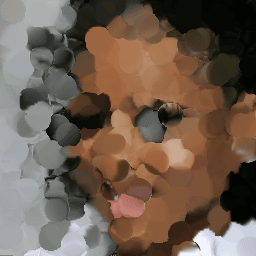} &
        \includegraphics[width=0.3\linewidth]{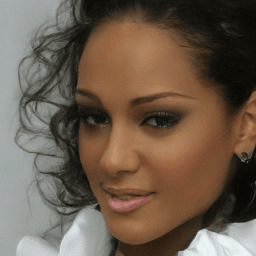} \\
        \includegraphics[width=0.3\linewidth]{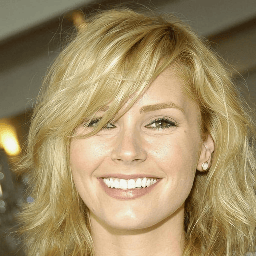} &
        \includegraphics[width=0.3\linewidth]{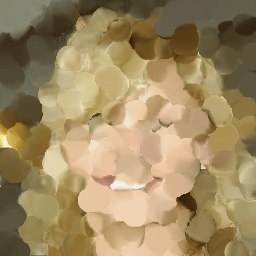} &
        \includegraphics[width=0.3\linewidth]{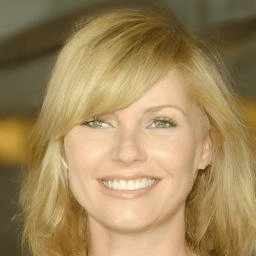} \\
        \includegraphics[width=0.3\linewidth]{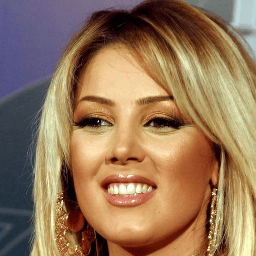} &
        \includegraphics[width=0.3\linewidth]{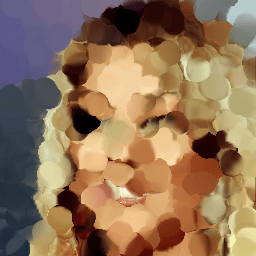} &
        \includegraphics[width=0.3\linewidth]{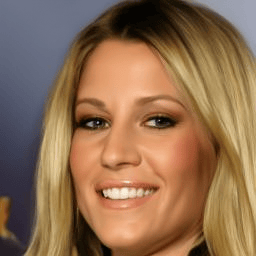} \\
        \includegraphics[width=0.3\linewidth]{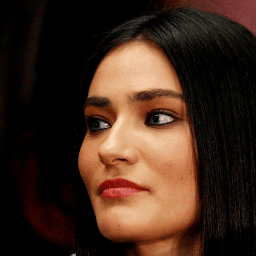} &
        \includegraphics[width=0.3\linewidth]{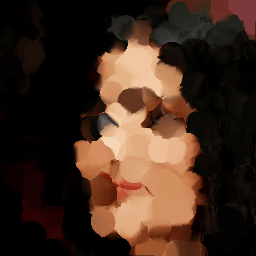} &
        \includegraphics[width=0.3\linewidth]{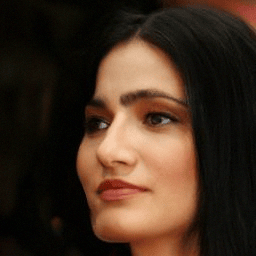} \\

        \multicolumn{1}{c}{\makecell{\scriptsize{Source}}} &
        {\makecell{\scriptsize{Input}}} &
        \multicolumn{1}{c}{\makecell{\scriptsize{\textbf{CODE (ours)}}}}  \\

        \end{tabular}
    \caption{Qualitative results of Fish-Eye Noise}
    \label{fig:appendix-fish-eye}
\end{figure}
\newpage


\subsection{Gaussian Blur}
\begin{table}[htbp]
    \centering
    \scriptsize
    \begin{tabularx}{\columnwidth}{@{}ccccc@{}}
    \toprule[1.5pt]
     & LPIPS-Source $\downarrow$ &  SSIM-Input $\uparrow$ & PSNR-Input $\uparrow$ & FID $\downarrow$ \\
    \midrule[1pt]
    \makecell{\textit{Inputs}} & 0.60 & - & - & 96.33 \\
    \midrule[1pt]
    \makecell{\textit{DDIM}\\w/o CBC} & 0.47 & 0.88 & 31.59 & 74.89 \\
    \midrule[1pt]
    \makecell{\textit{DDIM}\\w/ CBC} & 0.45 & 0.92 & 28.26 & 80.68 \\
    \midrule[1pt]
    \makecell{\textbf{CODE}\\\textbf{w/o CBC}} & 0.30 & 0.73 & 23.89 & 27.51 \\
    \midrule[1pt]
    \makecell{\textbf{CODE}\\\textbf{w/ CBC}} & 0.28 & 0.77 & 25.97 & 25.29 \\
    \midrule[1pt]
    \makecell{\textit{SDEdit}} & 0.29 & 0.70 & 21.74 & 34.72 \\
    \bottomrule[1.5pt]
    \end{tabularx}
    \caption{Quantitative results on Gaussian Blur}
\end{table}
\begin{figure}[h!]
    \centering
    \setlength{\tabcolsep}{0pt}
    \renewcommand{\arraystretch}{0} 
    \begin{tabular}{c@{}c@{}c}
        \includegraphics[width=0.3\linewidth]{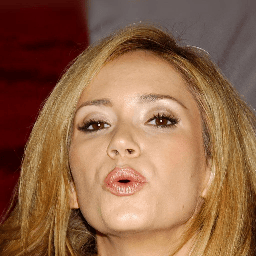} &
        \includegraphics[width=0.3\linewidth]{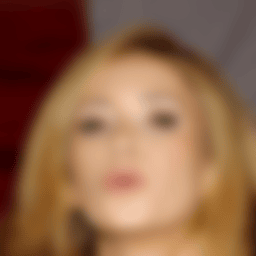} &
        \includegraphics[width=0.3\linewidth]{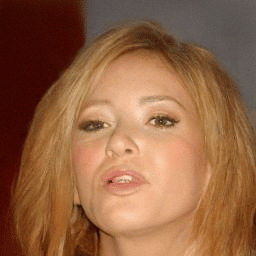} \\
        \includegraphics[width=0.3\linewidth]{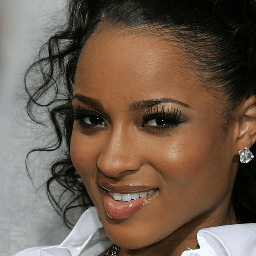} &
        \includegraphics[width=0.3\linewidth]{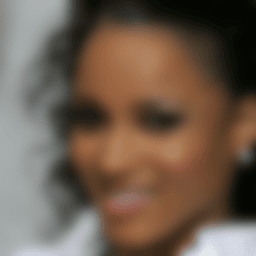} &
        \includegraphics[width=0.3\linewidth]{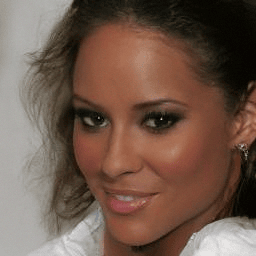} \\
        \includegraphics[width=0.3\linewidth]{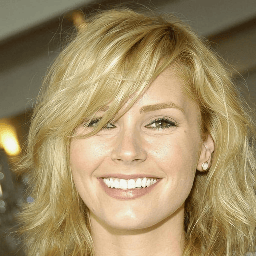} &
        \includegraphics[width=0.3\linewidth]{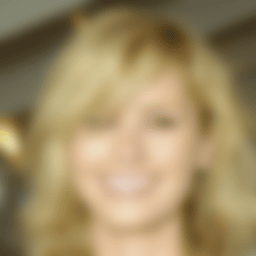} &
        \includegraphics[width=0.3\linewidth]{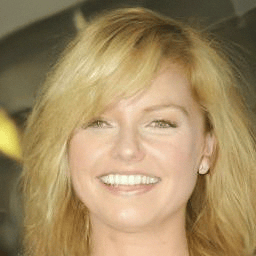} \\
        \includegraphics[width=0.3\linewidth]{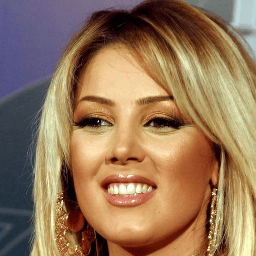} &
        \includegraphics[width=0.3\linewidth]{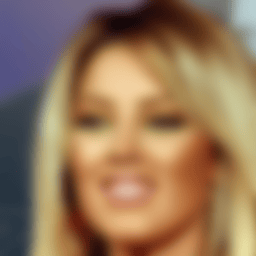} &
        \includegraphics[width=0.3\linewidth]{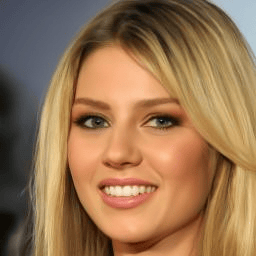} \\
        \includegraphics[width=0.3\linewidth]{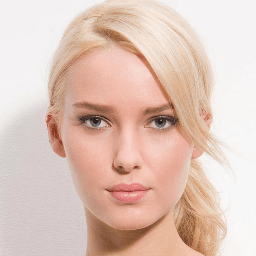} &
        \includegraphics[width=0.3\linewidth]{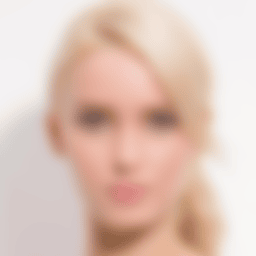} &
        \includegraphics[width=0.3\linewidth]{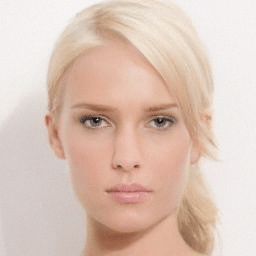} \\

        \multicolumn{1}{c}{\makecell{\scriptsize{Source}}} &
        {\makecell{\scriptsize{Input}}} &
        \multicolumn{1}{c}{\makecell{\scriptsize{\textbf{CODE (ours)}}}}  \\

        \end{tabular}
    \caption{Qualitative results of Gaussian Blur}
    \label{fig:appendix-gaussian-noise}
\end{figure}
\newpage


\subsection{Pinch and Twirl Noise}
\begin{table}[htbp]
    \centering
    \scriptsize
    \begin{tabularx}{\columnwidth}{@{}ccccc@{}}
    \toprule[1.5pt]
     & LPIPS-Source $\downarrow$ &  SSIM-Input $\uparrow$ & PSNR-Input $\uparrow$ & FID $\downarrow$ \\
    \midrule[1pt]
    \makecell{\textit{Inputs}} & 0.06 & - & - & 31.46 \\
    \midrule[1pt]
    \makecell{\textit{DDIM}\\w/o CBC} & 0.11 & 0.90 & 32.20 & 48.57 \\
    \midrule[1pt]
    \makecell{\textit{DDIM}\\w/ CBC} & 0.23 & 0.82 & 27.30 & 53.49 \\
    \midrule[1pt]
    \makecell{\textbf{CODE}\\\textbf{w/o CBC}} & 0.20 & 0.61 & 21.86 & 17.80 \\
    \midrule[1pt]
    \makecell{\textbf{CODE}\\\textbf{w/ CBC}} & 0.16 & 0.65 & 23.49 & 16.45 \\
    \midrule[1pt]
    \makecell{\textit{SDEdit}} & 0.17 & 0.64 & 22.33 & 28.07 \\
    \bottomrule[1.5pt]
    \end{tabularx}
    \caption{Quantitative results on Pinch and Twirl Noise}
\end{table}
\begin{figure}[h!]
    \centering
    \setlength{\tabcolsep}{0pt}
    \renewcommand{\arraystretch}{0} 
    \begin{tabular}{c@{}c@{}c}
        \includegraphics[width=0.3\linewidth]{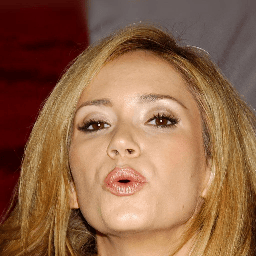} &
        \includegraphics[width=0.3\linewidth]{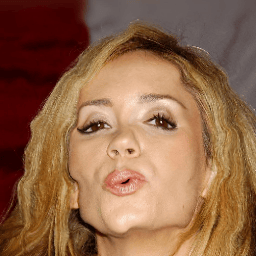} &
        \includegraphics[width=0.3\linewidth]{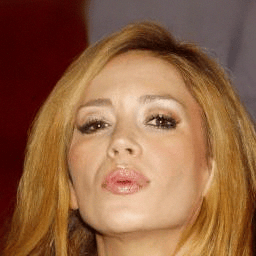} \\
        \includegraphics[width=0.3\linewidth]{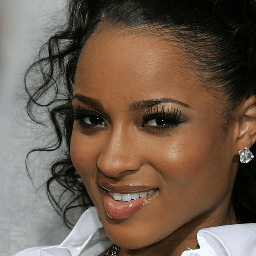} &
        \includegraphics[width=0.3\linewidth]{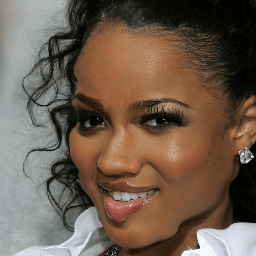} &
        \includegraphics[width=0.3\linewidth]{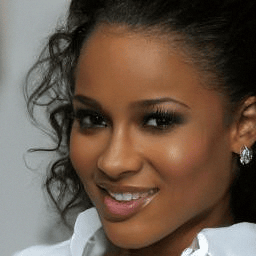} \\
        \includegraphics[width=0.3\linewidth]{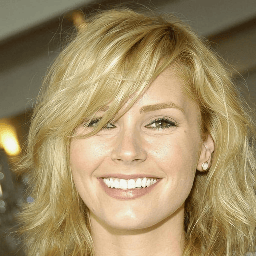} &
        \includegraphics[width=0.3\linewidth]{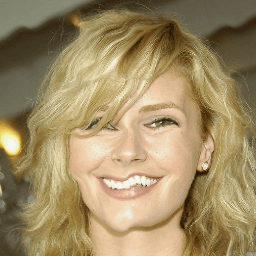} &
        \includegraphics[width=0.3\linewidth]{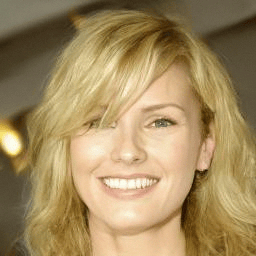} \\
        \includegraphics[width=0.3\linewidth]{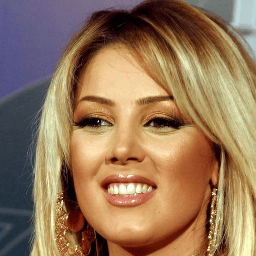} &
        \includegraphics[width=0.3\linewidth]{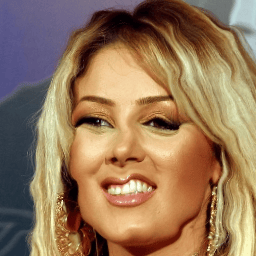} &
        \includegraphics[width=0.3\linewidth]{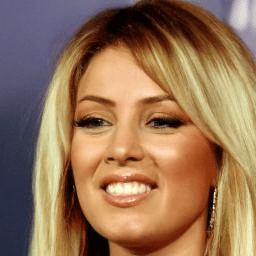} \\
        \includegraphics[width=0.3\linewidth]{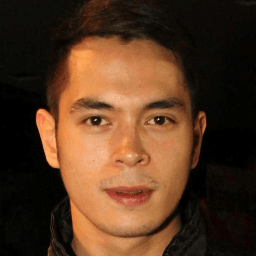} &
        \includegraphics[width=0.3\linewidth]{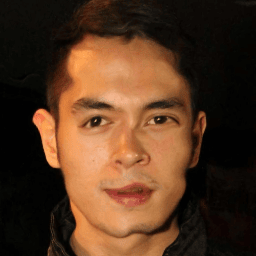} &
        \includegraphics[width=0.3\linewidth]{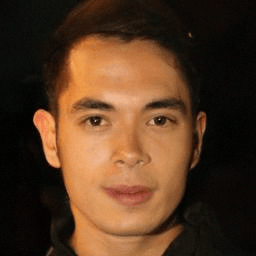} \\

        \multicolumn{1}{c}{\makecell{\scriptsize{Source}}} &
        {\makecell{\scriptsize{Input}}} &
        \multicolumn{1}{c}{\makecell{\scriptsize{\textbf{CODE (ours)}}}}  \\

        \end{tabular}
    \caption{Qualitative results of Pinch and Twirl Noise}
    \label{fig:appendix-pinch}
\end{figure}
\newpage


\subsection{Pixelate Noise}
\begin{table}[htbp]
    \centering
    \scriptsize
    \begin{tabularx}{\columnwidth}{@{}ccccc@{}}
    \toprule[1.5pt]
     & LPIPS-Source $\downarrow$ &  SSIM-Input $\uparrow$ & PSNR-Input $\uparrow$ & FID $\downarrow$ \\
    \midrule[1pt]
    \makecell{\textit{Inputs}} & 0.41 & - & - & 108.67 \\
    \midrule[1pt]
    \makecell{\textit{DDIM}\\w/o CBC} & 0.40 & 0.91 & 31.36 & 110.65 \\
    \midrule[1pt]
    \makecell{\textit{DDIM}\\w/ CBC} & 0.29 & 0.82 & 27.26 & 68.65 \\
    \midrule[1pt]
    \makecell{\textbf{CODE}\\\textbf{w/o CBC}} & 0.19 & 0.70 & 24.30 & 17.47 \\
    \midrule[1pt]
    \makecell{\textbf{CODE}\\\textbf{w/ CBC}} & 0.21 & 0.72 & 24.55 & 23.87 \\
    \midrule[1pt]
    \makecell{\textit{SDEdit}} & 0.22 & 0.62 & 20.63 & 25.76 \\
    \bottomrule[1.5pt]
    \end{tabularx}
    \caption{Quantitative results on Pixelate Noise}
\end{table}
\begin{figure}[h!]
    \centering
    \setlength{\tabcolsep}{0pt}
    \renewcommand{\arraystretch}{0} 
    \begin{tabular}{c@{}c@{}c}
        \includegraphics[width=0.3\linewidth]{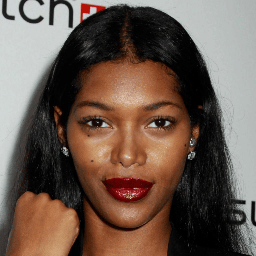} &
        \includegraphics[width=0.3\linewidth]{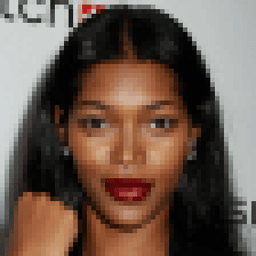} &
        \includegraphics[width=0.3\linewidth]{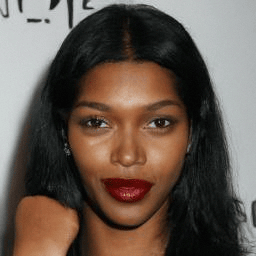} \\
        \includegraphics[width=0.3\linewidth]{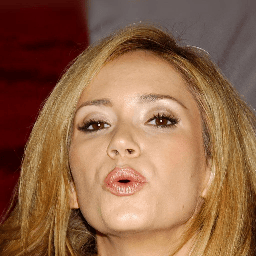} &
        \includegraphics[width=0.3\linewidth]{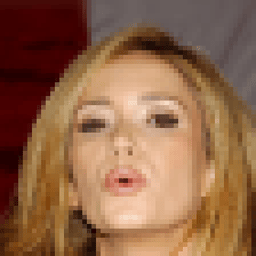} &
        \includegraphics[width=0.3\linewidth]{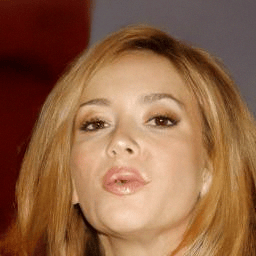} \\
        \includegraphics[width=0.3\linewidth]{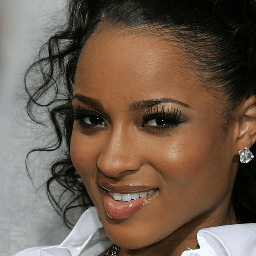} &
        \includegraphics[width=0.3\linewidth]{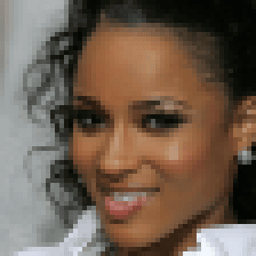} &
        \includegraphics[width=0.3\linewidth]{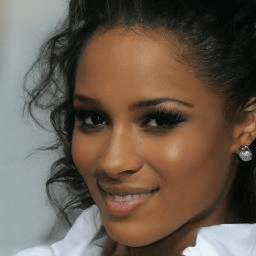} \\
        \includegraphics[width=0.3\linewidth]{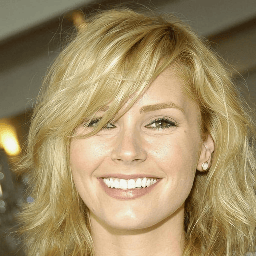} &
        \includegraphics[width=0.3\linewidth]{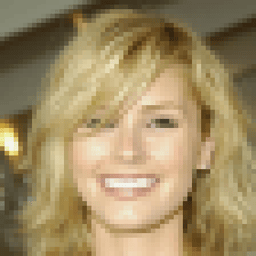} &
        \includegraphics[width=0.3\linewidth]{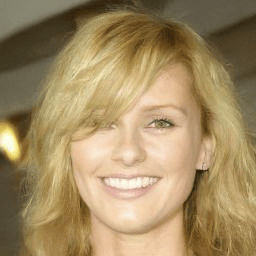} \\
        \includegraphics[width=0.3\linewidth]{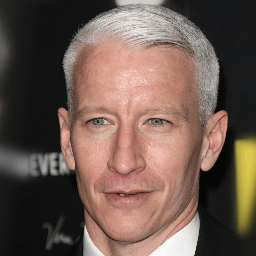} &
        \includegraphics[width=0.3\linewidth]{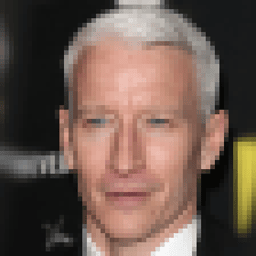} &
        \includegraphics[width=0.3\linewidth]{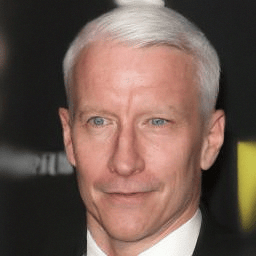} \\
        \multicolumn{1}{c}{\makecell{\scriptsize{Source}}} &
        {\makecell{\scriptsize{Input}}} &
        \multicolumn{1}{c}{\makecell{\scriptsize{\textbf{CODE (ours)}}}}  \\

        \end{tabular}
    \caption{Qualitative results of Pixelate Noise}
    \label{fig:appendix-pixelate}
\end{figure}
\newpage


\subsection{Scatter Noise}
\begin{table}[htbp]
    \centering
    \scriptsize
    \begin{tabularx}{\columnwidth}{@{}ccccc@{}}
    \toprule[1.5pt]
     & LPIPS-Source $\downarrow$ &  SSIM-Input $\uparrow$ & PSNR-Input $\uparrow$ & FID $\downarrow$ \\
    \midrule[1pt]
    \makecell{\textit{Inputs}} & 0.93 & - & - & 431.49 \\
    \midrule[1pt]
    \makecell{\textit{DDIM}\\w/o CBC} & 0.95 & 0.96 & 31.71 & 416.98 \\
    \midrule[1pt]
    \makecell{\textit{DDIM}\\w/ CBC} & 0.49 & 0.31 & 17.28 & 109.95 \\
    \midrule[1pt]
    \makecell{\textbf{CODE}\\\textbf{w/o CBC}} & 0.31 & 0.23 & 17.14 & 40.90 \\
    \midrule[1pt]
    \makecell{\textbf{CODE}\\\textbf{w/ CBC}} & 0.21 & 0.24 & 17.09 & 28.53 \\
    \midrule[1pt]
    \makecell{\textit{SDEdit}} & 0.31 & 0.23 & 17.07 & 49.21 \\
    \bottomrule[1.5pt]
    \end{tabularx}
    \caption{Quantitative results on Scatter Noise}
\end{table}
\begin{figure}[h!]
    \centering
    \setlength{\tabcolsep}{0pt}
    \renewcommand{\arraystretch}{0} 
    \begin{tabular}{c@{}c@{}c}
        \includegraphics[width=0.3\linewidth]{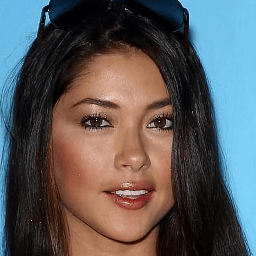} &
        \includegraphics[width=0.3\linewidth]{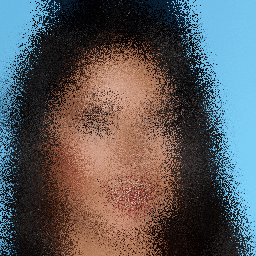} &
        \includegraphics[width=0.3\linewidth]{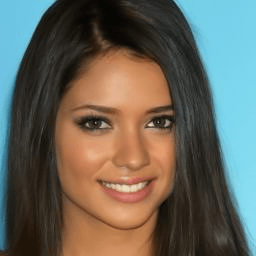} \\
        \includegraphics[width=0.3\linewidth]{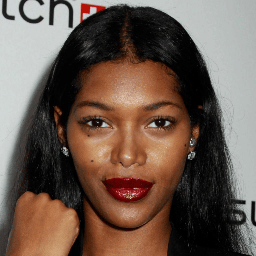} &
        \includegraphics[width=0.3\linewidth]{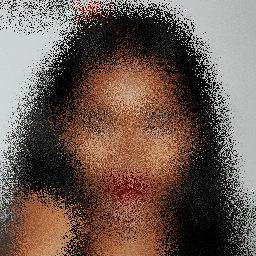} &
        \includegraphics[width=0.3\linewidth]{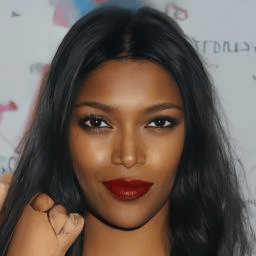} \\
        \includegraphics[width=0.3\linewidth]{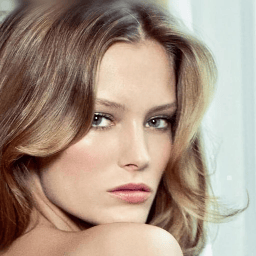} &
        \includegraphics[width=0.3\linewidth]{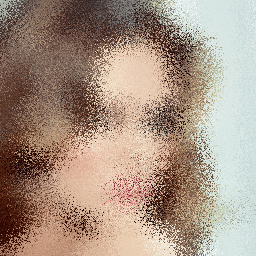} &
        \includegraphics[width=0.3\linewidth]{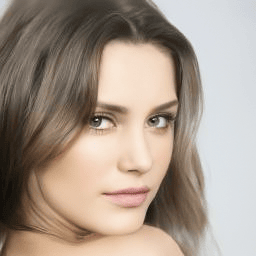} \\
        \includegraphics[width=0.3\linewidth]{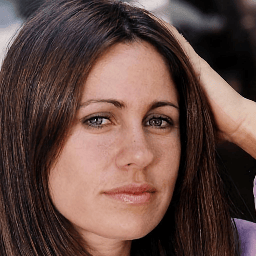} &
        \includegraphics[width=0.3\linewidth]{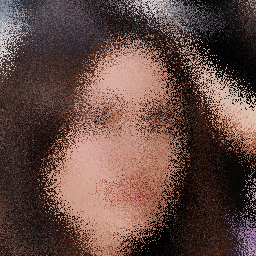} &
        \includegraphics[width=0.3\linewidth]{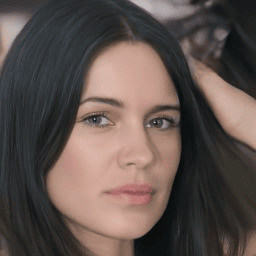} \\
        \includegraphics[width=0.3\linewidth]{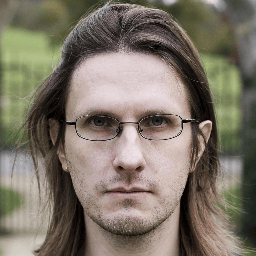} &
        \includegraphics[width=0.3\linewidth]{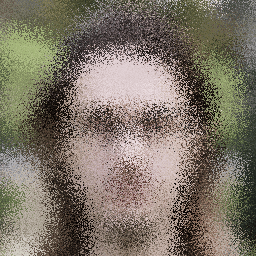} &
        \includegraphics[width=0.3\linewidth]{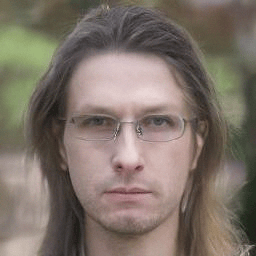} \\
        \multicolumn{1}{c}{\makecell{\scriptsize{Source}}} &
        {\makecell{\scriptsize{Input}}} &
        \multicolumn{1}{c}{\makecell{\scriptsize{\textbf{CODE (ours)}}}}  \\

        \end{tabular}
    \caption{Qualitative results of Scatter Noise}
    \label{fig:appendix-scatter}
\end{figure}
\newpage


\subsection{Water Drop Noise}
\begin{table}[htbp]
    \centering
    \scriptsize
    \begin{tabularx}{\columnwidth}{@{}ccccc@{}}
    \toprule[1.5pt]
     & LPIPS-Source $\downarrow$ &  SSIM-Input $\uparrow$ & PSNR-Input $\uparrow$ & FID $\downarrow$ \\
    \midrule[1pt]
    \makecell{\textit{Inputs}} & 0.11 & - & - & 40.65 \\
    \midrule[1pt]
    \makecell{\textit{DDIM}\\w/o CBC} & 0.18 & 0.87 & 28.44 & 64.28 \\
    \midrule[1pt]
    \makecell{\textit{DDIM}\\w/ CBC} & 0.26 & 0.80 & 26.42 & 64.48 \\
    \midrule[1pt]
    \makecell{\textbf{CODE}\\\textbf{w/o CBC}} & 0.23 & 0.58 & 20.73 & 20.7 \\
    \midrule[1pt]
    \makecell{\textbf{CODE}\\\textbf{w/ CBC}} & 0.19 & 0.62 & 22.62 & 18.86 \\
    \midrule[1pt]
    \makecell{\textit{SDEdit}} & 0.19 & 0.62 & 21.80 & 27.61 \\
    \bottomrule[1.5pt]
    \end{tabularx}
    \caption{Quantitative results on Water Drop Noise}
\end{table}
\begin{figure}[h!]
    \centering
    \setlength{\tabcolsep}{0pt}
    \renewcommand{\arraystretch}{0} 
    \begin{tabular}{c@{}c@{}c}
        \includegraphics[width=0.3\linewidth]{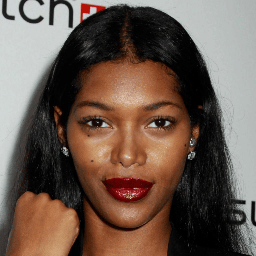} &
        \includegraphics[width=0.3\linewidth]{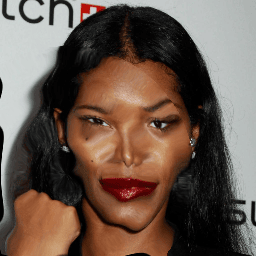} &
        \includegraphics[width=0.3\linewidth]{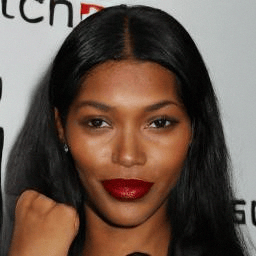} \\
        \includegraphics[width=0.3\linewidth]{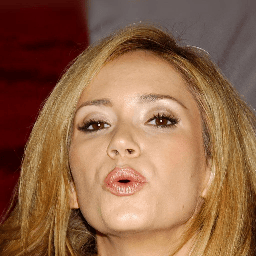} &
        \includegraphics[width=0.3\linewidth]{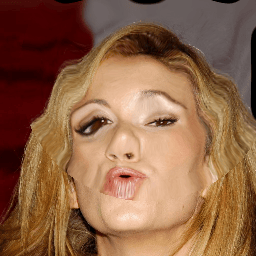} &
        \includegraphics[width=0.3\linewidth]{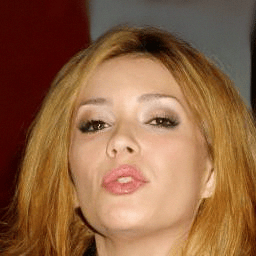} \\
        \includegraphics[width=0.3\linewidth]{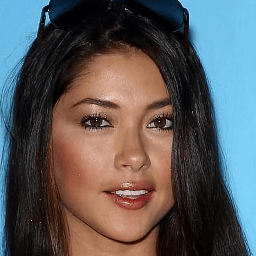} &
        \includegraphics[width=0.3\linewidth]{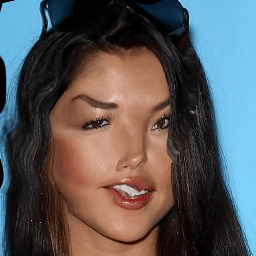} &
        \includegraphics[width=0.3\linewidth]{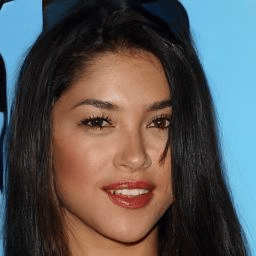} \\
        \includegraphics[width=0.3\linewidth]{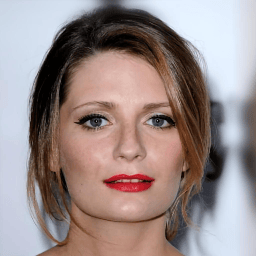} &
        \includegraphics[width=0.3\linewidth]{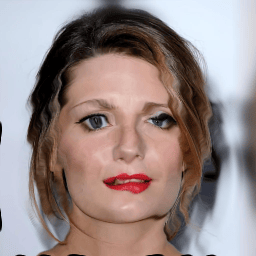} &
        \includegraphics[width=0.3\linewidth]{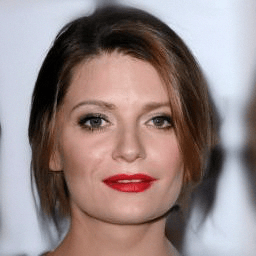} \\
        \includegraphics[width=0.3\linewidth]{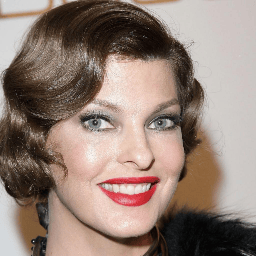} &
        \includegraphics[width=0.3\linewidth]{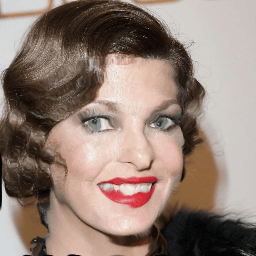} &
        \includegraphics[width=0.3\linewidth]{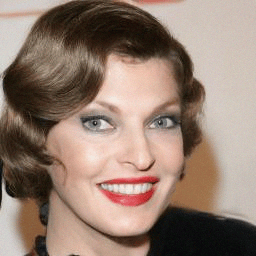} \\
        \multicolumn{1}{c}{\makecell{\scriptsize{Source}}} &
        {\makecell{\scriptsize{Input}}} &
        \multicolumn{1}{c}{\makecell{\scriptsize{\textbf{CODE (ours)}}}}  \\
        \end{tabular}
    \caption{Qualitative results of Water Drop Noise}
    \label{fig:appendix-waterdrop}
\end{figure}
\newpage


\subsection{Bleach Bypass}
\begin{table}[htbp]
    \centering
    \scriptsize
    \begin{tabularx}{\columnwidth}{@{}ccccc@{}}
    \toprule[1.5pt]
     & LPIPS-Source $\downarrow$ &  SSIM-Input $\uparrow$ & PSNR-Input $\uparrow$ & FID $\downarrow$ \\
    \midrule[1pt]
    \makecell{\textit{Inputs}} & 0.1 & - & - & 33.95 \\
    \midrule[1pt]
    \makecell{\textit{DDIM}\\w/o CBC} & 0.14 & 0.92 & 32.12 & 52.78 \\
    \midrule[1pt]
    \makecell{\textit{DDIM}\\w/ CBC} & 0.3 & 0.71 & 22.95 & 78.34 \\
    \midrule[1pt]
    \makecell{\textbf{CODE}\\\textbf{w/o CBC}} & 0.14 & 0.76 & 26.46 & 26.41 \\
    \midrule[1pt]
    \makecell{\textbf{CODE}\\\textbf{w/ CBC}} & 0.2 & 0.68 & 23.38 & 21.81 \\
    \midrule[1pt]
    \makecell{\textit{SDEdit}} & 0.19 & 0.62 & 21.52 & 26.98 \\
    \bottomrule[1.5pt]
    \end{tabularx}
    \caption{Quantitative results on Bleach Bypass}
\end{table}
\begin{figure}[h!]
    \centering
    \setlength{\tabcolsep}{0pt}
    \renewcommand{\arraystretch}{0} 
    \begin{tabular}{c@{}c@{}c}
        \includegraphics[width=0.3\linewidth]{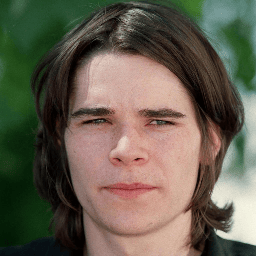} &
        \includegraphics[width=0.3\linewidth]{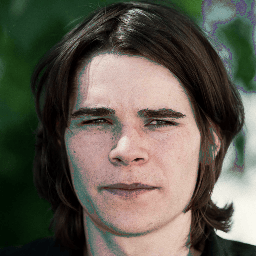} &
        \includegraphics[width=0.3\linewidth]{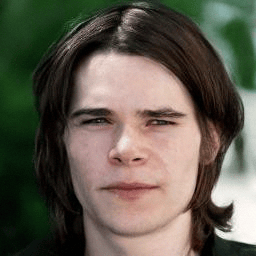} \\
        \includegraphics[width=0.3\linewidth]{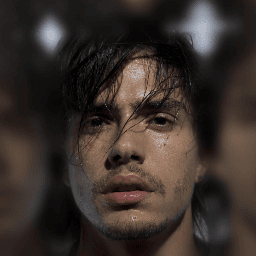} &
        \includegraphics[width=0.3\linewidth]{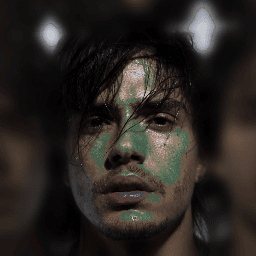} &
        \includegraphics[width=0.3\linewidth]{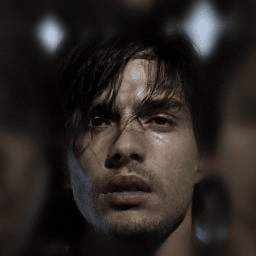} \\
        \includegraphics[width=0.3\linewidth]{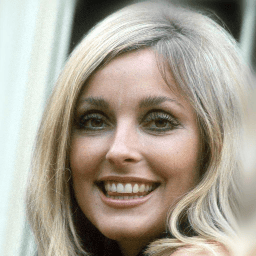} &
        \includegraphics[width=0.3\linewidth]{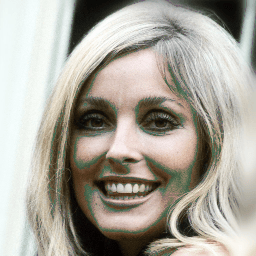} &
        \includegraphics[width=0.3\linewidth]{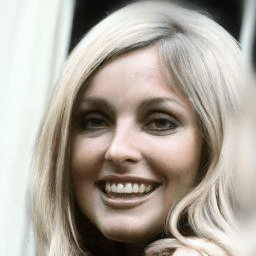} \\
        \includegraphics[width=0.3\linewidth]{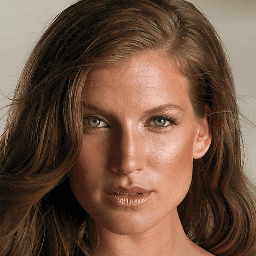} &
        \includegraphics[width=0.3\linewidth]{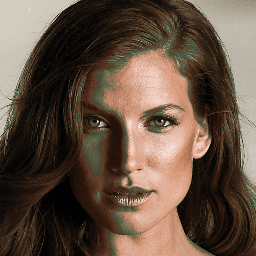} &
        \includegraphics[width=0.3\linewidth]{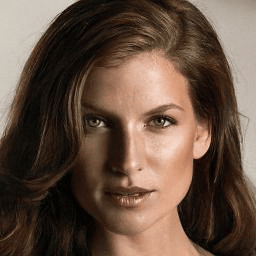} \\
        \includegraphics[width=0.3\linewidth]{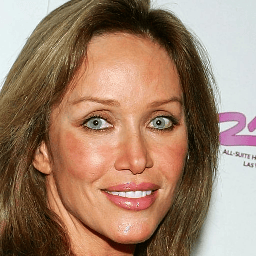} &
        \includegraphics[width=0.3\linewidth]{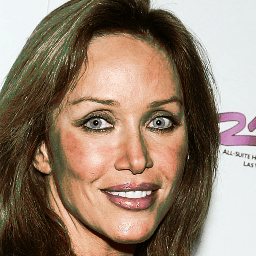} &
        \includegraphics[width=0.3\linewidth]{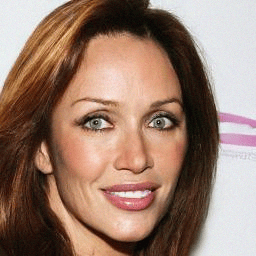} \\
        \multicolumn{1}{c}{\makecell{\scriptsize{Source}}} &
        {\makecell{\scriptsize{Input}}} &
        \multicolumn{1}{c}{\makecell{\scriptsize{\textbf{CODE (ours)}}}}  \\
        \end{tabular}
    \caption{Qualitative results of Bleach Bypass}
\end{figure}
\newpage


\subsection{Blue Noise}
\begin{table}[htbp]
    \centering
    \scriptsize
    \begin{tabularx}{\columnwidth}{@{}ccccc@{}}
    \toprule[1.5pt]
     & LPIPS-Source $\downarrow$ &  SSIM-Input $\uparrow$ & PSNR-Input $\uparrow$ & FID $\downarrow$ \\
    \midrule[1pt]
    \makecell{\textit{Inputs}} & 1.19 & - & - & 119.76 \\
    \midrule[1pt]
    \makecell{\textit{DDIM}\\w/o CBC} & 1.2 & 0.98 & 36.56 & 135.39 \\
    \midrule[1pt]
    \makecell{\textit{DDIM}\\w/ CBC} & 0.83 & 0.8 & 16.51 & 64.15 \\
    \midrule[1pt]
    \makecell{\textbf{CODE}\\\textbf{w/o CBC}} & 0.21 & 0.05 & 10.8 & 21.69 \\
    \midrule[1pt]
    \makecell{\textbf{CODE}\\\textbf{w/ CBC}} & 0.19 & 0.06 & 10.78 & 17.5 \\
    \midrule[1pt]
    \makecell{\textit{SDEdit}} & 0.23 & 0.03 & 10.72 & 30.71 \\
    \bottomrule[1.5pt]
    \end{tabularx}
    \caption{Quantitative results on Blue Noise}
\end{table}
\begin{figure}[h!]
    \centering
    \setlength{\tabcolsep}{0pt}
    \renewcommand{\arraystretch}{0} 
    \begin{tabular}{c@{}c@{}c}
        \includegraphics[width=0.3\linewidth]{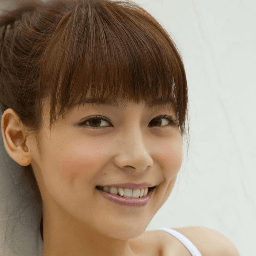} &
        \includegraphics[width=0.3\linewidth]{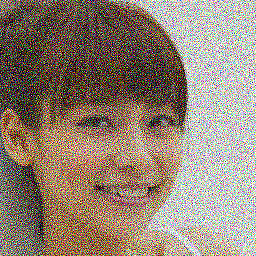} &
        \includegraphics[width=0.3\linewidth]{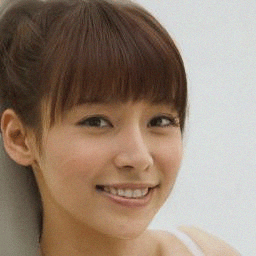} \\
        \includegraphics[width=0.3\linewidth]{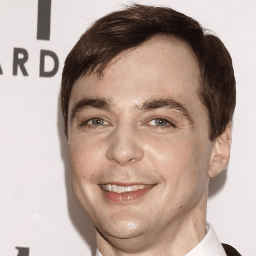} &
        \includegraphics[width=0.3\linewidth]{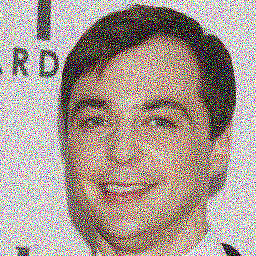} &
        \includegraphics[width=0.3\linewidth]{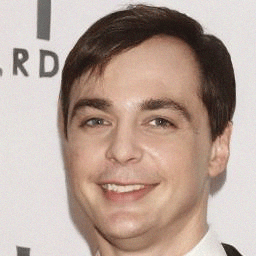} \\
        \includegraphics[width=0.3\linewidth]{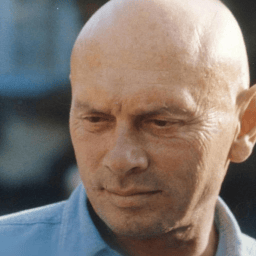} &
        \includegraphics[width=0.3\linewidth]{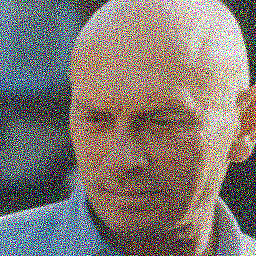} &
        \includegraphics[width=0.3\linewidth]{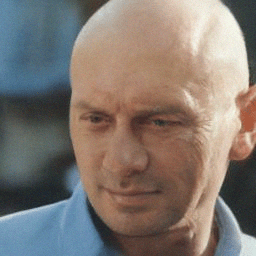} \\
        \includegraphics[width=0.3\linewidth]{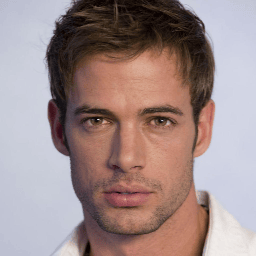} &
        \includegraphics[width=0.3\linewidth]{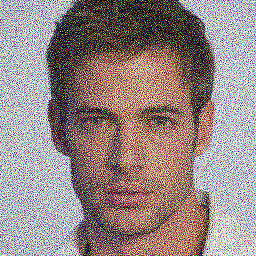} &
        \includegraphics[width=0.3\linewidth]{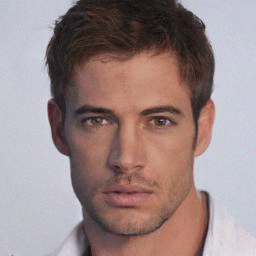} \\
        \includegraphics[width=0.3\linewidth]{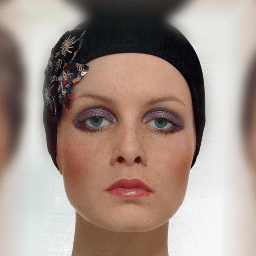} &
        \includegraphics[width=0.3\linewidth]{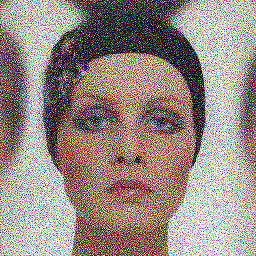} &
        \includegraphics[width=0.3\linewidth]{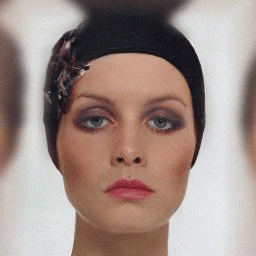} \\
        \multicolumn{1}{c}{\makecell{\scriptsize{Source}}} &
        {\makecell{\scriptsize{Input}}} &
        \multicolumn{1}{c}{\makecell{\scriptsize{\textbf{CODE (ours)}}}}  \\
        \end{tabular}
    \caption{Qualitative results of Blue Noise}
\end{figure}
\newpage


\subsection{Blue Noise Sample}
\begin{table}[htbp]
    \centering
    \scriptsize
    \begin{tabularx}{\columnwidth}{@{}ccccc@{}}
    \toprule[1.5pt]
     & LPIPS-Source $\downarrow$ &  SSIM-Input $\uparrow$ & PSNR-Input $\uparrow$ & FID $\downarrow$ \\
    \midrule[1pt]
    \makecell{\textit{Inputs}} & 0.95 & - & - & 208.28 \\
    \midrule[1pt]
    \makecell{\textit{DDIM}\\w/o CBC} & 0.93 & 0.99 & 37.88 & 206,76 \\
    \midrule[1pt]
    \makecell{\textit{DDIM}\\w/ CBC} & 0.23 & 0.33 & 13.17 & 62.09 \\
    \midrule[1pt]
    \makecell{\textbf{CODE}\\\textbf{w/o CBC}} & 0.24 & 0.17 & 13.91 & 24.13 \\
    \midrule[1pt]
    \makecell{\textbf{CODE}\\\textbf{w/ CBC}} & 0.19 & 0.23 & 13.33 & 24.19 \\
    \midrule[1pt]
    \makecell{\textit{SDEdit}} & 0.23 & 0.19 & 13.95 & 27.13 \\
    \bottomrule[1.5pt]
    \end{tabularx}
    \caption{Quantitative results on Blue Noise Sample}
\end{table}
\begin{figure}[h!]
    \centering
    \setlength{\tabcolsep}{0pt}
    \renewcommand{\arraystretch}{0} 
    \begin{tabular}{c@{}c@{}c}
        \includegraphics[width=0.3\linewidth]{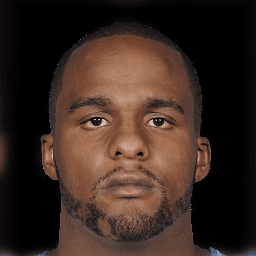} &
        \includegraphics[width=0.3\linewidth]{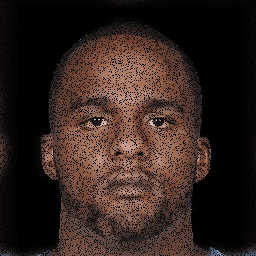} &
        \includegraphics[width=0.3\linewidth]{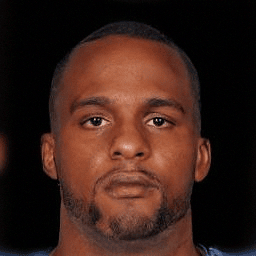} \\
        \includegraphics[width=0.3\linewidth]{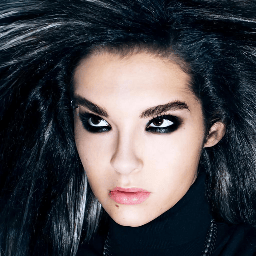} &
        \includegraphics[width=0.3\linewidth]{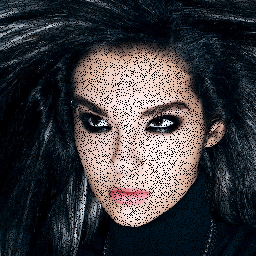} &
        \includegraphics[width=0.3\linewidth]{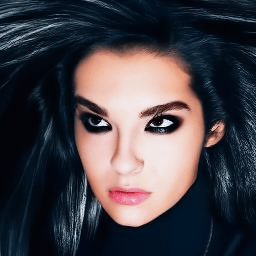} \\
        \includegraphics[width=0.3\linewidth]{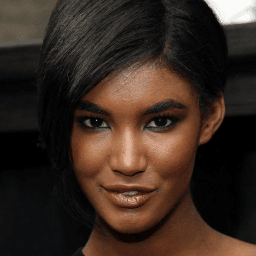} &
        \includegraphics[width=0.3\linewidth]{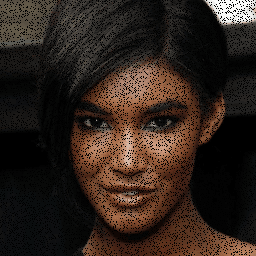} &
        \includegraphics[width=0.3\linewidth]{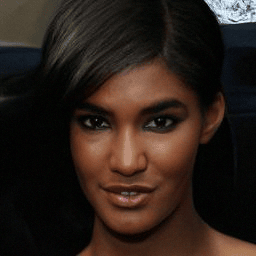} \\
        \includegraphics[width=0.3\linewidth]{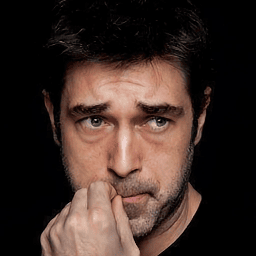} &
        \includegraphics[width=0.3\linewidth]{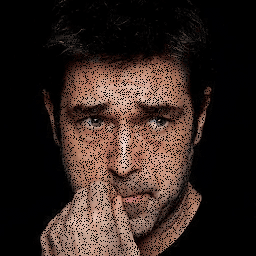} &
        \includegraphics[width=0.3\linewidth]{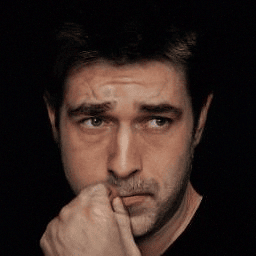} \\
        \includegraphics[width=0.3\linewidth]{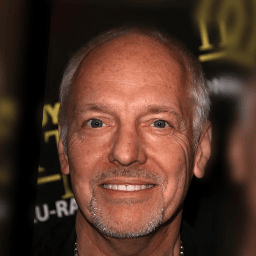} &
        \includegraphics[width=0.3\linewidth]{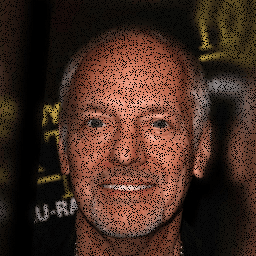} &
        \includegraphics[width=0.3\linewidth]{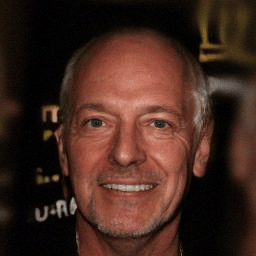} \\
        \multicolumn{1}{c}{\makecell{\scriptsize{Source}}} &
        {\makecell{\scriptsize{Input}}} &
        \multicolumn{1}{c}{\makecell{\scriptsize{\textbf{CODE (ours)}}}}  \\
        \end{tabular}
    \caption{Qualitative results of Blue Noise Sample}
\end{figure}
\newpage


\subsection{Brightness}
\begin{table}[htbp]
    \centering
    \scriptsize
    \begin{tabularx}{\columnwidth}{@{}ccccc@{}}
    \toprule[1.5pt]
     & LPIPS-Source $\downarrow$ &  SSIM-Input $\uparrow$ & PSNR-Input $\uparrow$ & FID $\downarrow$ \\
    \midrule[1pt]
    \makecell{\textit{Inputs}} & 0.26 & - & - & 47.17 \\
    \midrule[1pt]
    \makecell{\textit{DDIM}\\w/o CBC} & 0.27 & 0.95 & 36.3 & 52.37 \\
    \midrule[1pt]
    \makecell{\textit{DDIM}\\w/ CBC} & 0.41 & 0.75 & 25.47 & 72.34 \\
    \midrule[1pt]
    \makecell{\textbf{CODE}\\\textbf{w/o CBC}} & 0.39 & 0.56 & 18.91 & 38.65 \\
    \midrule[1pt]
    \makecell{\textbf{CODE}\\\textbf{w/ CBC}} & 0.42 & 0.56 & 18.75 & 42.0 \\
    \midrule[1pt]
    \makecell{\textit{SDEdit}} & 0.37 & 0.59 & 20.53 & 41.16 \\
    \bottomrule[1.5pt]
    \end{tabularx}
    \caption{Quantitative results on brightness}
\end{table}
\begin{figure}[h!]
    \centering
    \setlength{\tabcolsep}{0pt}
    \renewcommand{\arraystretch}{0} 
    \begin{tabular}{c@{}c@{}c}
        \includegraphics[width=0.3\linewidth]{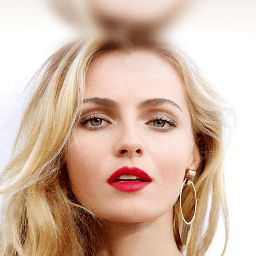} &
        \includegraphics[width=0.3\linewidth]{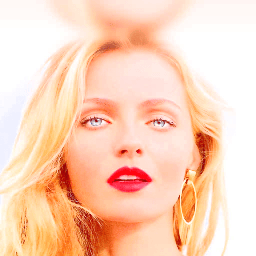} &
        \includegraphics[width=0.3\linewidth]{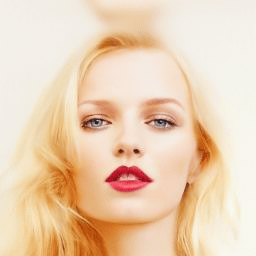} \\
        \includegraphics[width=0.3\linewidth]{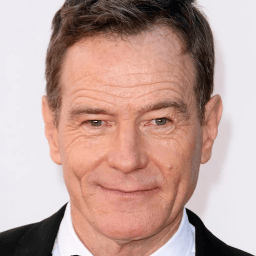} &
        \includegraphics[width=0.3\linewidth]{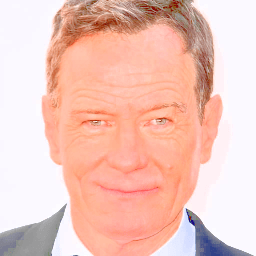} &
        \includegraphics[width=0.3\linewidth]{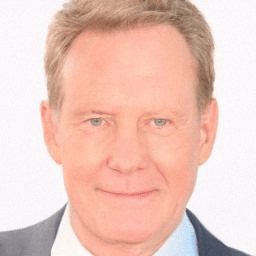} \\
        \includegraphics[width=0.3\linewidth]{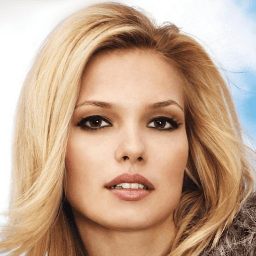} &
        \includegraphics[width=0.3\linewidth]{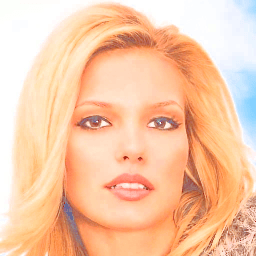} &
        \includegraphics[width=0.3\linewidth]{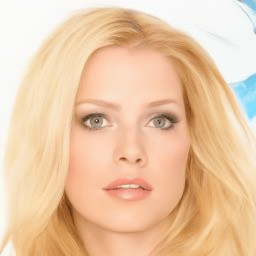} \\
        \includegraphics[width=0.3\linewidth]{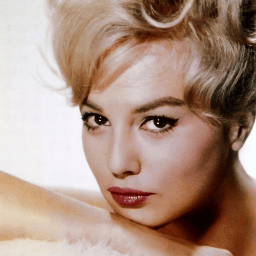} &
        \includegraphics[width=0.3\linewidth]{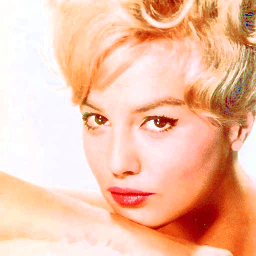} &
        \includegraphics[width=0.3\linewidth]{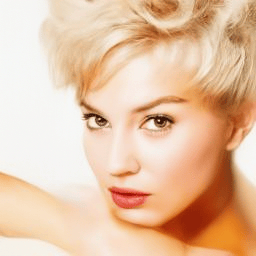} \\
        \includegraphics[width=0.3\linewidth]{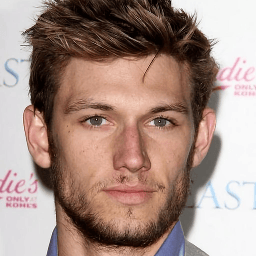} &
        \includegraphics[width=0.3\linewidth]{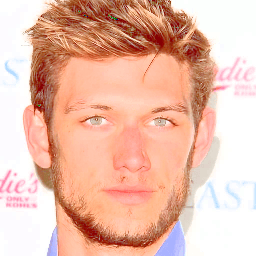} &
        \includegraphics[width=0.3\linewidth]{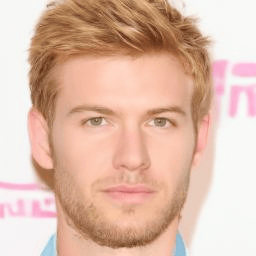} \\
        \multicolumn{1}{c}{\makecell{\scriptsize{Source}}} &
        {\makecell{\scriptsize{Input}}} &
        \multicolumn{1}{c}{\makecell{\scriptsize{\textbf{CODE (ours)}}}}  \\
        \end{tabular}
    \caption{Qualitative results of Brightness}
\end{figure}
\newpage


\subsection{Checkerboard Cutout}
\begin{table}[htbp]
    \centering
    \scriptsize
    \begin{tabularx}{\columnwidth}{@{}ccccc@{}}
    \toprule[1.5pt]
     & LPIPS-Source $\downarrow$ &  SSIM-Input $\uparrow$ & PSNR-Input $\uparrow$ & FID $\downarrow$ \\
    \midrule[1pt]
    \makecell{\textit{Inputs}} & 0.34 & - & - & 246.46 \\
    \midrule[1pt]
    \makecell{\textit{DDIM}\\w/o CBC} & 0.35 & 0.88 & 28.46 & 216.47 \\
    \midrule[1pt]
    \makecell{\textit{DDIM}\\w/ CBC} & 0.35 & 0.75 & 23.14 & 93.09 \\
    \midrule[1pt]
    \makecell{\textbf{CODE}\\\textbf{w/o CBC}} & 0.33 & 0.53 & 18.58 & 24.64 \\
    \midrule[1pt]
    \makecell{\textbf{CODE}\\\textbf{w/ CBC}} & 0.3 & 0.54 & 18.79 & 17.89 \\
    \midrule[1pt]
    \makecell{\textit{SDEdit}} & 0.35 & 0.51 & 18.17 & 35.4 \\
    \bottomrule[1.5pt]
    \end{tabularx}
    \caption{Quantitative results on Checkerboard Cutout}
\end{table}
\begin{figure}[h!]
    \centering
    \setlength{\tabcolsep}{0pt}
    \renewcommand{\arraystretch}{0} 
    \begin{tabular}{c@{}c@{}c}
        \includegraphics[width=0.3\linewidth]{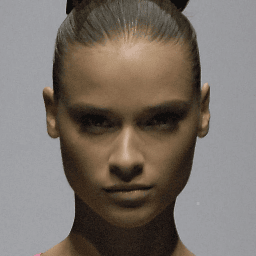} &
        \includegraphics[width=0.3\linewidth]{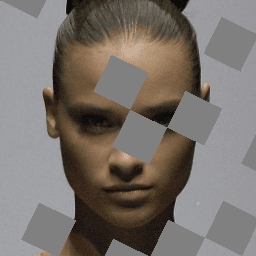} &
        \includegraphics[width=0.3\linewidth]{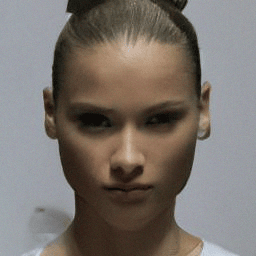} \\
        \includegraphics[width=0.3\linewidth]{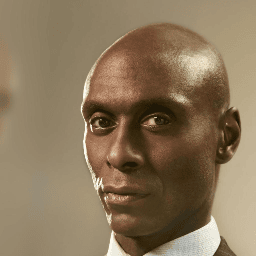} &
        \includegraphics[width=0.3\linewidth]{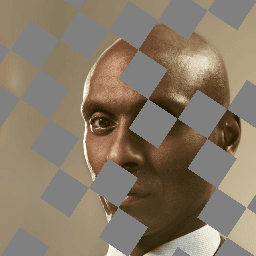} &
        \includegraphics[width=0.3\linewidth]{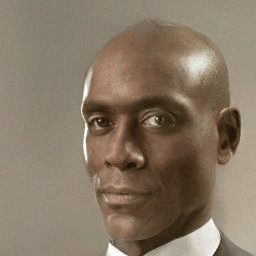} \\
        \includegraphics[width=0.3\linewidth]{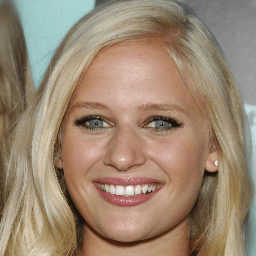} &
        \includegraphics[width=0.3\linewidth]{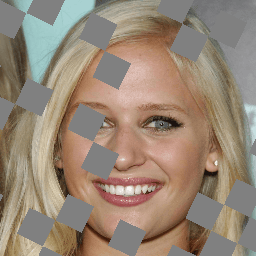} &
        \includegraphics[width=0.3\linewidth]{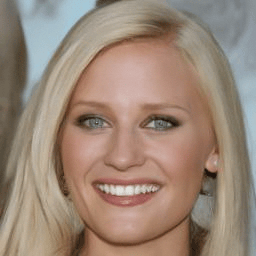} \\
        \includegraphics[width=0.3\linewidth]{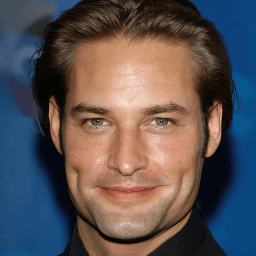} &
        \includegraphics[width=0.3\linewidth]{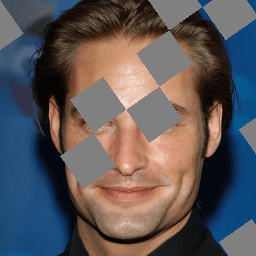} &
        \includegraphics[width=0.3\linewidth]{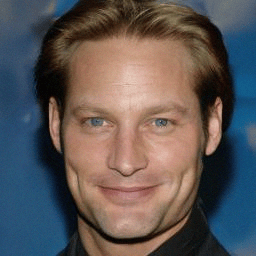} \\
        \includegraphics[width=0.3\linewidth]{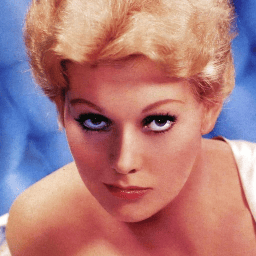} &
        \includegraphics[width=0.3\linewidth]{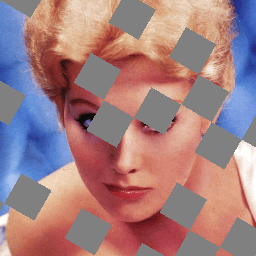} &
        \includegraphics[width=0.3\linewidth]{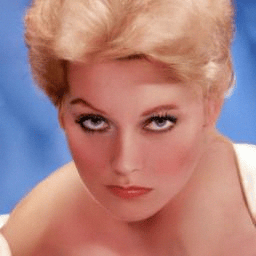} \\
        \multicolumn{1}{c}{\makecell{\scriptsize{Source}}} &
        {\makecell{\scriptsize{Input}}} &
        \multicolumn{1}{c}{\makecell{\scriptsize{\textbf{CODE (ours)}}}}  \\
        \end{tabular}
    \caption{Qualitative results of Checkerboard Cutout}
\end{figure}
\newpage


\subsection{Chromatic Aberration}
\begin{table}[htbp]
    \centering
    \scriptsize
    \begin{tabularx}{\columnwidth}{@{}ccccc@{}}
    \toprule[1.5pt]
     & LPIPS-Source $\downarrow$ &  SSIM-Input $\uparrow$ & PSNR-Input $\uparrow$ & FID $\downarrow$ \\
    \midrule[1pt]
    \makecell{\textit{Inputs}} & 0.53 & - & - & 188.53 \\
    \midrule[1pt]
    \makecell{\textit{DDIM}\\w/o CBC} & 0.52 & 0.95 & 37.67 & 186.9 \\
    \midrule[1pt]
    \makecell{\textit{DDIM}\\w/ CBC} & 0.6 & 0.79 & 24.22 & 190.87 \\
    \midrule[1pt]
    \makecell{\textbf{CODE}\\\textbf{w/o CBC}} & 0.44 & 0.44 & 16.74 & 57.56 \\
    \midrule[1pt]
    \makecell{\textbf{CODE}\\\textbf{w/ CBC}} & 0.44 & 0.46 & 17.25 & 56.05 \\
    \midrule[1pt]
    \makecell{\textit{SDEdit}} & 0.44 & 0.47 & 16.91 & 61.02 \\
    \bottomrule[1.5pt]
    \end{tabularx}
    \caption{Quantitative results on Chromatic Abberation}
\end{table}
\begin{figure}[h!]
    \centering
    \setlength{\tabcolsep}{0pt}
    \renewcommand{\arraystretch}{0} 
    \begin{tabular}{c@{}c@{}c}
        \includegraphics[width=0.3\linewidth]{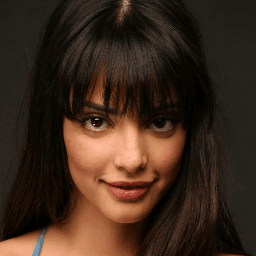} &
        \includegraphics[width=0.3\linewidth]{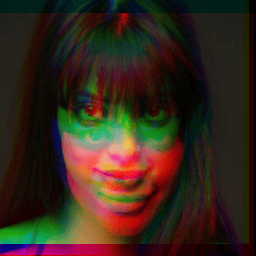} &
        \includegraphics[width=0.3\linewidth]{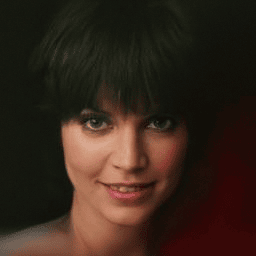} \\
        \includegraphics[width=0.3\linewidth]{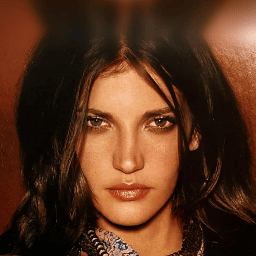} &
        \includegraphics[width=0.3\linewidth]{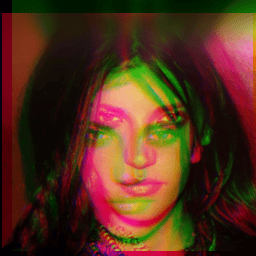} &
        \includegraphics[width=0.3\linewidth]{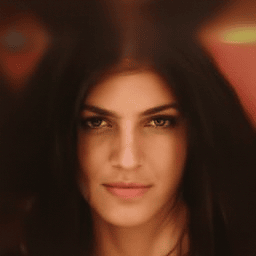} \\
        \includegraphics[width=0.3\linewidth]{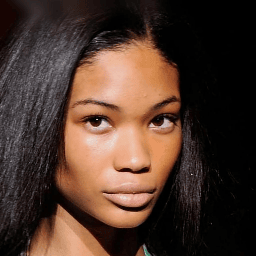} &
        \includegraphics[width=0.3\linewidth]{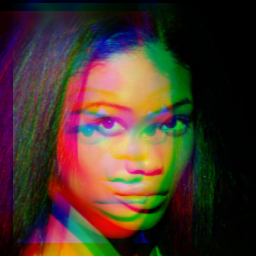} &
        \includegraphics[width=0.3\linewidth]{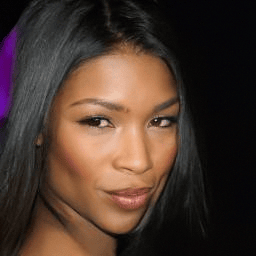} \\
        \includegraphics[width=0.3\linewidth]{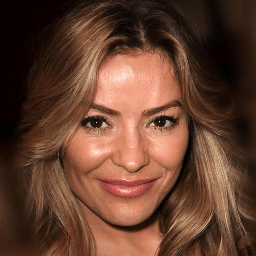} &
        \includegraphics[width=0.3\linewidth]{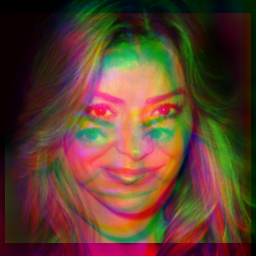} &
        \includegraphics[width=0.3\linewidth]{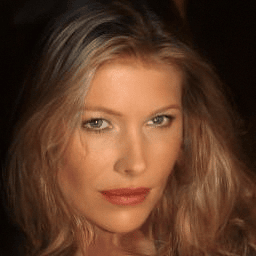} \\
        \includegraphics[width=0.3\linewidth]{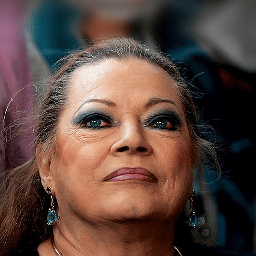} &
        \includegraphics[width=0.3\linewidth]{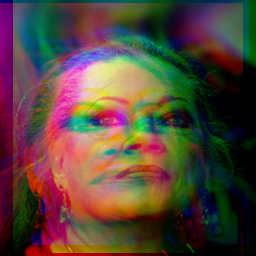} &
        \includegraphics[width=0.3\linewidth]{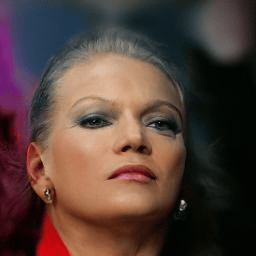} \\
        \multicolumn{1}{c}{\makecell{\scriptsize{Source}}} &
        {\makecell{\scriptsize{Input}}} &
        \multicolumn{1}{c}{\makecell{\scriptsize{\textbf{CODE (ours)}}}}  \\
        \end{tabular}
    \caption{Qualitative results of Chromatic Aberration}
\end{figure}
\newpage


\subsection{Circular Motion Blur}
\begin{table}[htbp]
    \centering
    \scriptsize
    \begin{tabularx}{\columnwidth}{@{}ccccc@{}}
    \toprule[1.5pt]
     & LPIPS-Source $\downarrow$ &  SSIM-Input $\uparrow$ & PSNR-Input $\uparrow$ & FID $\downarrow$ \\
    \midrule[1pt]
    \makecell{\textit{Inputs}} & 0.33 & - & - & 164.88 \\
    \midrule[1pt]
    \makecell{\textit{DDIM}\\w/o CBC} & 0.32 & 0.87 & 30.3 & 154.69 \\
    \midrule[1pt]
    \makecell{\textit{DDIM}\\w/ CBC} & 0.35 & 0.9 & 29.19 & 98.65 \\
    \midrule[1pt]
    \makecell{\textbf{CODE}\\\textbf{w/o CBC}} & 0.26 & 0.67 & 22.14 & 34.1 \\
    \midrule[1pt]
    \makecell{\textbf{CODE}\\\textbf{w/ CBC}} & 0.23 & 0.71 & 23.94 & 26.98 \\
    \midrule[1pt]
    \makecell{\textit{SDEdit}} & 0.23 & 0.71 & 23.61 & 37.33 \\
    \bottomrule[1.5pt]
    \end{tabularx}
    \caption{Quantitative results on Circular Motion Blur}
\end{table}
\begin{figure}[h!]
    \centering
    \setlength{\tabcolsep}{0pt}
    \renewcommand{\arraystretch}{0} 
    \begin{tabular}{c@{}c@{}c}
        \includegraphics[width=0.3\linewidth]{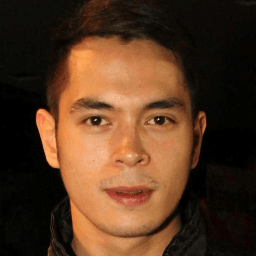} &
        \includegraphics[width=0.3\linewidth]{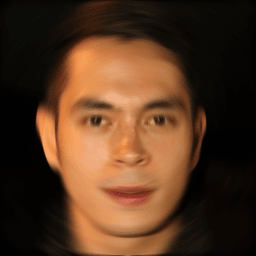} &
        \includegraphics[width=0.3\linewidth]{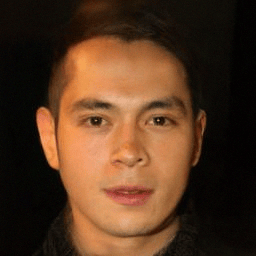} \\
        \includegraphics[width=0.3\linewidth]{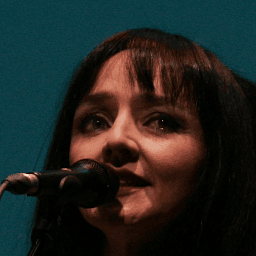} &
        \includegraphics[width=0.3\linewidth]{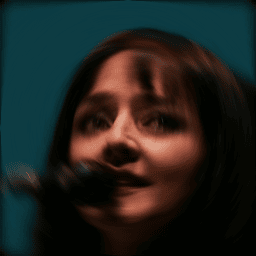} &
        \includegraphics[width=0.3\linewidth]{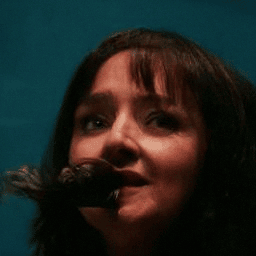} \\
        \includegraphics[width=0.3\linewidth]{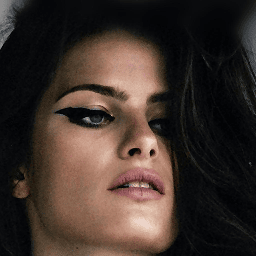} &
        \includegraphics[width=0.3\linewidth]{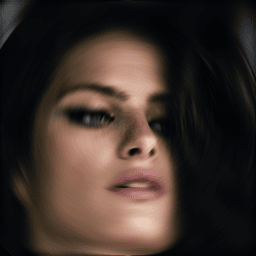} &
        \includegraphics[width=0.3\linewidth]{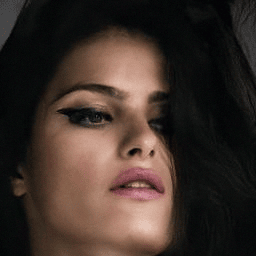} \\
        \includegraphics[width=0.3\linewidth]{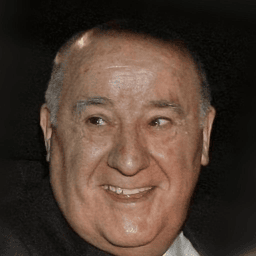} &
        \includegraphics[width=0.3\linewidth]{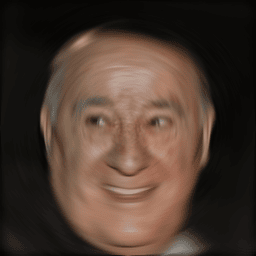} &
        \includegraphics[width=0.3\linewidth]{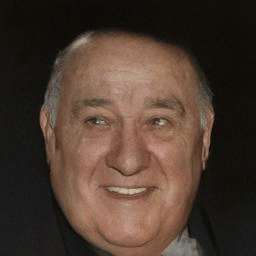} \\
        \includegraphics[width=0.3\linewidth]{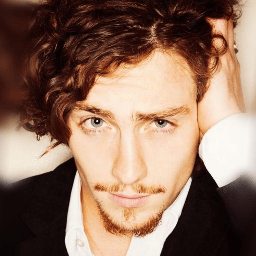} &
        \includegraphics[width=0.3\linewidth]{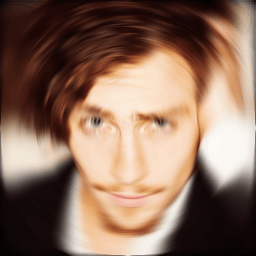} &
        \includegraphics[width=0.3\linewidth]{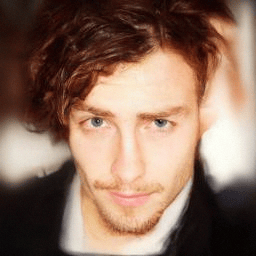} \\
        \multicolumn{1}{c}{\makecell{\scriptsize{Source}}} &
        {\makecell{\scriptsize{Input}}} &
        \multicolumn{1}{c}{\makecell{\scriptsize{\textbf{CODE (ours)}}}}  \\
        \end{tabular}
    \caption{Qualitative results of Circular Motion Blur}
\end{figure}
\newpage


\subsection{Color Dither}
\begin{table}[htbp]
    \centering
    \scriptsize
    \begin{tabularx}{\columnwidth}{@{}ccccc@{}}
    \toprule[1.5pt]
     & LPIPS-Source $\downarrow$ &  SSIM-Input $\uparrow$ & PSNR-Input $\uparrow$ & FID $\downarrow$ \\
    \midrule[1pt]
    \makecell{\textit{Inputs}} & 0.1 & - & - & 35.14 \\
    \midrule[1pt]
    \makecell{\textit{DDIM}\\w/o CBC} & 0.12 & 0.97 & 35.19 & 48.96 \\
    \midrule[1pt]
    \makecell{\textit{DDIM}\\w/ CBC} & 0.15 & 0.71 & 24.45 & 34.59 \\
    \midrule[1pt]
    \makecell{\textbf{CODE}\\\textbf{w/o CBC}} & 0.11 & 0.7 & 26.32 & 26.55 \\
    \midrule[1pt]
    \makecell{\textbf{CODE}\\\textbf{w/ CBC}} & 0.14 & 0.51 & 22.88 & 16.17 \\
    \midrule[1pt]
    \makecell{\textit{SDEdit}} & 0.16 & 0.43 & 21.12 & 25.31 \\
    \bottomrule[1.5pt]
    \end{tabularx}
    \caption{Quantitative results on Color Dither}
\end{table}
\begin{figure}[h!]
    \centering
    \setlength{\tabcolsep}{0pt}
    \renewcommand{\arraystretch}{0} 
    \begin{tabular}{c@{}c@{}c}
        \includegraphics[width=0.3\linewidth]{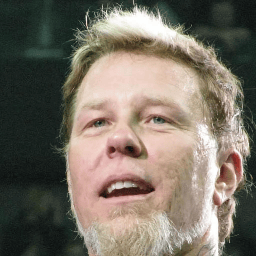} &
        \includegraphics[width=0.3\linewidth]{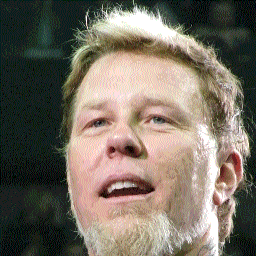} &
        \includegraphics[width=0.3\linewidth]{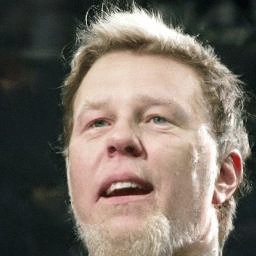} \\
        \includegraphics[width=0.3\linewidth]{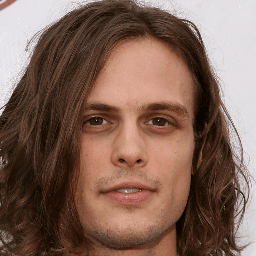} &
        \includegraphics[width=0.3\linewidth]{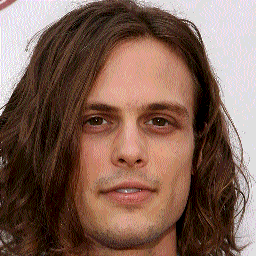} &
        \includegraphics[width=0.3\linewidth]{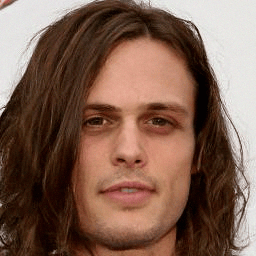} \\
        \includegraphics[width=0.3\linewidth]{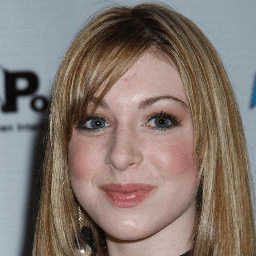} &
        \includegraphics[width=0.3\linewidth]{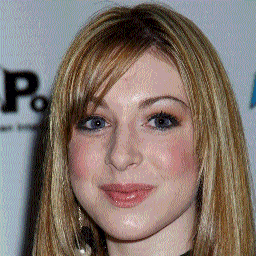} &
        \includegraphics[width=0.3\linewidth]{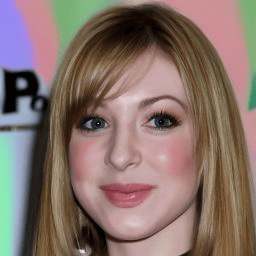} \\
        \includegraphics[width=0.3\linewidth]{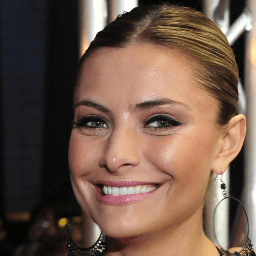} &
        \includegraphics[width=0.3\linewidth]{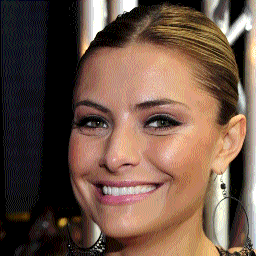} &
        \includegraphics[width=0.3\linewidth]{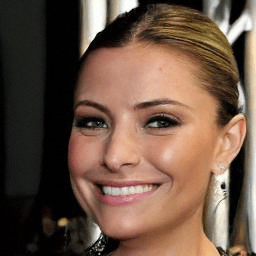} \\
        \includegraphics[width=0.3\linewidth]{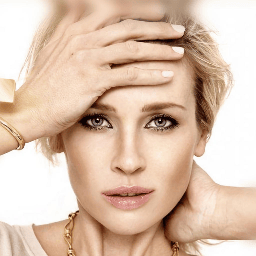} &
        \includegraphics[width=0.3\linewidth]{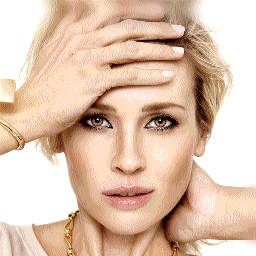} &
        \includegraphics[width=0.3\linewidth]{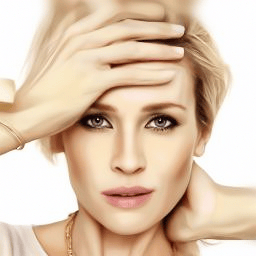} \\
        \multicolumn{1}{c}{\makecell{\scriptsize{Source}}} &
        {\makecell{\scriptsize{Input}}} &
        \multicolumn{1}{c}{\makecell{\scriptsize{\textbf{CODE (ours)}}}}  \\
        \end{tabular}
    \caption{Qualitative results of Color Dither}
\end{figure}
\newpage


\subsection{Contrast}
\begin{table}[htbp]
    \centering
    \scriptsize
    \begin{tabularx}{\columnwidth}{@{}ccccc@{}}
    \toprule[1.5pt]
     & LPIPS-Source $\downarrow$ &  SSIM-Input $\uparrow$ & PSNR-Input $\uparrow$ & FID $\downarrow$ \\
    \midrule[1pt]
    \makecell{\textit{Inputs}} & 0.75 & - & - & 83.87 \\
    \midrule[1pt]
    \makecell{\textit{DDIM}\\w/o CBC} & 0.6 & 0.76 & 28.54 & 92.48 \\
    \midrule[1pt]
    \makecell{\textit{DDIM}\\w/ CBC} & 0.85 & 0.97 & 36.55 & 279.15 \\
    \midrule[1pt]
    \makecell{\textbf{CODE}\\\textbf{w/o CBC}} & 0.54 & 0.72 & 23.71 & 76.05 \\
    \midrule[1pt]
    \makecell{\textbf{CODE}\\\textbf{w/ CBC}} & 0.65 & 0.82 & 25.73 & 107.31 \\
    \midrule[1pt]
    \makecell{\textit{SDEdit}} & 0.79 & 0.81 & 26.56 & 259.42 \\
    \bottomrule[1.5pt]
    \end{tabularx}
    \caption{Quantitative results on Contrast}
\end{table}
\begin{figure}[h!]
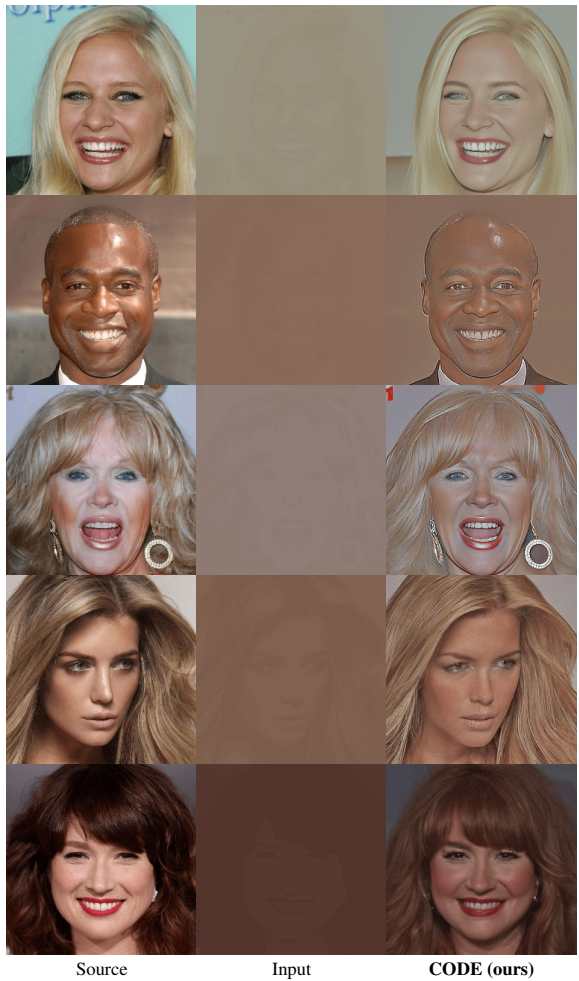

    \centering
    \setlength{\tabcolsep}{0pt}
    \renewcommand{\arraystretch}{0} 
    %
    \caption{Qualitative results of Contrast}
\end{figure}
\newpage


\subsection{Elastic Transform}
    \caption{Quantitative results on Elastic Transform}
\end{table}
\begin{figure}[h!]
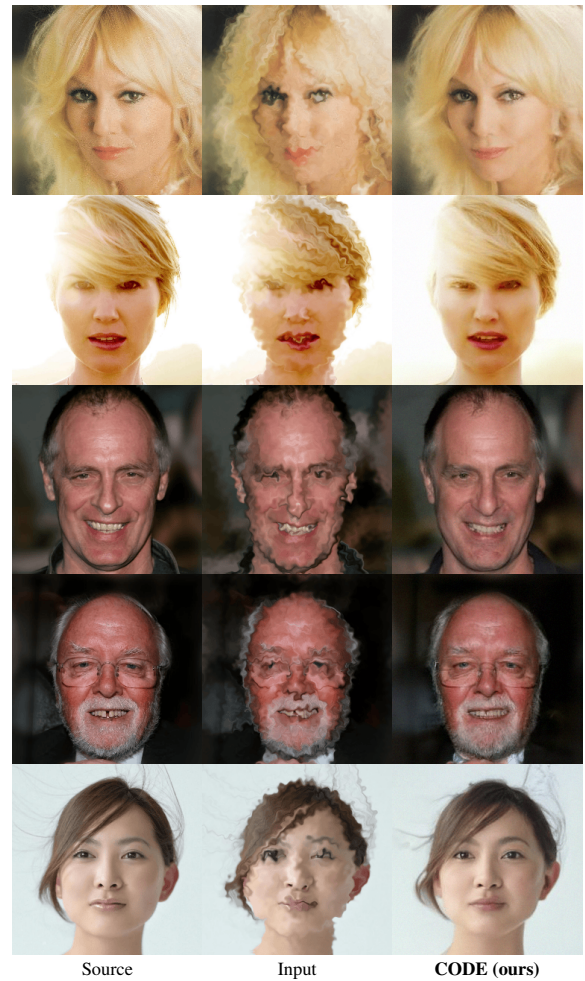

    \centering
    \setlength{\tabcolsep}{0pt}
    \renewcommand{\arraystretch}{0} 
    %
    \caption{Qualitative results of Elastic Transform}
\end{figure}
\newpage


\subsection{Fog}
    \caption{Quantitative results on Fog}
\end{table}
\begin{figure}[h!]
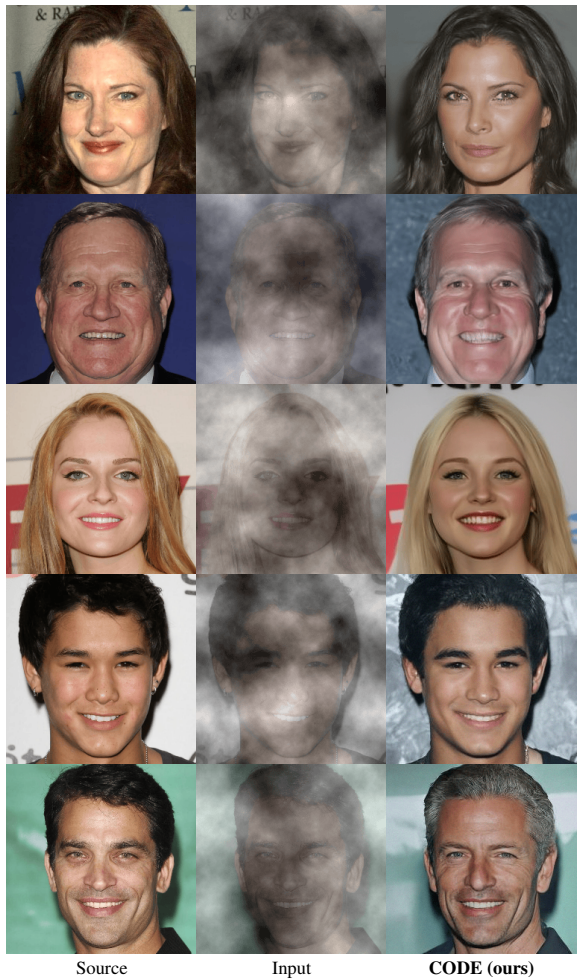

    \centering
    \setlength{\tabcolsep}{0pt}
    \renewcommand{\arraystretch}{0} 
    %
    \caption{Qualitative results of Fog}
\end{figure}
\newpage


\subsection{Frost}
    \caption{Quantitative results on Frost}
\end{table}
\begin{figure}[h!]
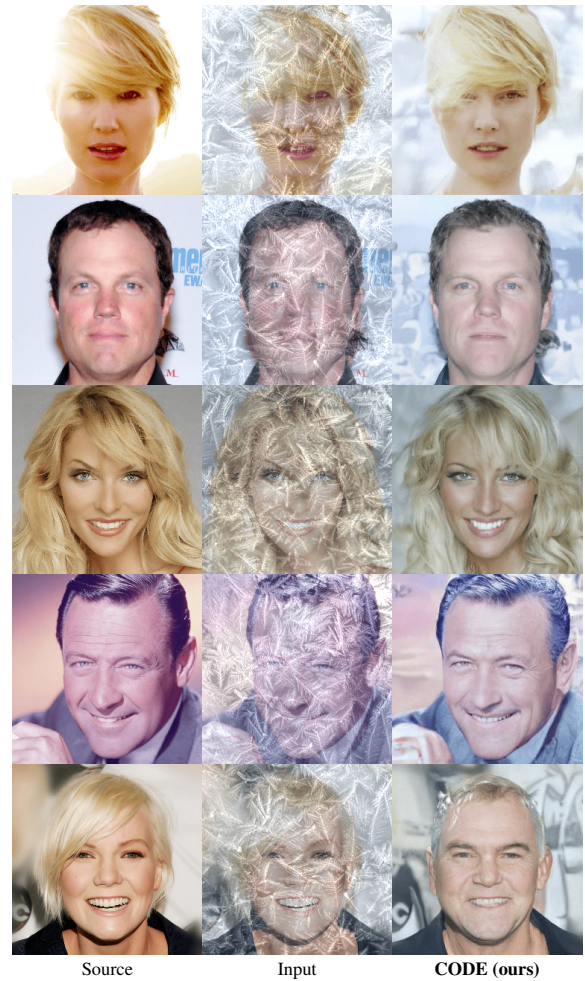

    \centering
    \setlength{\tabcolsep}{0pt}
    \renewcommand{\arraystretch}{0} 
    %
    \caption{Qualitative results of Frost}
\end{figure}
\newpage


\subsection{Gaussian Noise}
    \caption{Quantitative results on Gaussian Noise}
\end{table}
\begin{figure}[h!]
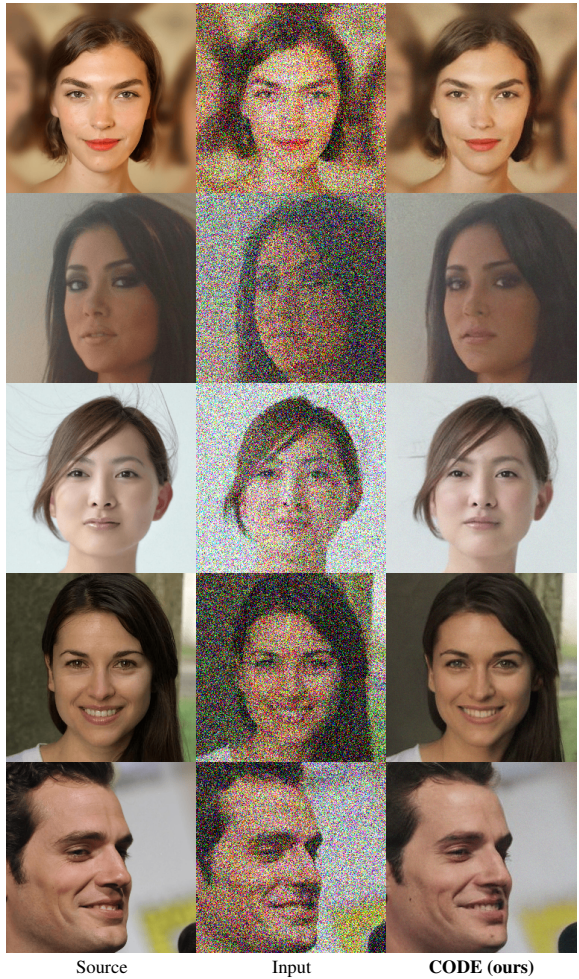

    \centering
    \setlength{\tabcolsep}{0pt}
    \renewcommand{\arraystretch}{0} 
    %
    \caption{Qualitative results of Gaussian Noise}
\end{figure}
\newpage


\subsection{Glass Blur}
    \caption{Quantitative results on Glass Blur}
\end{table}
\begin{figure}[h!]
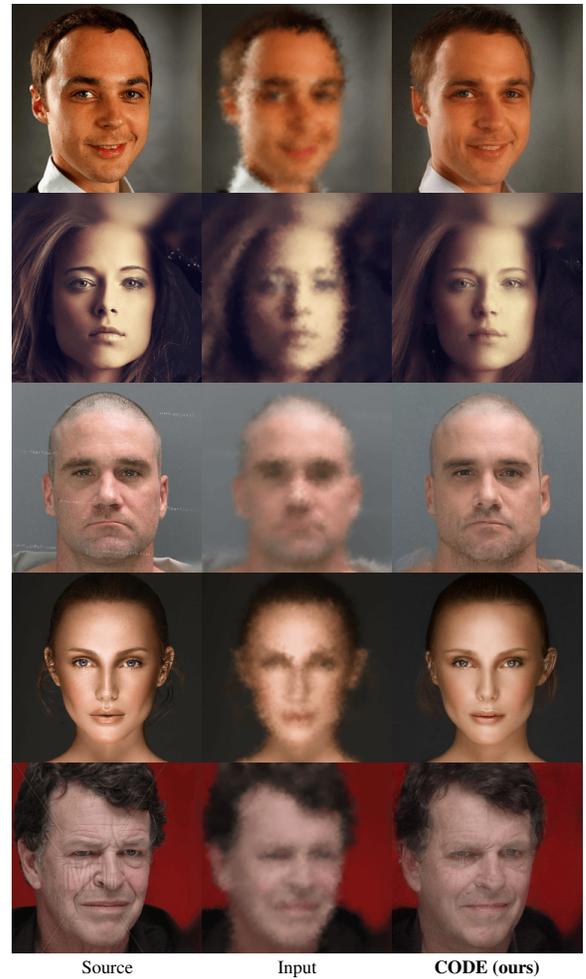

    \centering
    \setlength{\tabcolsep}{0pt}
    \renewcommand{\arraystretch}{0} 
    %
    \caption{Qualitative results of Glass Blur}
\end{figure}
\newpage


\subsection{Hue Shift}
\begin{table}[htbp]
    \centering
    \scriptsize
    \begin{tabularx}{\columnwidth}{@{}ccccc@{}}
    \toprule[1.5pt]
     & LPIPS-Source $\downarrow$ &  SSIM-Input $\uparrow$ & PSNR-Input $\uparrow$ & FID $\downarrow$ \\
    \midrule[1pt]
    \makecell{\textit{Inputs}} & 0.26 & - & - & 56.48 \\
    \midrule[1pt]
    \makecell{\textit{DDIM}\\w/o CBC} & 0.32 & 0.9 & 31.38 & 82.7 \\
    \midrule[1pt]
    \makecell{\textit{DDIM}\\w/ CBC} & 0.44 & 0.78 & 25.39 & 89.07 \\
    \midrule[1pt]
    \makecell{\textbf{CODE}\\\textbf{w/o CBC}} & 0.34 & 0.71 & 23.26 & 49.63 \\
    \midrule[1pt]
    \makecell{\textbf{CODE}\\\textbf{w/ CBC}} & 0.38 & 0.65 & 22.79 & 37.72 \\
    \midrule[1pt]
    \makecell{\textit{SDEdit}} & 0.38 & 0.6 & 20.85 & 48.58 \\
    \bottomrule[1.5pt]
    \end{tabularx}
    \caption{Quantitative results on Hue Shift}
\end{table}
    \caption{Qualitative results of Hue Shift}
\end{figure}
\newpage


\subsection{Impulse Noise}
    \caption{Quantitative results on Impulse Noise}
\end{table}
\begin{figure}[h!]
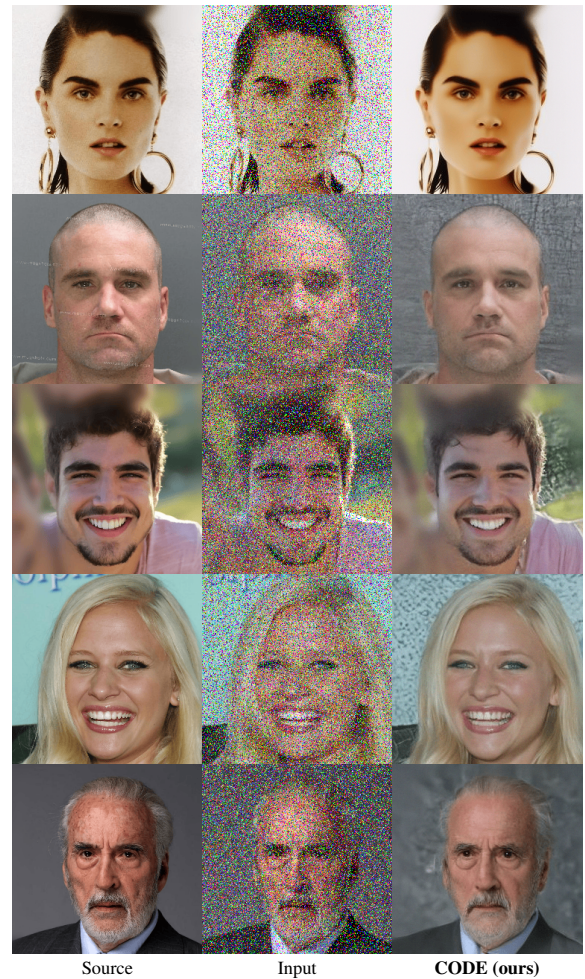

    \centering
    \setlength{\tabcolsep}{0pt}
    \renewcommand{\arraystretch}{0} 
    %
    \caption{Qualitative results of Impulse Noise}
\end{figure}
\newpage


\subsection{Inverse Sparkles}
    \caption{Quantitative results on Inverse Sparkles}
\end{table}
\begin{figure}[h!]
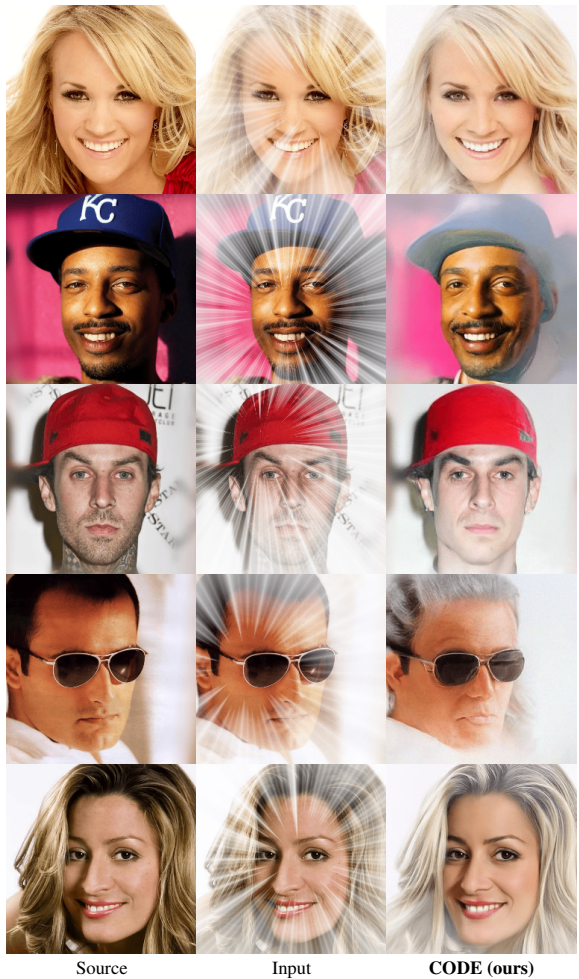

    \centering
    \setlength{\tabcolsep}{0pt}
    \renewcommand{\arraystretch}{0} 
    %
    \caption{Qualitative results of Inverse Sparkles}
\end{figure}
\newpage


\subsection{Jpeg Compression}
    \caption{Quantitative results on Jpeg Compression}
\end{table}
\begin{figure}[h!]
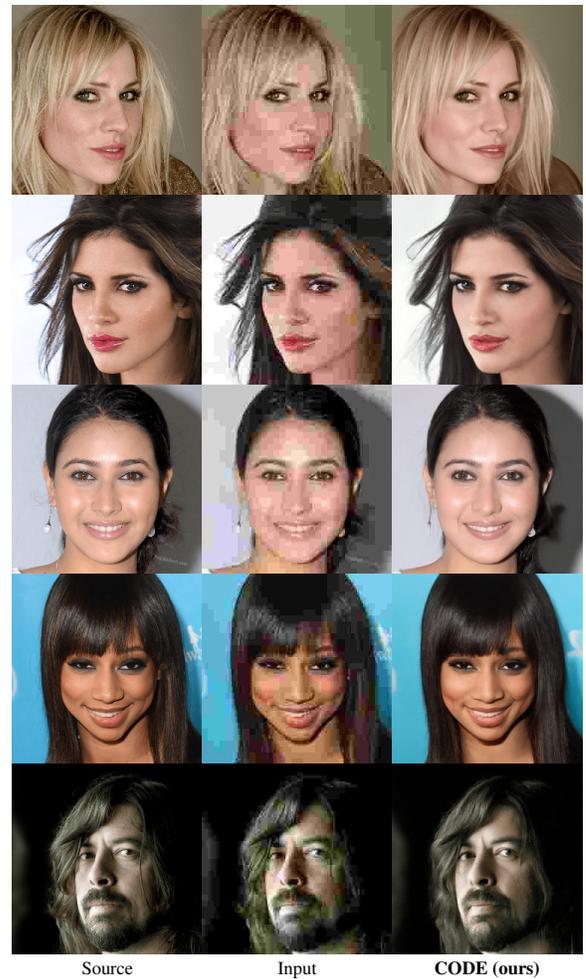

    \centering
    \setlength{\tabcolsep}{0pt}
    \renewcommand{\arraystretch}{0} 
    %
    \caption{Qualitative results of Jpeg Compression}
\end{figure}
\newpage


\subsection{Lines}
    \caption{Quantitative results on Lines}
\end{table}
\begin{figure}[h!]
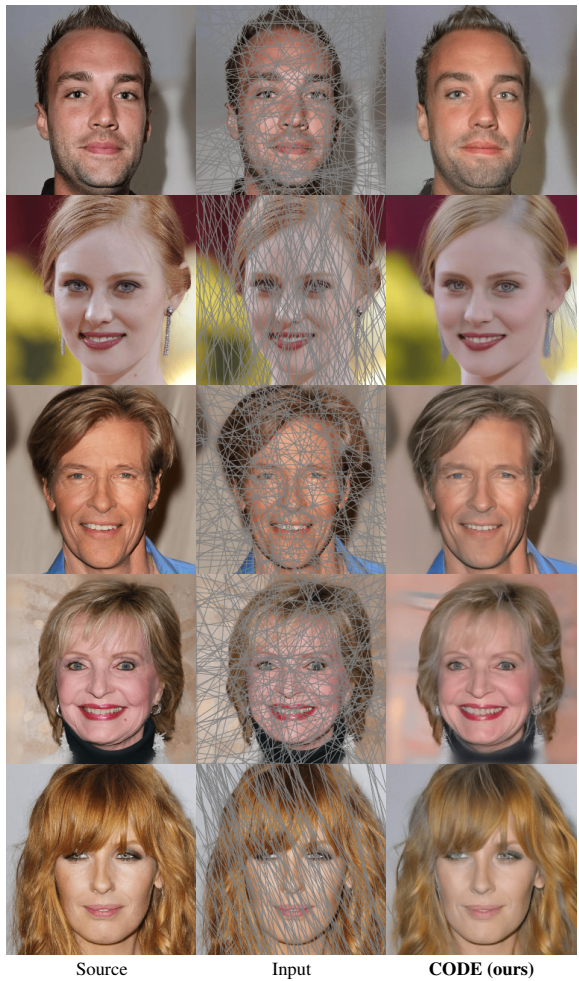

    \centering
    \setlength{\tabcolsep}{0pt}
    \renewcommand{\arraystretch}{0} 
    %
    \caption{Qualitative results of Lines}
\end{figure}
\newpage


\subsection{Masking Random Color}
    \caption{Quantitative results on Masking Random Color}
\end{table}
\begin{figure}[h!]
    \centering
    \setlength{\tabcolsep}{0pt}
    \renewcommand{\arraystretch}{0} 
    %
    \caption{Qualitative results of Masking Random Color}
\end{figure}
\newpage


\subsection{Masking Vline Random Color}
    \caption{Quantitative results on Masking Vline Random Color}
\end{table}
\begin{figure}[h!]
    \centering
    \setlength{\tabcolsep}{0pt}
    \renewcommand{\arraystretch}{0} 
    %
    \caption{Qualitative results of Masking Vline Random Color}
\end{figure}
\newpage


\subsection{Motion Blur}
\begin{table}[htbp]
    \centering
    \scriptsize
    %
    \caption{Quantitative results on Motion Blur}
\end{table}
    \caption{Qualitative results of Motion Blur}
\end{figure}
\newpage


\subsection{Perlin Noise}
    \caption{Quantitative results on Perlin Noise}
\end{table}
\begin{figure}[h!]
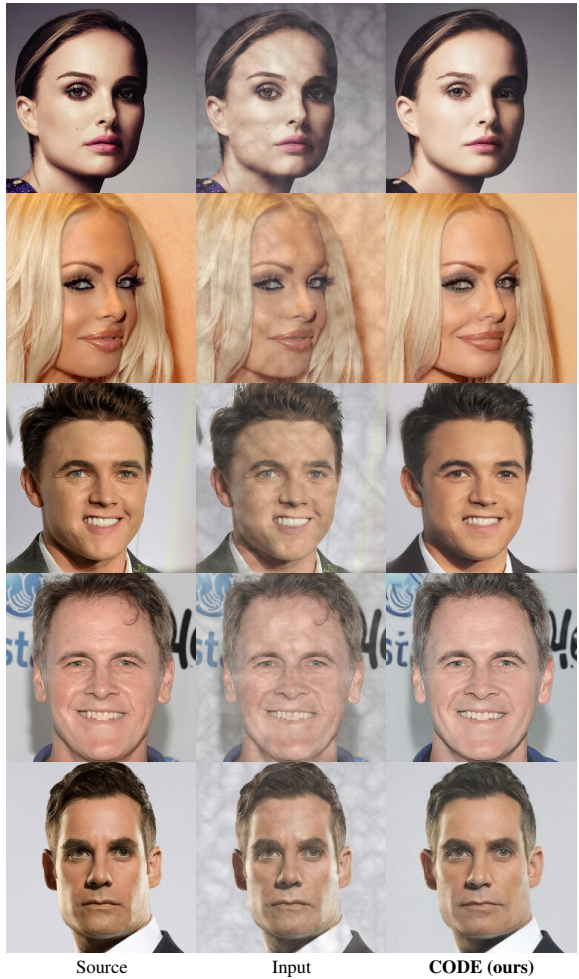

    \centering
    \setlength{\tabcolsep}{0pt}
    \renewcommand{\arraystretch}{0} 
    %
    \caption{Qualitative results of Perlin Noise}
\end{figure}
\newpage


\subsection{Perspective No Bars}
    \caption{Quantitative results on Perspective No Bars}
\end{table}
\begin{figure}[h!]
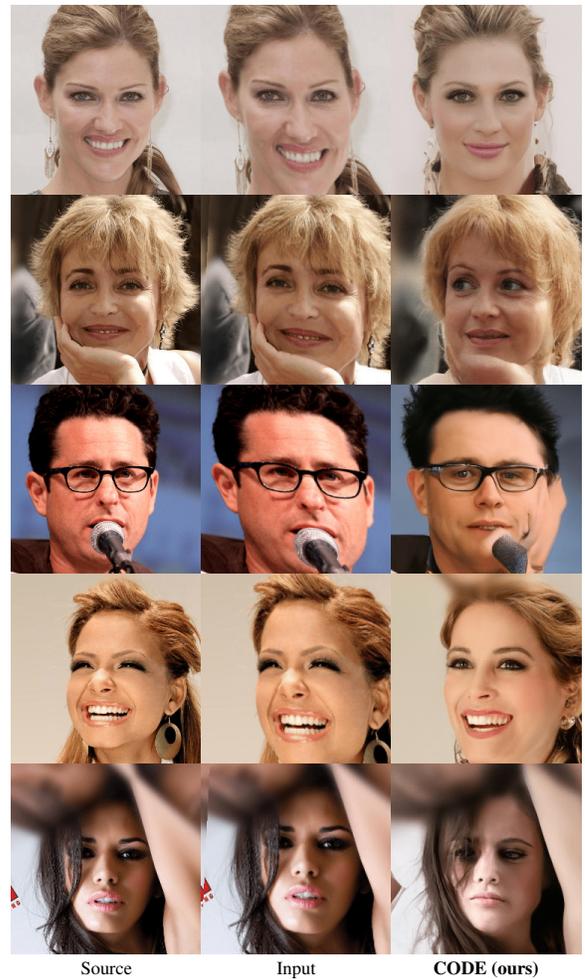

    \centering
    \setlength{\tabcolsep}{0pt}
    \renewcommand{\arraystretch}{0} 
    %
    \caption{Qualitative results of Perspective No Bars}
\end{figure}
\newpage


\subsection{Plasma Noise}
\begin{table}[htbp]
    \centering
    \scriptsize
    %
    \caption{Quantitative results on Plasma Noise}
\end{table}
    \caption{Qualitative results of Plasma Noise}
\end{figure}
\newpage


\subsection{Pseudocolor}
    \caption{Quantitative results on Pseudocolor}
\end{table}
\begin{figure}[h!]
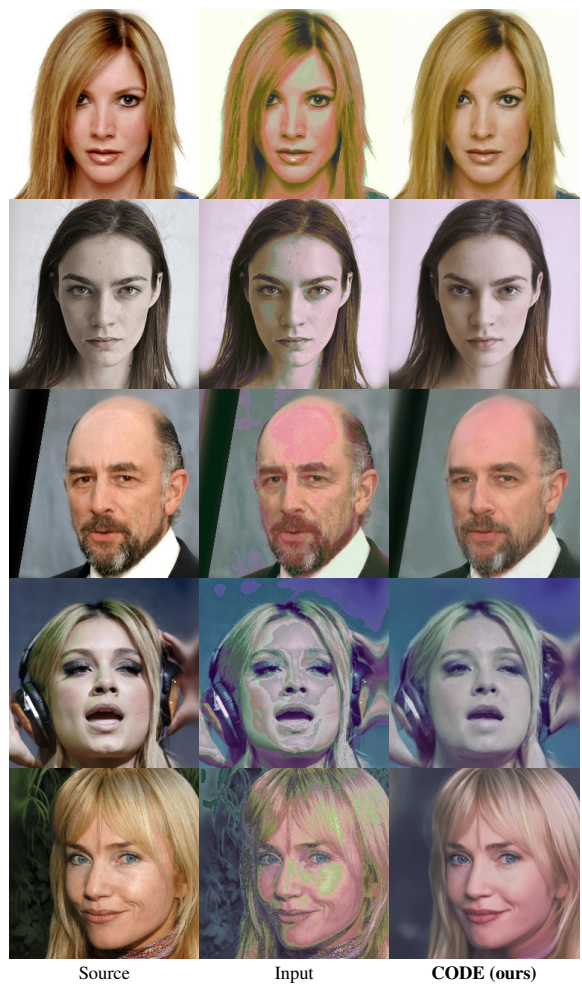

    \centering
    \setlength{\tabcolsep}{0pt}
    \renewcommand{\arraystretch}{0} 
    %
    \caption{Qualitative results of Pseudocolor}
\end{figure}
\newpage


\subsection{Quadrilateral No Bars}
    \caption{Quantitative results on Quadrilateral No Bars}
\end{table}
\begin{figure}[h!]
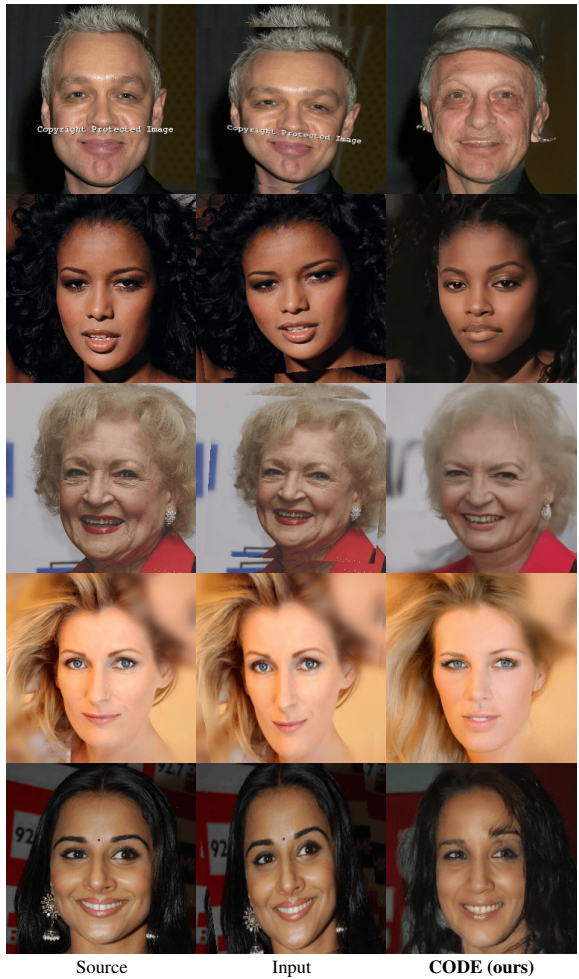

    \centering
    \setlength{\tabcolsep}{0pt}
    \renewcommand{\arraystretch}{0} 
    %
    \caption{Qualitative results of Quadrilateral No Bars}
\end{figure}
\newpage


\subsection{Ripple}
\begin{table}[htbp]
    \centering
    \scriptsize
    %
    \caption{Quantitative results on Ripple}
\end{table}
    \caption{Qualitative results of Ripple}
\end{figure}
\newpage


\subsection{Saturate}
    \caption{Quantitative results on Saturate}
\end{table}
\begin{figure}[h!]
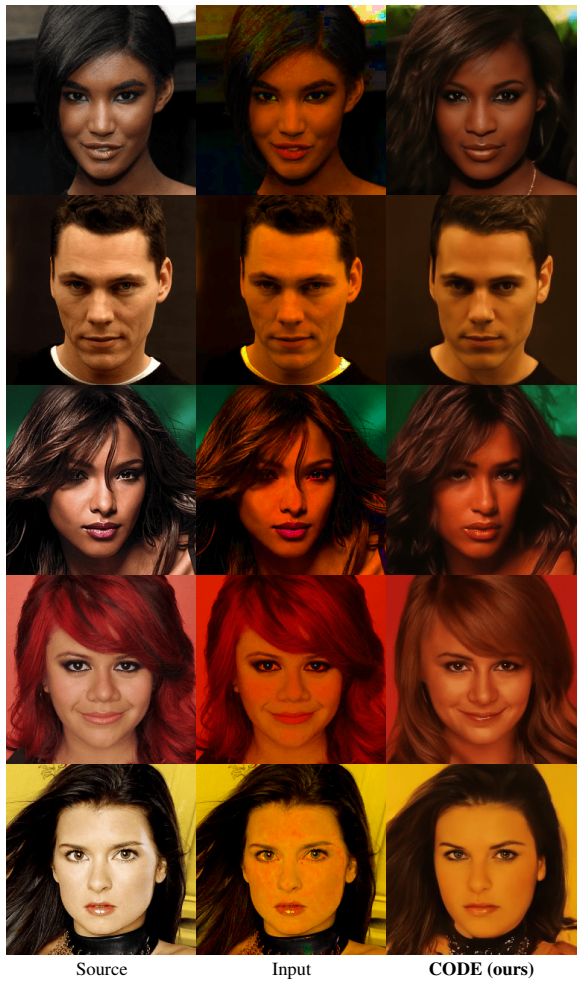

    \centering
    \setlength{\tabcolsep}{0pt}
    \renewcommand{\arraystretch}{0} 
    %
    \caption{Qualitative results of Saturate}
\end{figure}
\newpage


\subsection{Shot Noise}
    \caption{Quantitative results on Shot Noise}
\end{table}
\begin{figure}[h!]
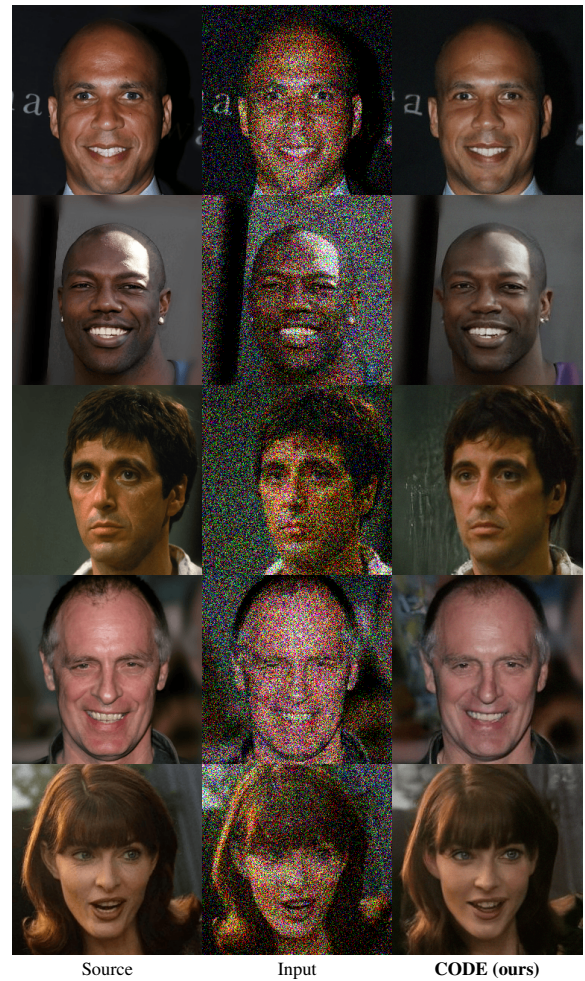

    \centering
    \setlength{\tabcolsep}{0pt}
    \renewcommand{\arraystretch}{0} 
    %
    \caption{Qualitative results of Shot Noise}
\end{figure}
\newpage


\subsection{Single Frequency Greyscale}
    \caption{Quantitative results on Single Frequency Greyscale}
\end{table}
\begin{figure}[h!]
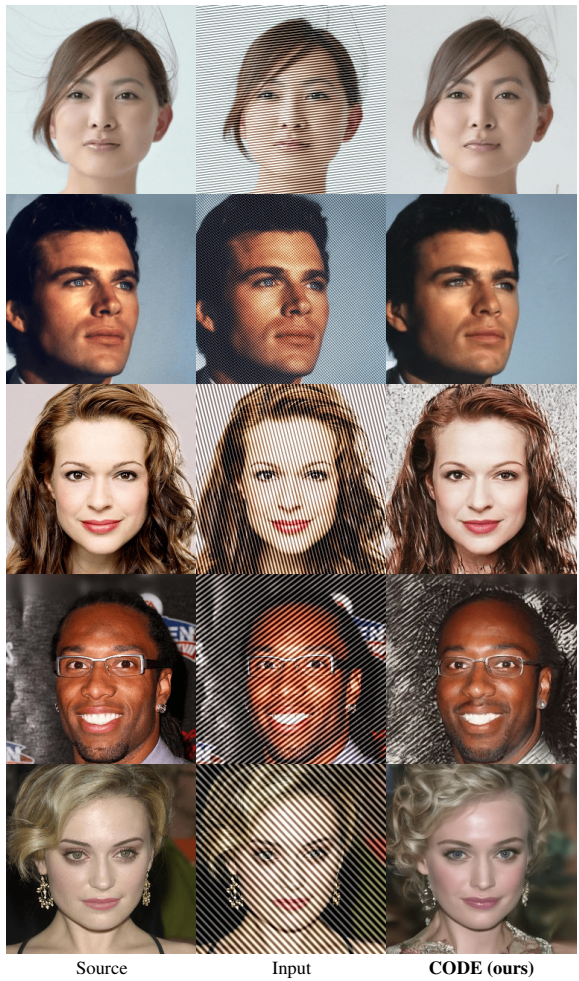

    \centering
    \setlength{\tabcolsep}{0pt}
    \renewcommand{\arraystretch}{0} 
    %
    \caption{Qualitative results of Single Frequency Greyscale}
\end{figure}
\newpage


\subsection{Snow}
    \caption{Quantitative results on Snow}
\end{table}
\begin{figure}[h!]
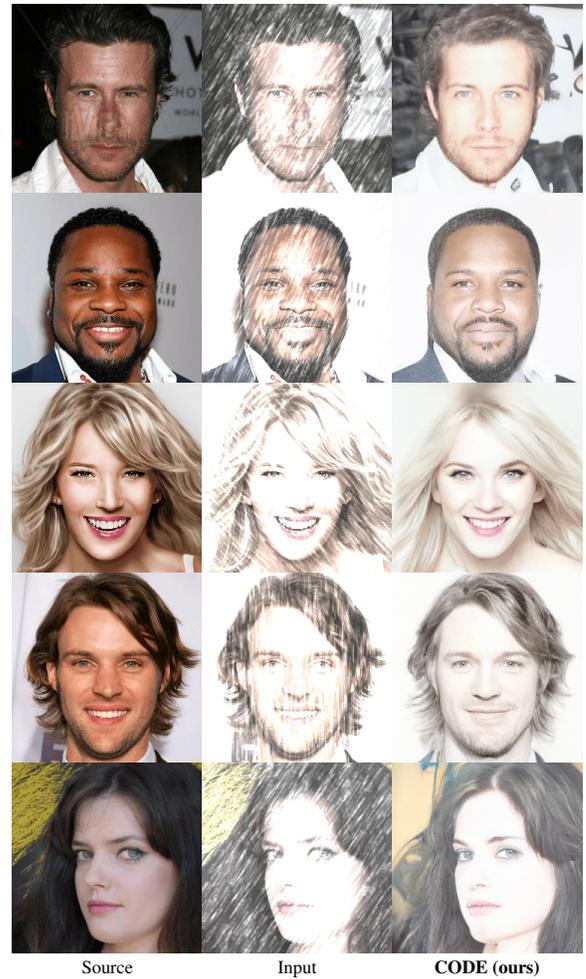

    \centering
    \setlength{\tabcolsep}{0pt}
    \renewcommand{\arraystretch}{0} 
    %
    \caption{Qualitative results of Snow}
\end{figure}
\newpage


\subsection{Spatter}
    \caption{Quantitative results on Spatter}
\end{table}
\begin{figure}[h!]
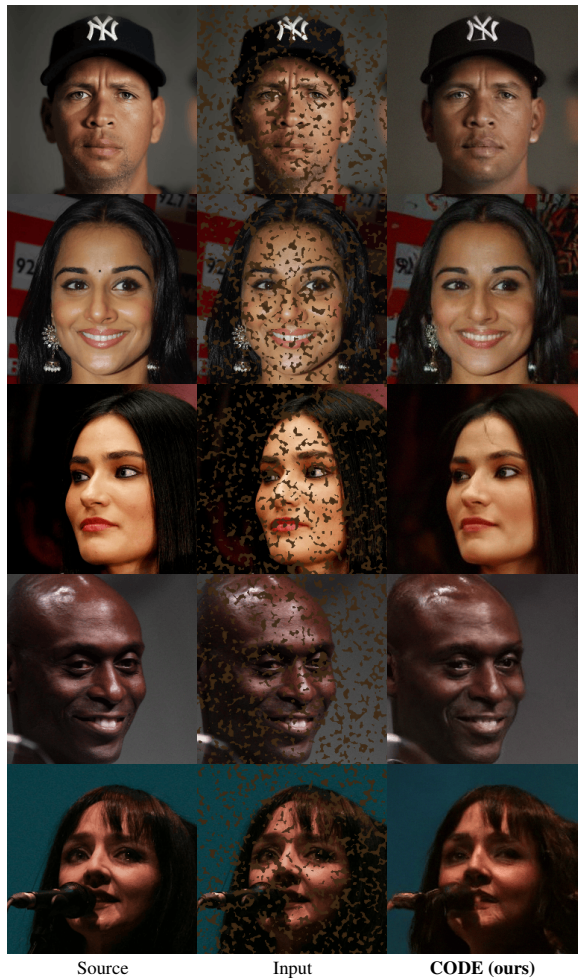

    \centering
    \setlength{\tabcolsep}{0pt}
    \renewcommand{\arraystretch}{0} 
    %
    \caption{Qualitative results of Spatter}
\end{figure}
\newpage


\subsection{Speckle Noise}
    \caption{Quantitative results on Speckle Noise}
\end{table}
\begin{figure}[h!]
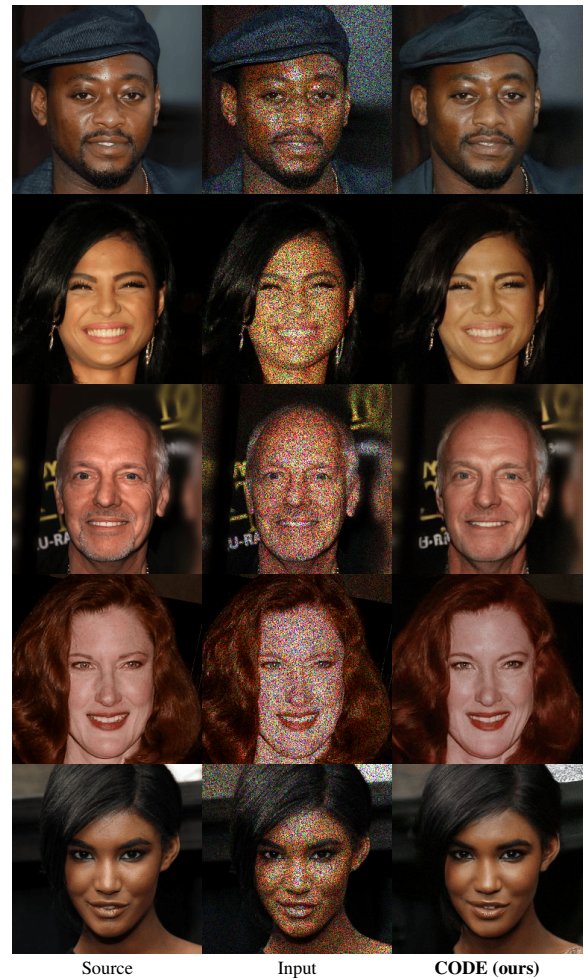

    \centering
    \setlength{\tabcolsep}{0pt}
    \renewcommand{\arraystretch}{0} 
    %
    \caption{Qualitative results of Speckle Noise}
\end{figure}
\newpage


\subsection{Technicolor}
    \caption{Quantitative results on Technicolor}
\end{table}
\begin{figure}[h!]
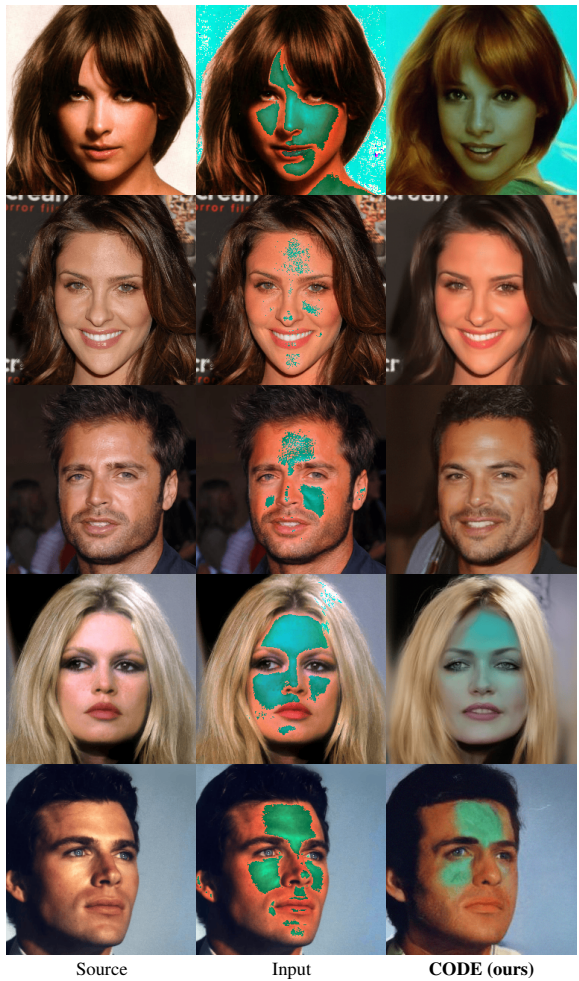

    \centering
    \setlength{\tabcolsep}{0pt}
    \renewcommand{\arraystretch}{0} 
    %
    \caption{Qualitative results of Technicolor}
\end{figure}
\newpage


\subsection{Transverse Chromatic Abberation}
    \caption{Quantitative results on Transverse Chromatic Abberation}
\end{table}
\begin{figure}[h!]
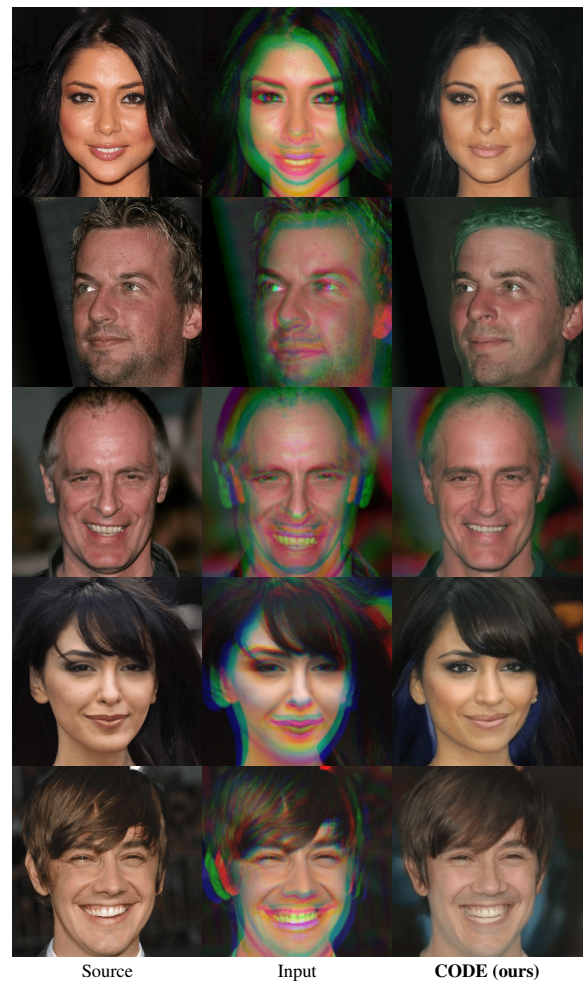

    \centering
    \setlength{\tabcolsep}{0pt}
    \renewcommand{\arraystretch}{0} 
    %
    \caption{Qualitative results of Transverse Chromatic Abberation}
\end{figure}
\newpage


\subsection{Voronoi Noise}
    \caption{Quantitative results on Voronoi Noise}
\end{table}
\begin{figure}[h!]
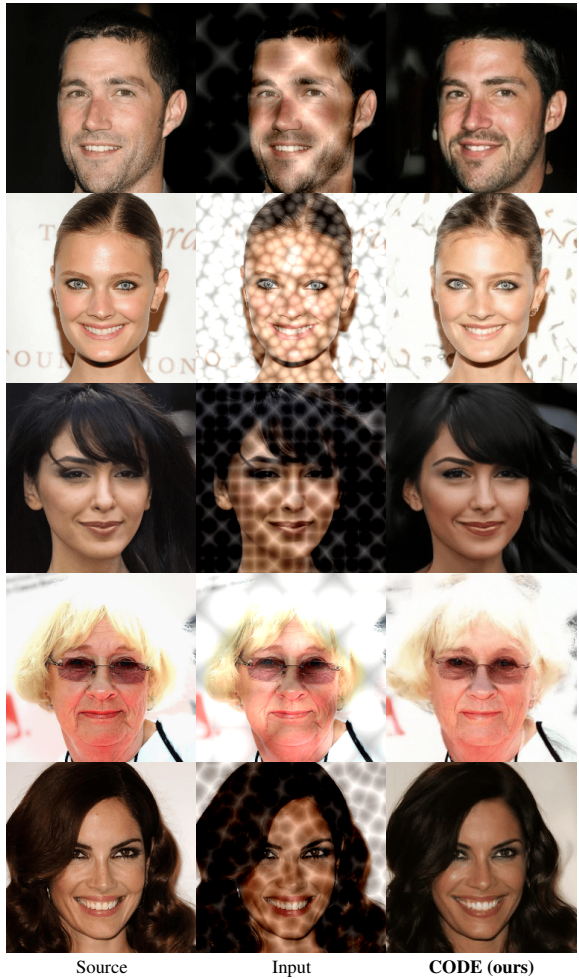

    \centering
    \setlength{\tabcolsep}{0pt}
    \renewcommand{\arraystretch}{0} 
    %
    \caption{Qualitative results of Voronoi Noise}
\end{figure}



\end{document}